\documentclass[10pt]{wlscirep}
\usepackage[utf8]{inputenc}
\usepackage[T1]{fontenc}
\usepackage{colortbl}
\usepackage{bm}
\usepackage{marvosym}
\usepackage{float}
\usepackage{hyperref}
\usepackage{lineno}
\usepackage{booktabs}
\usepackage{tabularx}
\usepackage{ragged2e}
\usepackage{array}
\usepackage{placeins}
\usepackage{threeparttable}
 \usepackage{multirow}
 \usepackage{makecell}
\usepackage[table]{xcolor}
\usepackage{longtable}
\usepackage{graphicx}
\usepackage{afterpage}
\usepackage{xurl}
\usepackage{seqsplit}
\newlength{\heatw}
\newcolumntype{H}{>{\centering\arraybackslash}m{\heatw}}

\makeatletter
\let\oldthebibliography\thebibliography
\let\endoldthebibliography\endthebibliography

\renewenvironment{thebibliography}[1]
  {\oldthebibliography{#1}%
   \setlength{\itemsep}{0pt}%
   \setlength{\parsep}{0pt}%
   \setlength{\parskip}{0pt}}
  {\endoldthebibliography}
\makeatother

\hypersetup{breaklinks=true}

\newcolumntype{Y}{>{\RaggedRight\arraybackslash}X}

\usepackage[justification=raggedright,singlelinecheck=false]{caption}
\hypersetup{
    colorlinks=true,
    urlcolor=black,
}

\title{A Pathology Foundation Model for Gastric Cancer with Real-World Validation}

\author[1,*]{Ling Liang}
\author[1,*]{Jiabo Ma}
\author[2,3,4,*]{Zhengyu Zhang}
\author[1]{Fengtao Zhou}
\author[1]{Yingxue Xu}
\author[1]{Yihui Wang}

\author[1]{Cheng Jin}
\author[1]{Zhengrui Guo}

\author[5,6]{On Ki Tang}

\author[2,3,4]{Zhijian Cen}
\author[2,3,4]{Zhen Wang}
\author[2,3,4]{Qi Xie}
\author[2,3,4]{Chengyu Lu}
\author[2,3,4,7]{Chenglong Zhao}
\author[2,3,4]{Feifei Wang}
\author[8]{Yu Cai}
\author[1]{Hongyi Wang}

\author[9]{Jing Zhang}
\author[2,3,4]{Yaping Ye} 
\author[10]{Shijun Sun} 
\author[11,12]{Shenglei Li} 
\author[13]{Yu Wang} 

\author[14]{Zhenhui Li} 
\author[5,6]{Ronald Cheong Kin Chan}
\author[9]{Xiuming Zhang} 
\author[15,16 \Letter]{Zhe Wang}
\author[1,17,18,19,20 \Letter]{Hao Chen}
\author[2,3,4,21 \Letter]{Li Liang}

\affil[1]{Department of Computer Science and Engineering, The Hong Kong University of Science and Technology, Hong Kong SAR, China}

\affil[2]{Department of Pathology, Nanfang Hospital, Southern Medical University, Guangzhou, China}
\affil[3]{Department of Pathology, School of Basic Medical Sciences, Southern Medical University, Guangzhou, China}
\affil[4]{Guangdong Province Key Laboratory of Molecular Tumor Pathology, Guangzhou, China}

\affil[5]{Department of Anatomical and Cellular Pathology, The Chinese University of Hong Kong, Hong Kong SAR, China}
\affil[6]{Pathology Artificial Intelligence Development and Assessment Laboratory, State Key Laboratory of Translational Oncology, The Chinese University of Hong Kong, Hong Kong SAR, China}

\affil[7]{Department of Pathology, Shandong Provincial Qianfoshan Hospital, Jinan, China}

\affil[8]{Department of Electronic and Computer Engineering, The Hong Kong University of Science and Technology, Hong Kong SAR, China}

\affil[9]{Department of Pathology, The First Affiliated Hospital, School of Medicine, Zhejiang University, Hangzhou, China}

\affil[10]{Department of Pathology, Zhongshan People's Hospital, Zhongshan, China}

\affil[11]{Department of Pathology, The First Affiliated Hospital of Zhengzhou University, Zhengzhou, China}

\affil[12]{Department of Pathology, School of Basic Medicine, Zhengzhou University, Zhengzhou, China}

\affil[13]{Department of Pathology, Zhujiang Hospital, Southern Medical University, Guangzhou, China}

\affil[14]{Department of Radiology, The Third Affiliated Hospital of Kunming Medical University, Yunnan Cancer Hospital, Kunming, China}

\affil[15]{State Key Laboratory of Holistic Integrative Management of Gastrointestinal Cancers, Department of Pathology, School of Basic Medicine and Xijing Hospital, Fourth Military Medical University, Xi'an, China}

\affil[16]{Department of Pathology, The First Affiliated Hospital of USTC, Division of Life Sciences and Medicine, University of Science and Technology of China, Hefei, China}

\affil[17]{Department of Chemical and Biological Engineering, The Hong Kong University of Science and Technology, Hong Kong SAR, China}
\affil[18]{Division of Life Science, The Hong Kong University of Science and Technology, Hong Kong SAR, China}
\affil[19]{HKUST Shenzhen-Hong Kong Collaborative Innovation Research Institute, Futian, Shenzhen, China}
\affil[20]{State Key Laboratory of Nervous System Disorders, The Hong Kong University of Science and Technology, Hong Kong SAR, China}
\affil[21]{Jinfeng Laboratory, Chongqing, China}

\affil[*]{Contributed Equally}
\affil[\Letter]{Corresponding Authors}
\affil[ ]{\textbf{Li Liang (lli@smu.edu.cn), Hao Chen (jhc@ust.hk), Zhe Wang (zhwang@fmmu.edu.cn)}}

\begin{abstract}
Gastric cancer remains one of the leading causes of cancer-related mortality worldwide. 
Histopathological examination serves as the cornerstone of gastric cancer diagnosis, yet the disease's remarkable phenotypic and molecular heterogeneity poses substantial challenges. 
Although existing pancancer pathology foundation models (PFMs) offer robust general-purpose representations, their performance frequently plateaus on fine-grained gastric endpoints. 
Furthermore, few of these models have been rigorously evaluated through prospective validation or clinical reader studies.
To bridge these gaps, we present GRACE, a \textbf{G}astric-specific foundation model for \textbf{R}eal-world \textbf{A}ssessment and \textbf{C}linical d\textbf{E}cision support, covering gastric mucosal lesion diagnosis, tumor histological assessment, molecular profiling, and prognostic prediction. 
To develop GRACE, we leveraged a large-scale, multicenter gastric pathology dataset totaling 48,364 primarily H\&E-stained whole-slide images from 37,493 patients, and adapted Virchow2 through gastric-specific continued pretraining using a low-rank adaptation strategy.
When comprehensively evaluated on 28 clinically relevant tasks (28 internal, 11 external, and 33 prospective validation cohorts), GRACE consistently outperformed representative pancancer PFMs, achieving a macro-AUC of 0.9188.
Specifically, the model exhibited exceptional discriminative power in classifying precancerous lesions across the Correa cascade (macro-AUC: 0.9322) and tumor histopathology classification (macro-AUC: 0.9119).
In addition, GRACE demonstrated robust capability for molecular profiling directly from routine H\&E-stained slides, yielding an average macro-AUC of 0.8682. 
Beyond performance benchmarks, GRACE’s translational value was substantiated through a rigorous evidence chain. Under safety-gated criteria requiring 100\% NPV for rule-out and 100\% PPV for rule-in, GRACE could streamline review for up to 69.6\% of malignancy-diagnosis cases and triage 46.8\% of MMR-IHC follow-up requests to non-urgent confirmatory testing.
This translational feasibility was further strengthened by a randomized crossover reader study evaluating pathologist-artificial intelligence collaboration within routine pathology practice. 
With GRACE assistance, diagnostic accuracy significantly improved from 82.0\% to 89.9\%, yielding nearly twofold higher adjusted odds of a correct diagnosis (OR = 1.987, $P < 0.05$) alongside concurrent gains in sensitivity and specificity.
AI assistance also reduced diagnostic time by 14.9\%, elevated diagnostic confidence by 9.0\%, and markedly improved inter-rater agreement.
When calibrated to maintain non-inferior performance to senior pathologists, the AI-assisted workflow could triage 60.7\% of atrophy and 82.7\% of intestinal metaplasia cases.
Ultimately, GRACE advances pathology foundation modeling beyond retrospective benchmarking toward clinically bounded decision support, representing a translatable paradigm for AI-driven precision oncology in gastric disease.

\end{abstract}

\begin{document}
    \flushbottom
    \maketitle
    \thispagestyle{empty}
    \section*{Introduction}
According to GLOBOCAN 2022 data \cite{filho2025globocan}, gastric cancer (GC) remains one of the most aggressive and lethal malignancies, ranking fifth globally in incidence and serving as a leading cause of cancer-related mortality.
Despite substantial regional variations in incidence \cite{lin2024global}, GC represents a major global health challenge, accounting for approximately one million new cases \cite{sundar_gastric_2025,thrift_burden_2020-1} and over 700,000 deaths annually \cite{lin2024global}.
The disease burden is particularly pronounced in East Asian countries, including Japan, Mongolia, the Republic of Korea, and China, which exhibit the highest age-standardized incidence rates globally \cite{shin_updated_2023,mousavi_epidemiology_2025}.
Furthermore, a disproportionate rise in young-onset GC within these populations \cite{morgan_current_2022} continues to amplify its societal and healthcare impact.

The pathogenesis of GC is a multistep, disease-specific process driven by the complex interplay of genetic factors\cite{cancer2014comprehensive}, \textit{Helicobacter pylori} (\textit{H. pylori}) infection \cite{sundar_gastric_2025,thrift_burden_2020-1,Li2023-om,iwu_gastric_2023}, and environmental or lifestyle exposures \cite{qin_burden_2025,liang_risk_2025}.
Morphologically, this progression manifests as a continuum of pathological changes, evolving from chronic gastritis (CG) to atrophy, intestinal metaplasia (IM), and ultimately invasive carcinoma.
Because treatment benefit depends strongly on stage at diagnosis \cite{negura_regulatory_2023,yada2013current,satolli2015gastric}, precise histopathological evaluation of these premalignant lesions and carcinoma is critical for risk stratification and personalized treatment planning.
To address the growing demand for precise pathological evaluation, pathology foundation models (PFMs), such as Virchow2\cite{virchow2}, CONCH\cite{lu_visual-language_2024}, and UNI\cite{chen2024uni}, have been developed, with recent efforts integrating advanced training strategies like unified distillation frameworks\cite{ma2025generalizable} and multimodal context integration\cite{xu2025multimodal}.
\begin{figure}
    \captionsetup{labelformat=empty}
    \centering
    \includegraphics[width=0.95\textwidth]{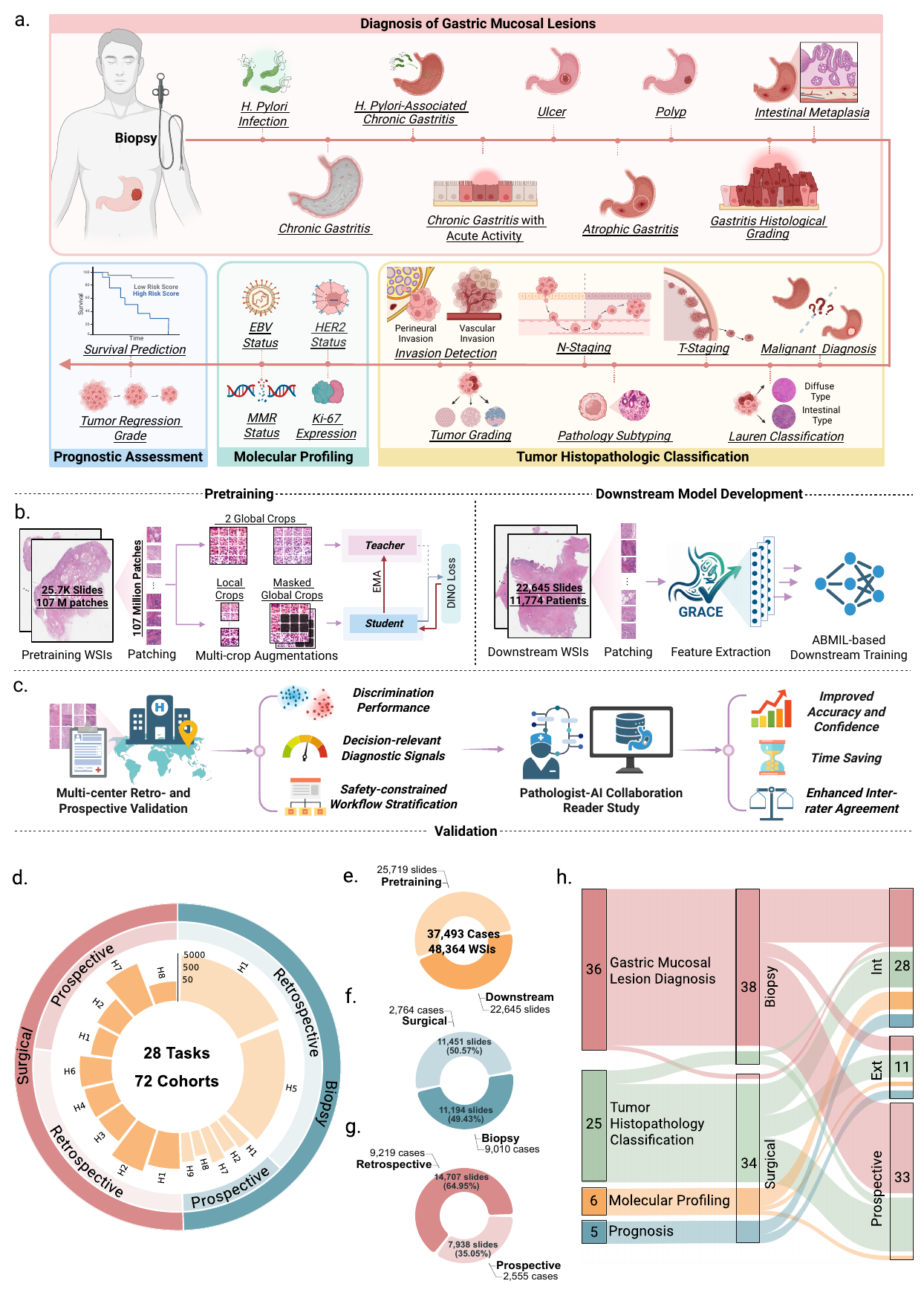}
    \caption{}
\end{figure}
\begin{figure}
    \ContinuedFloat
    \captionsetup{justification=justified, singlelinecheck=false}
    \caption{%
    \textbf{Overview of GRACE.}
        (\textbf{a}) GRACE spans the key clinical endpoints along the gastric pathology workflow, integrating gastric mucosal lesion diagnosis, tumor histological classification, molecular profiling, and prognostic prediction.
        (\textbf{b}) Overview of the GRACE development pipeline, including LoRA-based continued pretraining and downstream model development using the ABMIL framework.
        (\textbf{c}) GRACE was evaluated in multicenter retrospective and prospective external validations to assess diagnostic robustness across diverse clinical settings. Model discrimination was then translated into decision-relevant diagnostic signals to support workflow optimization. We also conducted a pathologist-artificial intelligence (AI) reader study to evaluate the impact of GRACE assistance on diagnostic accuracy, confidence, efficiency, and consistency.
        (\textbf{d}) Cohort distribution across 28 downstream tasks. For each center, the inner bar angular width is proportional to the number of unique slides, and the bar height is log-scaled by the same quantity.
        (\textbf{e}) Slide composition of the full gastric dataset used for model pretraining and downstream evaluation.
        (\textbf{f}-\textbf{g}) Composition of downstream evaluation slides by specimen type (\textbf{f}) and study design (\textbf{g}), shown as percentages of all downstream evaluation slides.
        (\textbf{h}) Sankey diagram showing the distribution of the 72 validation cohorts by task type, specimen type, and evaluation split.
        }
        \label{Fig:main} 
        \end{figure}
        
Despite these methodological advances, prevailing PFMs have been developed under a pancancer paradigm designed for broad transferability, which may be insufficient for GC due to its highly diverse histological patterns and inherent diagnostic ambiguities.
Recent studies indicate that cross-disease pretraining does not consistently translate into strong performance on gastric-specific tasks\cite{wang2025fully,xu2025multimodal,neidlinger2025benchmarking}.
Additionally, because current PFM evaluations remain largely retrospective, prospective clinical robustness remains insufficiently established and real-world utility may be overestimated\cite{ochi2025pathology,cheng2025role}.
To address these limitations, we introduce GRACE, a \textbf{G}astric-specific foundation model for \textbf{R}eal-world \textbf{A}ssessment and \textbf{C}linical d\textbf{E}cision support. 
GRACE was developed using a large-scale, multi-institutional gastric pathology dataset comprising 48,364 predominantly hematoxylin and eosin (H\&E)-stained whole-slide images (WSIs) from 37,493 unique patients (Fig.~\ref{Fig:main}e). 
To capture the distinct morphological features of GC, we designed a gastric-specific continued pretraining strategy. 
Specifically, we adapted Virchow2, which has been validated as a leading pancancer foundation model\cite{ma2025pathbench}, using low-rank adaptation (LoRA) within a DINO\cite{DINO} framework.
The adaptation was performed on a subset of 25,719 biopsy and surgical WSIs, yielding 107 million patches (Fig.~\ref{Fig:main}b). 
Patch embeddings from this adapted backbone were then aggregated via an attention-based multiple instance learning (ABMIL) module\cite{ilse2018attention} to generate patient-level predictions.

We systematically evaluated GRACE across a 28-task benchmark spanning the entire gastric pathology continuum (Fig.~\ref{Fig:main}a), ranging from diagnostic triage to prognostic and molecular predictions.
The downstream evaluation data comprised 22,645 biopsy and surgical WSIs from 11,774 patients (Fig.~\ref{Fig:main}d, f, h). 
Crucially, to overcome the limitations of purely retrospective evaluations, over 35\% of this dataset (7,938 unique WSIs from 5 hospitals) was prospectively collected to support robustness on newly accrued cases (Fig.~\ref{Fig:main}g). 
Across both retrospective and prospective external validation cohorts, GRACE consistently outperformed representative pancancer PFMs, achieving the highest macro-AUC (Fig.~\ref{Fig:performance}b, c).
To translate this discriminative power into decision-relevant workflow signals, we applied fixed retrospective thresholds to prospective cohorts for safety-gated triage. 
By identifying high-confidence prospective subsets with 100\% positive predictive value (PPV) and negative predictive value (NPV), GRACE could achieve substantial workload reductions, triaging up to 69.6\% of cases in biopsy malignancy detection (rule-in and rule-out), 46.8\% in MMR prediction (rule-out), and 28.8\% in Lauren classification (intestinal-type rule-in).    
Finally, we conducted a prospective randomized crossover reader study to quantify GRACE’s impact on human-AI collaboration in clinically realistic workflows (Fig.~\ref{Fig:main}c). 
Compared to unassisted review, GRACE assistance demonstrated concurrent improvements across key workflow endpoints: diagnostic accuracy increased from 82.0\% to 89.9\% (a 43.6\% relative error reduction), geometric-mean interpretation time was shortened by 14.9\%, and diagnostic confidence increased by 9.0\%, alongside improved interobserver agreement. 
These findings substantiate the prospective validity and translational relevance of GRACE, highlighting its substantial potential to streamline and enhance diagnostic workflows.

\begin{figure}
    \captionsetup{labelformat=empty}
    \centering
    \includegraphics[width=0.95\textwidth]{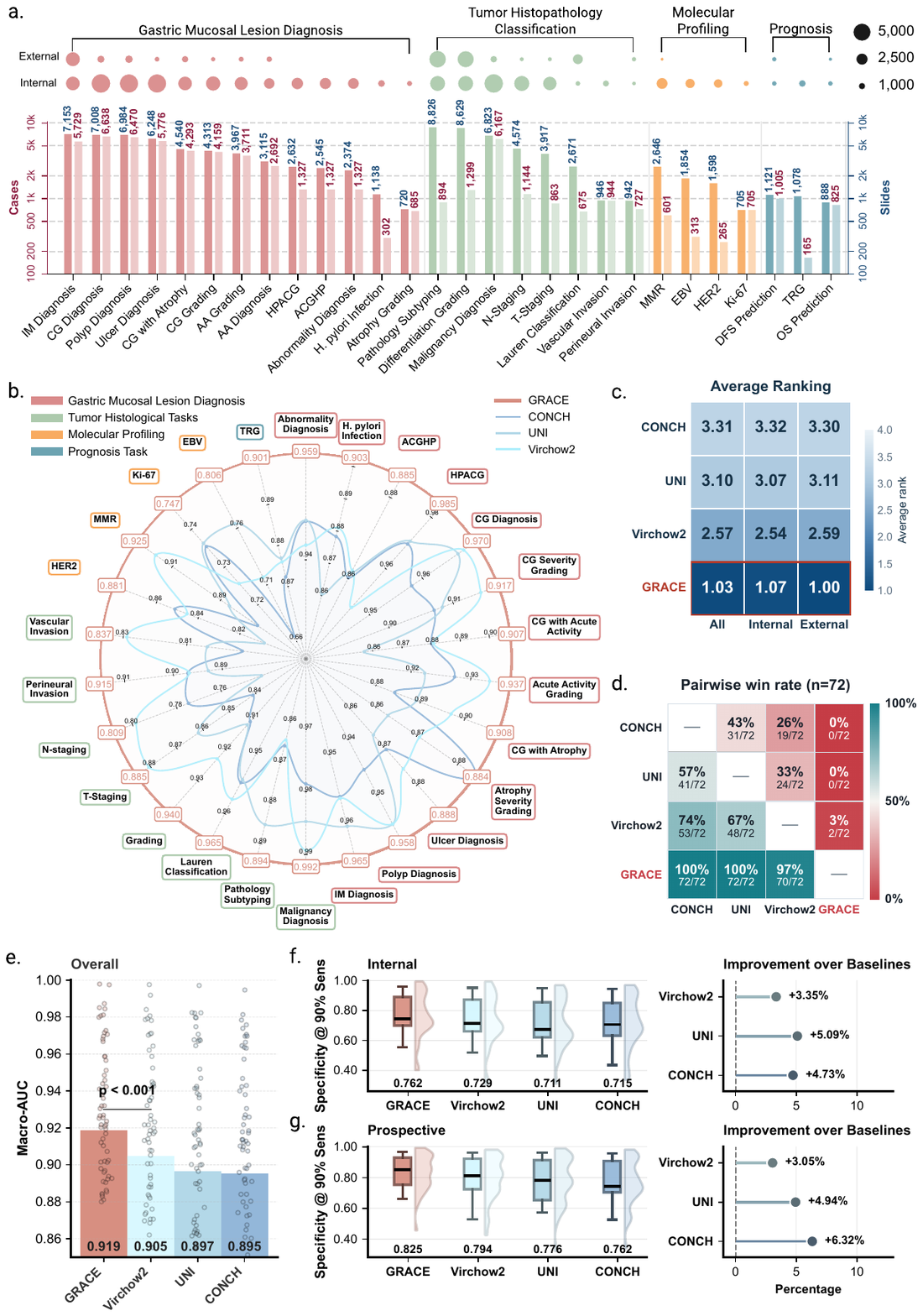}
    \caption{}
    \end{figure}
    \begin{figure}
    \ContinuedFloat
    \caption{%
    \textbf{Overall performance of GRACE.}
    (\textbf{a}) Distribution of slides across the internal and external validation cohorts for all 28 downstream tasks. Bubble area represents the number of slides, while the corresponding case and slide counts for each task are shown in the bar plot. (\textbf{b}) Comparison of task-level macro-AUC between GRACE and pancancer PFMs for 26 histological classification and molecular profiling tasks. Task-level performance was calculated by aggregating all cohorts belonging to each task. 
    (\textbf{c}) Average model rank across all, internal, and external cohorts. Lower rank indicates better performance. 
    (\textbf{d}) Pairwise model dominance across 72 validation cohorts. For each cell, the percentage represents the fraction of cohorts in which the model in the row outperformed the model in the column, and the value below shows the corresponding wins/total cohorts. 
    (\textbf{e}) Comparison of mean macro-AUC across 26 histological classification and molecular profiling tasks. GRACE demonstrated significantly superior performance relative to pancancer PFMs (one-sided Wilcoxon signed-rank test, $P < 0.001$). (\textbf{f}-\textbf{g}) Comparison of specificity at 90\% sensitivity in the internal cohort and prospective external validation cohorts across 26 histological classification and molecular profiling tasks. GRACE's absolute improvements over the baselines are shown on the right.
    }
    \label{Fig:performance}
\end{figure}
\section*{Results}
GRACE is a comprehensive decision-support model evaluated across the routine gastric pathology workflow, encompassing mucosal lesion diagnosis, tumor histology, molecular profiling, and prognosis prediction (Fig.~\ref{Fig:performance}a).
Overall, the model demonstrated robust performance, achieving a mean macro-AUC of 0.9188 across 26 classification tasks (68 cohorts, Fig.~\ref{Fig:performance}e) and an average concordance index (C-index) of 0.7094 for prognosis prediction (4 cohorts).
Compared with pancancer PFMs, GRACE achieved the strongest aggregate benchmark performance (Fig.~\ref{Fig:performance}b).
It secured the best overall average rank (1.03, Friedman test $P < 0.001$), outperforming the second-best-performing model Virchow2 (2.57, Fig.~\ref{Fig:performance}c). GRACE further showed a uniformly leading macro-AUC profile across external validation cohorts (rank 1.00).
Pairwise comparisons across 72 cohorts further confirmed this superiority: GRACE outperformed both CONCH and UNI in all evaluations (100\%), and surpassed Virchow2 in 97\% (70 of 72) of the comparisons (Fig.~\ref{Fig:performance}d). 
Furthermore, at a clinically relevant high-sensitivity operating point (90\%), GRACE maintained robust specificity. 
In both retrospective and prospective evaluations, it achieved the highest average specificity, surpassing Virchow2 by over 3 percentage points ($P < 0.05$, Fig.~\ref{Fig:performance}f, g).

\subsection*{Accurate diagnosis of gastric mucosal lesions}

\begin{figure}
\captionsetup{labelformat=empty}
\centering
\includegraphics[width=\textwidth]{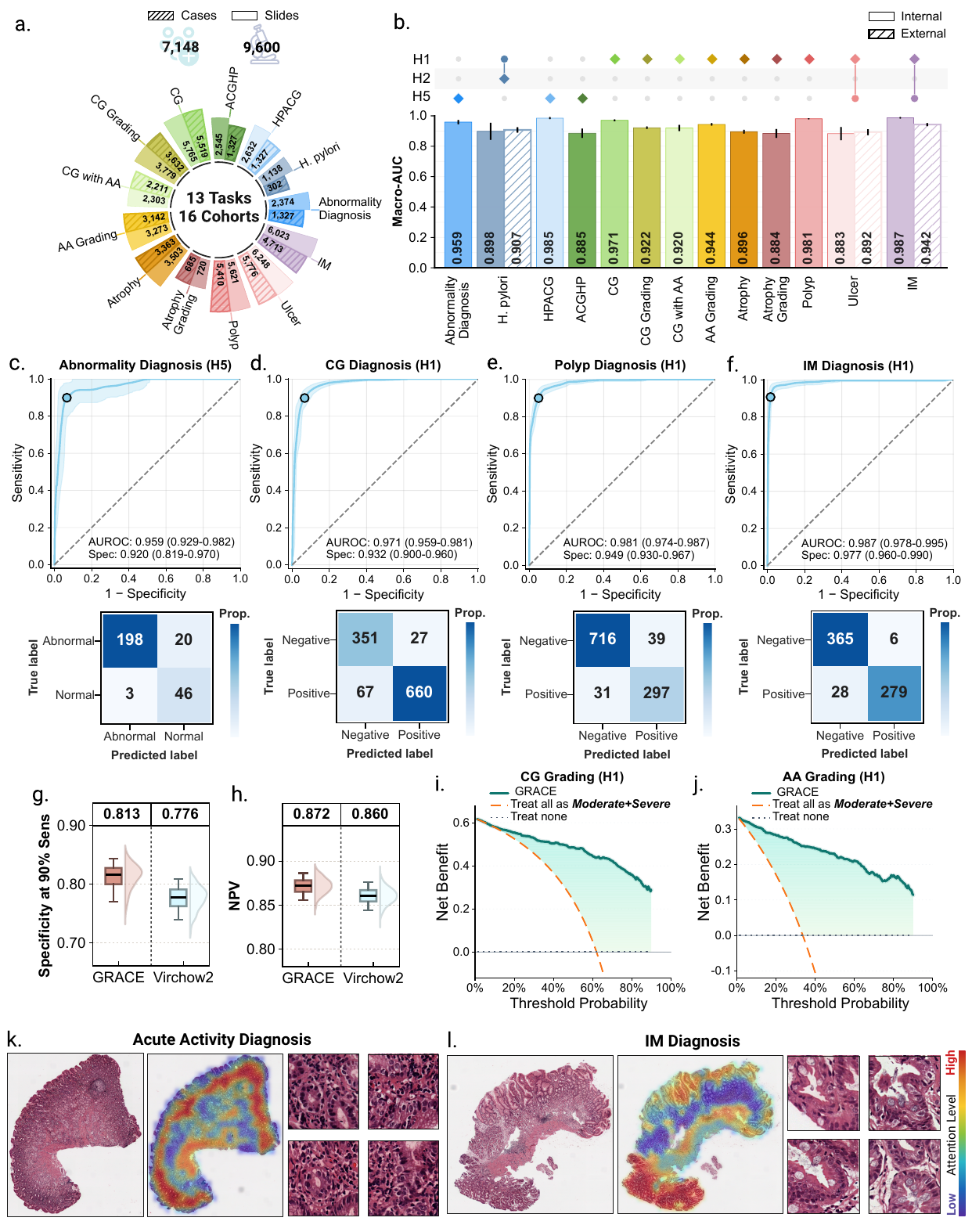}
\caption{}
\end{figure}
\begin{figure}
    \ContinuedFloat
    \caption{%
    \textbf{Performance of GRACE on gastric mucosal lesion diagnosis.}
    (\textbf{a}) Distribution of patients and slides across gastric mucosal lesion diagnosis.
    (\textbf{b}) Cohort-level macro-AUC of GRACE. Bars represent bootstrap mean estimates with error bars showing mean $\pm$ standard deviation (SD). Unless otherwise specified, cohort composition for each task is presented at top with diamonds marking internal cohorts and circles marking external cohorts.
    (\textbf{c}-\textbf{f}) Receiver operating characteristic (ROC) curves with AUROC (95\% CI) and ROC-derived specificity at 90\% sensitivity (95\% CI). Confusion matrices show empirical counts at a discrete operating threshold selected to achieve at least 90\% sensitivity. ROC curves for the remaining tasks are shown in Extended Data Figs.~\ref{fig:AUROC-1} and \ref{fig:AUROC-2}. 
    (\textbf{g}-\textbf{h}) Comparison of specificity at 90\% sensitivity (\textbf{g}) and NPV (\textbf{h}) between GRACE and Virchow2 across the 16 cohorts. Boxplots with half-violin overlays show the distribution of metric values.
    (\textbf{i}-\textbf{j}) DCA for CG grading and acute activity grading. GRACE showed higher net benefit than treating all cases as moderate/severe grades, or treating none.
    (\textbf{k}-\textbf{l}) Attention heatmaps with top-ranked patches, highlighting diagnostically relevant regions driving model predictions. 
    }
    \label{fig:precancerous}
\end{figure}

GC prevention relies on accurate risk stratification of mucosal lesions along the Correa cascade\cite{toh_pathways_2020}.
To evaluate whether GRACE could support clinically relevant diagnosis of gastric mucosal lesions, we established a benchmark using 9,600 WSIs from 7,148 patients across 13 tasks (16 validation cohorts, Fig.~\ref{fig:precancerous}a).
GRACE demonstrated robust discrimination performance (mean macro-AUC 0.9286, Fig.~\ref{fig:precancerous}b) and a strong average NPV (0.8717, Fig.~\ref{fig:precancerous}h). 
Notably, GRACE significantly reduced the false-positive rate compared to Virchow2 (decreased by 3.67 percentage points at 90\% sensitivity, $P < 0.05$, Fig.~\ref{fig:precancerous}g).

\textbf{Screening and etiological triage.} 
For front-line biopsy triage, GRACE distinguished abnormal from normal gastric tissue with a strong balance between sensitivity and false-positive burden (AUROC 0.9589, specificity 92.04\% at 90\% sensitivity, Fig.~\ref{fig:precancerous}c). 
When evaluated for etiological stratification, GRACE exhibited robust \textit{H. pylori} detection performance, yielding AUROCs of 0.8980 in the internal cohort and 0.9072 in the external cohort.
Beyond direct \textit{H. pylori} detection, GRACE successfully captured the broader inflammatory context of infection-related gastritis.
The model demonstrated strong discrimination for \textit{H. pylori}-associated chronic gastritis (HPACG), maintaining a low false-negative rate even under a conservative specificity constraint (AUROC 0.9851, sensitivity 96.39\% at 90\% specificity, NPV 0.9473). 
Meanwhile, it retained clinically useful exclusion performance for autoimmune chronic gastritis with concurrent \textit{H. pylori} infection (ACGHP), yielding an AUROC of 0.8854 and an NPV of 0.9161.

\textbf{Diagnosis and grading of gastric inflammation and atrophy.}
GRACE captured diverse histological patterns of gastric mucosal disease, spanning CG, acute inflammatory activity, and atrophic damage. 
For CG diagnosis, the model not only achieved high diagnostic precision (AUROC 0.9707, 93.26\% sensitivity at 90\% specificity, PPV 0.9427, Fig.~\ref{fig:precancerous}d) but also reliably graded inflammatory severity (macro-AUC 0.9218).
This stratification yielded higher net benefit than non-stratified reference strategies, demonstrating its potential utility for risk-guided pathological review (Fig.~\ref{fig:precancerous}i).
Similarly, GRACE effectively detected acute inflammatory activity (AUROC 0.9203, PPV 0.9472) and classified its intensity into three grades (macro-AUC 0.9439).
Decision-curve analysis (DCA) highlighted the potential clinical value of this grading, showing higher net benefit than standard reference strategies  (Fig.~\ref{fig:precancerous}j).
Crucially, attention heatmaps confirmed that these predictions were morphology-grounded, with high-scoring patches accurately localized to regions of active inflammation (Fig.~\ref{fig:precancerous}k).
Furthermore, as structural mucosal changes can arise from persistent inflammation, GRACE demonstrated favorable performance in identifying gastric atrophy (AUROC 0.8960, PPV 0.8761) and grading its severity (AUROC 0.8844).

\textbf{Detection of cancer-relevant mucosal abnormalities.}
Extending beyond inflammatory and atrophic gastric injury, GRACE effectively identified cancer-relevant mucosal changes. 
It showed reliable rule-in potential for gastric polyps (AUROC 0.9807, PPV 0.9515, specificity 0.9491 at 90\% sensitivity, Fig.~\ref{fig:precancerous}e) and accurate ulcer detection (AUROCs $>$ 0.8828). 
For IM, GRACE achieved near-ceiling internal discrimination (AUROC 0.9873, Fig.~\ref{fig:precancerous}f) and generalized well to the external cohort (AUROC 0.9423), with predictions accurately mapping to morphologically relevant regions (Fig.~\ref{fig:precancerous}l).

Together, GRACE can maintain high case capture while limiting false-positive burden.
In a setting where gastric lesion assessment is subject to interobserver variation\cite{leja2013interobserver}, GRACE’s consistent support at subtle diagnostic boundaries positions it as a scalable framework for more standardized pathological review.

\subsection*{Robust clinicopathological characterization of gastric cancer}
Beyond early mucosal lesion detection,  definitive pathological evaluation of GC demands a multi-dimensional analysis to guide risk stratification and therapeutic intervention.
We evaluated GRACE across this workflow using 11,457 slides from 7,457 patients across 9 tumor histology tasks (14 cohorts, Fig.~\ref{fig:malignant}a). 
GRACE outperformed all pancancer PFM baselines, achieving an average macro-AUC of 0.9053 (Fig.~\ref{fig:malignant}b) and a 3.44-percentage-point macro-ACC improvement over the second-best model, CONCH.

\textbf{Malignancy detection.} 
GRACE accurately discriminated malignant from non-malignant cases, yielding an AUROC of 0.9979 and maintaining a specificity of 0.9946 at a 90\% sensitivity threshold (Fig.~\ref{fig:malignant}c). 
The model demonstrated strong predictive reliability (PPV 100\%, NPV 99.73\%, Fig.~\ref{fig:malignant}h, i), supporting accurate case escalation while minimizing false-positive referrals.
Complementing these quantitative results, attention heatmaps highlighted morphologically suspicious regions, suggesting that malignancy classification was anchored in pathologically relevant tissue areas (Fig.~\ref{fig:malignant}p).

\textbf{Histologic tumor characterization.}
GRACE provided robust Signet-ring cell carcinoma (SRC)-focused subtype stratification, distinguishing SRC, tubular adenocarcinoma, and non-specified adenocarcinoma with strong multiclass performance (macro-AUC 0.9019) while maintaining useful exclusion of SRC among model-negative cases (SRC NPV 0.9229). 
GRACE also achieved favorable performance in Lauren classification with promising rule-out capacity for intestinal-type disease (AUROC 0.9566, intestinal-type NPV 0.9698), potentially informing prognostic interpretation\cite{burgart2022protocol, mohri2010prognostic} (Fig.~\ref{fig:malignant}d).
Beyond subtype assignment, the model consistently differentiated poorly differentiated (G3) from well-to-moderately differentiated (G1-G2) tumors, generalizing from internal validation (AUROC 0.9585) to external cohorts (H3 AUROC 0.9320, H4 AUROC 0.9255, Fig.~\ref{fig:malignant}e).

\textbf{Tumor aggressiveness and treatment-response assessment.}
GRACE further captured local invasion depth across T1-T4 with generalizable multiclass discrimination (internal macro-AUC 0.8927, external macro-AUC 0.8800). 
Extending this assessment from local tumor extent to regional spread, GRACE successfully extracted nodal-risk signals from primary-tumor WSIs. 
It distinguished N0 from N+ status and outperformed all pancancer baselines (internal AUROC 0.8381, external AUROC 0.8009, all $P < 0.05$), suggesting a complementary role when conventional nodal assessment is constrained by lymph-node yield or examination rigor\cite{chen2021recent}.
In invasion-related assessment, GRACE demonstrated robust performance in detecting adverse invasion features and showed reliable positive calls for perineural invasion (PNI, internal AUROC 0.9381 and PPV 0.9339, external AUROC 0.9141, Fig.~\ref{fig:malignant}f). 
GRACE additionally provided informative classification of vascular invasion (VI) with an AUROC of 0.8371. Finally, GRACE extended this aggressiveness assessment to treatment response, quantifying tumor regression grade (TRG) with an AUROC of 0.9013.

Collectively, GRACE enables accurate, efficient identification of key tumor characteristics that inform subtyping, nodal metastatic risk, and treatment response. By translating routine WSIs into clinically structured signals, GRACE may provide a foundation for more actionable postoperative interpretation and downstream clinical decision-making.

\begin{figure}
    \captionsetup{labelformat=empty}
    \centering
    \includegraphics[width=0.95\textwidth]{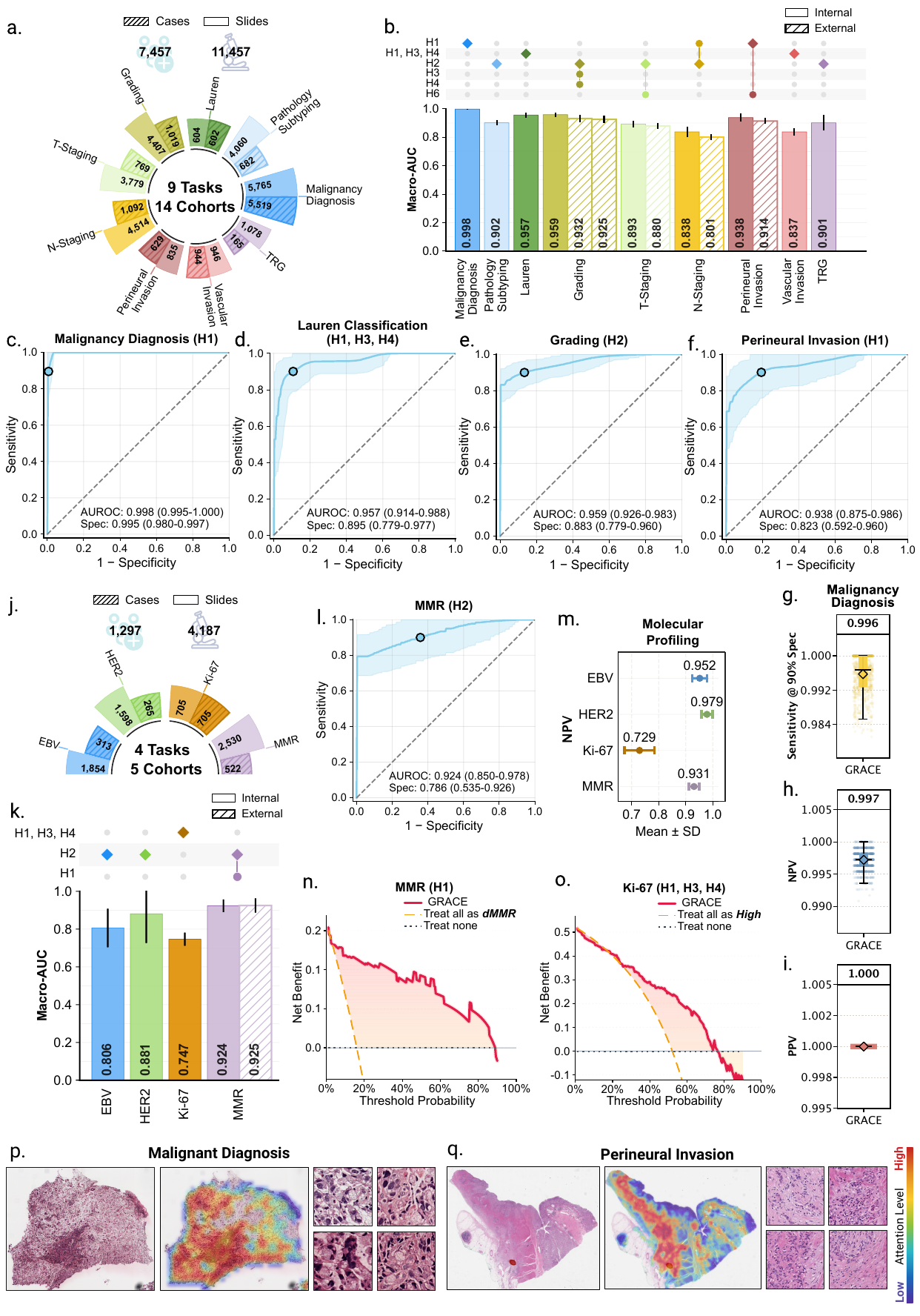}
    \caption{}
\end{figure}
\begin{figure}
    \ContinuedFloat
    \caption{%
    \textbf{Performance of GRACE on tumor histological classification and molecular profiling tasks.}
        (\textbf{a}, \textbf{j}) Distribution of patients and slides across tumor histological classification (\textbf{a}) and molecular profiling (\textbf{j}) tasks.
        (\textbf{b}, \textbf{k}) Cohort-level macro-AUC performance of GRACE on the tumor histological classification (\textbf{b}) and molecular profiling (\textbf{k}) tasks. Bars represent bootstrap mean estimates, and error bars indicate SD.
        (\textbf{c}-\textbf{f}, \textbf{l}) ROC curves with AUROC (95\% CI) and specificity at 90\% sensitivity (95\% CI). "H1, H3, H4" indicates that data from Hospital H1, H3, and H4 were combined into one evaluation cohort to increase the effective sample size, unless otherwise specified. ROC curves for the remaining tasks are shown in Extended Data Figs.~\ref{fig:AUROC-2}, ~\ref{fig:AUROC-3} and \ref{fig:AUROC-4}. 
        (\textbf{g}-\textbf{i}) Malignancy diagnosis performance in the internal cohort, shown by sensitivity at 90\% specificity, NPV, and PPV. Unless otherwise specified, jittered points show bootstrap estimates; diamonds, horizontal ticks, thick colored bars, and capped black lines denote bootstrap means, medians, interquartile ranges, and 95\% CIs, respectively.
        (\textbf{m}) Mean NPV for each molecular profiling task, averaged across internal and retrospective external validation cohorts. 
        (\textbf{n}-\textbf{o}) DCA for MMR and Ki-67 predictions showed that GRACE provided higher net benefit across clinically relevant threshold probabilities over treat-all and treat-none strategies.
        (\textbf{p}-\textbf{q}) WSI attention heatmaps with top-ranked patches illustrating regions contributing most to predictions. More visualization results are reported in Extended Data Fig.~\ref{fig:heatmap}.}
        \label{fig:malignant} 
        \end{figure}

\subsection*{Molecular profiling from H\&E slides}
Routine molecular and immunohistochemistry (IHC) assays in GC inform treatment selection and prognostic assessment, but remain constrained by tissue consumption, turnaround time, and labor-intensive workflows\cite{kock_am_brink_intratumoral_2023,lee_spatial_2025}.
To evaluate the potential of GRACE as a rapid, sample-sparing prescreening tool, we tested its ability to predict several clinically relevant molecular biomarker states, including HER2 status, mismatch repair (MMR) status, Ki-67 expression, and EBV status, directly from routine H\&E slides. 
Leveraging 4,187 surgical resection slides from 1,297 patients (Fig.~\ref{fig:malignant}j), GRACE achieved a mean AUROC of 0.8565 in 5 retrospective cohorts (Fig.~\ref{fig:malignant}k).
For actionable therapeutic targets such as HER2\cite{ajani_gastric_2025,alsina_current_2023,joshi2021current}, GRACE demonstrated strong discrimination (macro-AUC 0.8811, NPV 0.9785, Fig.~\ref{fig:malignant}m). 
This high negative-call reliability extended to MMR status prediction. 
GRACE successfully distinguished microsatellite instability-high/deficient mismatch repair (MSI-H/dMMR) from microsatellite stable/proficient mismatch repair (MSS/pMMR) in both internal and retrospective external cohorts (AUROCs of 0.9236 and 0.9252, respectively), establishing a strong capacity to exclude MSI-H/dMMR tumors (average NPV 0.9306, Fig.~\ref{fig:malignant}l).
Beyond therapy-relevant biomarkers, GRACE also stratified Ki-67 expression into low ($\leq 10\%$), intermediate ($10\% < x \leq 50\%$), and high ($>50\%$) positive-cell categories\cite{xiong_ki-67mki67_2019}, achieving a macro-AUC of 0.7467, and predicted EBV status with high negative-call reliability (NPV 0.9518). 
Across the evaluated molecular endpoints, GRACE significantly surpassed pancancer baselines ($P < 0.05$), while DCA underscored its potential clinical utility in MMR and Ki-67 prediction tasks (Fig.~\ref{fig:malignant}n, o). 
Together, these findings support the integration of GRACE as a high-confidence, exclusion-oriented molecular prescreening tool for rapid, tissue-sparing precision medicine in GC.

\subsection*{H\&E morphology-derived survival prediction and risk stratification}
To assess the potential utility of GRACE for postoperative risk stratification, we evaluated its prognostic performance on surgically resected WSIs.
The model robustly predicted disease-free survival (DFS), achieving a C-index of 0.7329 in the internal cohort and retaining a C-index of 0.7032 in external validation. 
Similar generalization was observed for overall survival (OS), with C-indices of 0.6869 internally and 0.7144 externally.
Kaplan-Meier analyses further confirmed that GRACE stratified patients into distinct high- and low-risk groups for both DFS and OS (log-rank test, $P < 0.0001$, Extended Data Fig.~\ref{fig:KM}a-d).
The results demonstrate that GRACE could extract clinically meaningful prognostic signals directly from routine postoperative WSIs, enabling an initial stratification of patients by their predicted recurrence and survival risks. 
By providing risk estimates at the time of surgical pathology review, GRACE may complement established clinicopathologic factors to identify patients who warrant closer multidisciplinary review or follow-up.

    \begin{figure}
    \captionsetup{labelformat=empty}
    \centering
    \includegraphics[width=\textwidth]{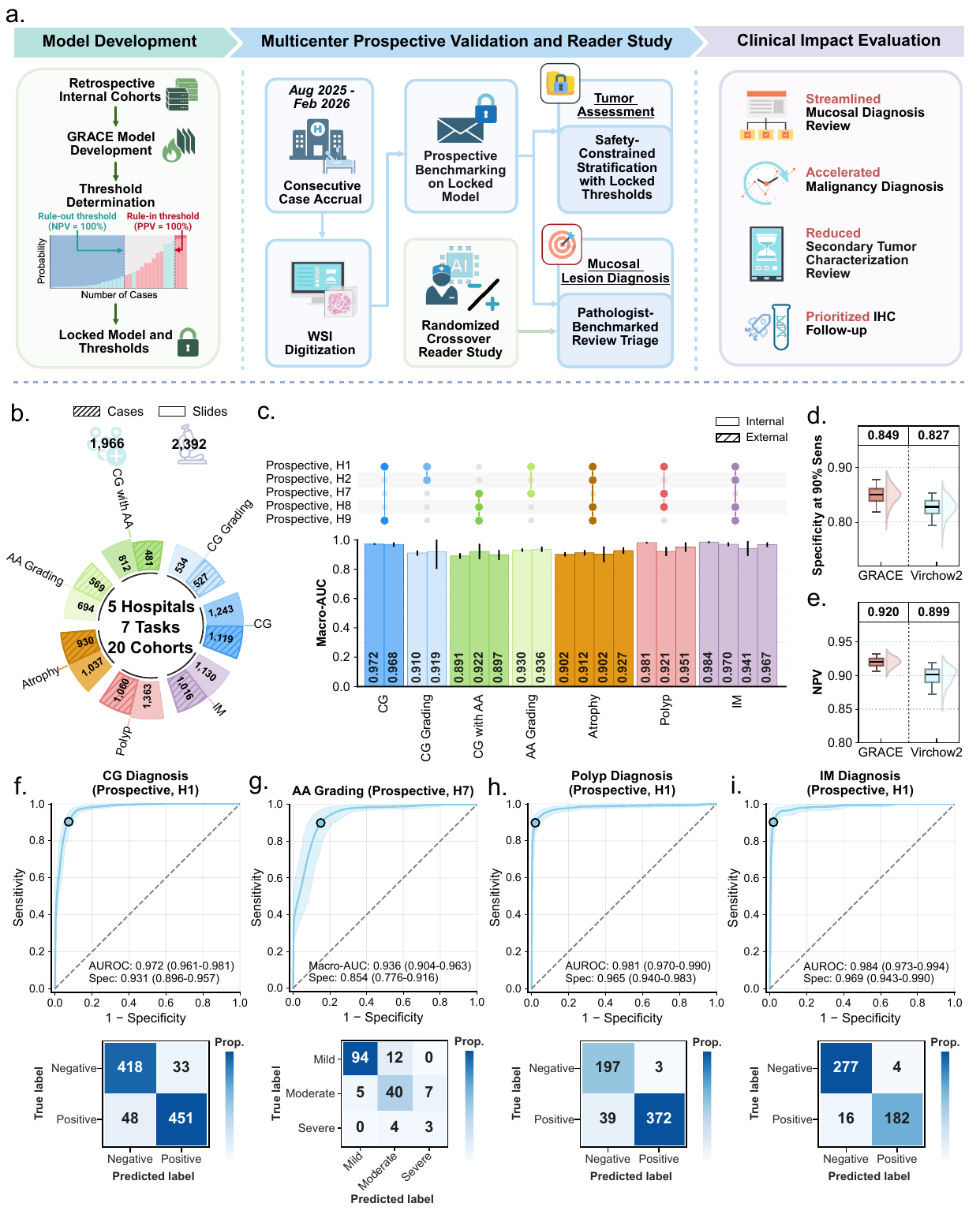}
    \caption{}
\end{figure}
\begin{figure}
    \ContinuedFloat
    \caption{%
    \textbf{Prospective validation of GRACE on gastric mucosal lesion diagnosis.}
    (\textbf{a}) Overview of the clinical utility evaluation. Multicenter consecutive cases underwent prospective validation using the locked model, followed by prospective safety-constrained tumor assessment with retrospectively locked rule-in/rule-out thresholds. Pathologist-referenced review triage for mucosal lesion diagnosis was evaluated against reader study benchmarks.
    (\textbf{b}) Distribution of patients and slides across prospective validation of gastric mucosal lesion tasks.
    (\textbf{c}) Overall macro-AUC performance of GRACE. Bars represent bootstrap mean estimates, and error bars indicate SD.
    (\textbf{d}-\textbf{e}) Comparisons of specificity at 90\% sensitivity (\textbf{d}) and NPV (\textbf{e}) between GRACE and Virchow2 across the 20 cohorts. Boxplots with half-violin overlays show the distribution of metric values.
    (\textbf{f}-\textbf{i}) ROC curves with AUROC (95\% CI) and ROC-derived specificity at 90\% sensitivity (95\% CI). Confusion matrices show empirical counts at a discrete operating threshold selected to achieve at least 90\% sensitivity. For the AA grading confusion matrix, moderate/severe cases are treated as positive. ROC curves for other tasks are included in Extended Data Figs.~\ref{fig:AUROC-1} and \ref{fig:AUROC-2}.
    }
    \label{fig:prospective}
\end{figure}

\subsection*{Prospective real-world validation and triage utility}
For AI systems intended for clinical pathology, prospective evaluation is essential to determine whether strong retrospective performance translates into safe, workflow-compatible diagnostic support under real-world case-mix diversity. 
Accordingly, we conducted a multicenter prospective observational evaluation using consecutive routine biopsy and surgical cases from 5 hospitals, spanning gastric mucosal disease (2,392 slides from 1,966 patients, Fig.~\ref{fig:prospective}b) and tumor histological and molecular prediction (6,264 slides from 1,039 patients, Fig.~\ref{fig:prospective_malignant}a).

\subsubsection*{Endpoint-aware triage in gastric mucosal disease}
GRACE generalized robustly to prospective gastric mucosal assessment, maintaining strong discrimination and clinically useful rule-out performance across 7 gastric mucosal lesion endpoints, achieving a mean macro-AUC of 0.9352 (20 cohorts, Fig.~\ref{fig:prospective}c). 
At a strict 90\% sensitivity operating point, GRACE retained a mean specificity of 0.8492 and a mean NPV of 0.9201, with specificity exceeding Virchow2 by 2.25 percentage points ($P < 0.05$, Fig.~\ref{fig:prospective}d, e). 

For CG diagnosis, GRACE achieved a mean AUROC of 0.9698, with a high specificity of 0.9378 at 90\% sensitivity and a PPV of 0.9303 (Fig.~\ref{fig:prospective}f), indicating limited false-positive calls and reliable positive predictions.
This strong positive-call performance extended to CG severity grading, where GRACE identified moderate/severe CG with high precision (mean PPV 0.9653) and achieved a three-class mean macro-AUC of 0.9149.
Conversely, for acute inflammatory activity and chronic atrophic gastritis diagnoses, GRACE demonstrated strong rule-out capabilities with very few missed cases among negative predictions (acute activity: mean AUROC 0.9031, mean NPV 0.9910; atrophy: mean AUROC 0.9107, mean NPV 0.9395). Furthermore, it maintained strong multiclass stratification for grading acute activity severity (mean macro-AUC 0.9332, Fig.~\ref{fig:prospective}g).
For lesion-level and premalignant findings, GRACE provided positive-call utility with false-positive control, most prominently for gastric polyp recognition and also for IM classification at the 90\% sensitivity operating point (polyp: PPV 0.9968, specificity 0.9649 at H1; IM: PPV 0.9265, specificity 0.9690 at H1; Fig.~\ref{fig:prospective}h, i).
Overall, GRACE demonstrated adaptive diagnostic capabilities across different clinical endpoints.
Its reliable positive predictions for CG, gastric polyps, and IM support its use in early lesion stratification and the prioritization of high-risk cases. 
Simultaneously, its robust negative predictions effectively rule out acute inflammatory activity and atrophy with minimal missed cases.
This prospective pattern highlights the potential of GRACE as an endpoint-aware triage aid that can help focus assessment on diagnostically ambiguous or clinically relevant cases in high-volume gastric biopsy practice.

\subsubsection*{Robust generalization across gastric tumor histological and biomarker endpoints}
Beyond inflammatory mucosal disease, GRACE demonstrated robust prospective generalization across tumor histological classification and biomarker prediction tasks.
Across 8 endpoints (13 cohorts), GRACE reached a mean macro-AUC of 0.9201 (Fig.~\ref{fig:prospective_malignant}b). 
Under the fixed 90\% sensitivity criterion, GRACE showed higher specificity than baseline models, with a 4.27-percentage-point gain over the second-best model ($P < 0.05$).

\begin{figure}
    \captionsetup{labelformat=empty}
    \centering
    \includegraphics[width=0.95\textwidth]{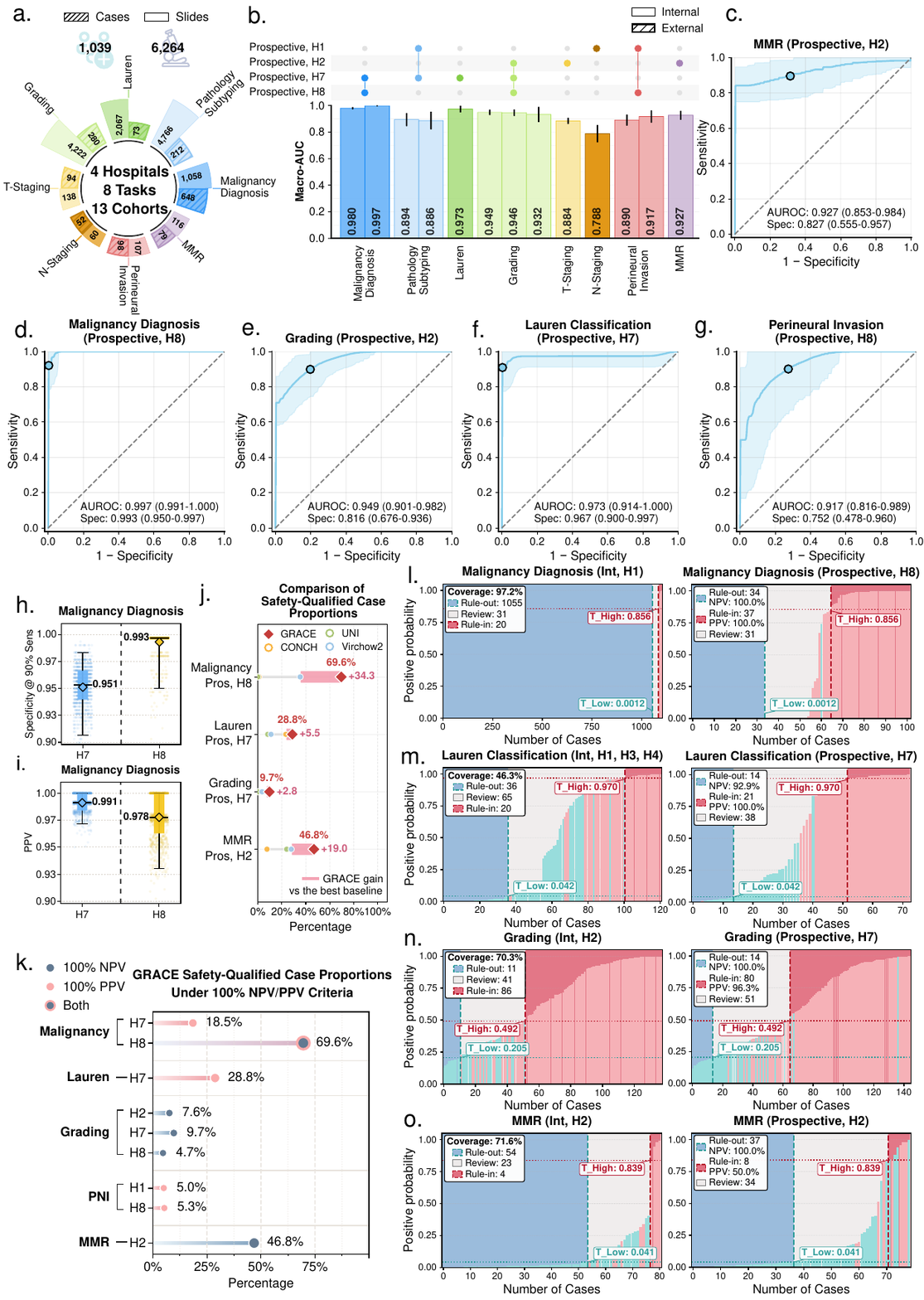}
    \caption{}
    \end{figure}
    \begin{figure}
    \ContinuedFloat
    \caption{%
    \textbf{Prospective validation of GRACE on tumor histological classification and molecular profiling tasks.}
    (\textbf{a}) Distribution of patients and slides across prospective validation of the tasks.
    (\textbf{b}) Cohort-level macro-AUC performance of GRACE. Bars represent bootstrap mean estimates, and error bars indicate SD.
    (\textbf{c}-\textbf{g}) ROC curves with AUROC (95\% CI) and specificity at 90\% sensitivity (95\% CI). ROC curves for the remaining tasks are included in Extended Data Figs.~\ref{fig:AUROC-2}, \ref{fig:AUROC-3} and \ref{fig:AUROC-4}. 
    (\textbf{h}-\textbf{i}) Specificity at 90\% sensitivity (\textbf{h}) and PPV (\textbf{i}) for prospective malignancy diagnosis in Hospitals H7 and H8. 
    (\textbf{j}) Comparison of safety-qualified case proportions for the prospective cohorts shown in (\textbf{l}-\textbf{o}). Cases were counted as safety-qualified only when the prospective rule-out subset maintained 100\% NPV and/or the rule-in subset maintained 100\% PPV.
    (\textbf{k}) GRACE safety-qualified case proportions across cohorts, annotated by whether the 100\% NPV and/or 100\% PPV criteria were met.
    (\textbf{l}-\textbf{o}) Probability-ranked plots of prospective cases (right) under thresholds locked from the corresponding internal retrospective cohort (left). Bars represent individual cases, with height indicating predicted positive probability and color indicating ground-truth class. Shaded regions mark the rule-out, manual review, and rule-in regions. The legend reports the number of prospective cases assigned to the high-confidence regions and their observed predictive values.
    }
    \label{fig:prospective_malignant}
\end{figure}
In malignancy diagnosis, GRACE achieved consistently strong cancer-identification performance with low false-positive burden across cohorts (mean AUROC 0.9887, mean specificity 0.9720 at 90\% sensitivity, mean PPV 0.9844, Fig.~\ref{fig:prospective_malignant}h, i). 
This performance was particularly evident in the H8 cohort, where GRACE achieved near-ceiling discrimination and high specificity at 90\% sensitivity (AUROC 0.9975, specificity 0.9930, Fig.~\ref{fig:prospective_malignant}d).  

Beyond malignancy detection, GRACE excelled in prospective histological characterization.
It outperformed Virchow2 in pathology subtyping (mean macro-AUC 0.8900, +4.35 percentage points), and maintained high predictive power for Lauren classification (AUROC 0.9727, Fig.~\ref{fig:prospective_malignant}f) and tumor differentiation grading (mean AUROC 0.9421, Fig.~\ref{fig:prospective_malignant}e).
Across staging-related endpoints, GRACE delivered clinically meaningful negative predictions for advanced local invasion and nodal involvement, identifying patients unlikely to have T3/T4 (NPV 0.9753) or N+ stages (NPV 0.8671).
Furthermore, GRACE generalized prospectively to PNI classification (mean AUROC 0.9037, Fig.~\ref{fig:prospective_malignant}g).
Extending beyond histological classification, GRACE further achieved strong prospective dMMR/MSI-H prediction, combining robust discrimination with high-confidence negative calls and few missed dMMR/MSI-H cases (AUROC 0.9265, NPV 0.9693, Fig.~\ref{fig:prospective_malignant}c).

\subsubsection*{Safety-constrained tumor assessment using retrospectively locked thresholds}

Reducing pathology turnaround time holds clinical significance for timely patient management. To explore whether GRACE could support clinically meaningful prospective workflow optimization, we performed a simulated safety-gated triage analysis for malignancy detection, postoperative tumor characterization, and biomarker prediction (Fig.~\ref{fig:prospective}a). Retrospectively locked thresholds were applied to prospective cohorts, with safety-qualified subsets defined as those maintaining 100\% NPV for rule-out or 100\% PPV for rule-in.

First, GRACE could accelerate malignancy diagnosis by automating triage while reserving uncertain biopsies for pathologist review.
In the prospective H8 cohort, 69.6\% of cases fell into safety-qualified regions, indicating a possible reduction in immediate full-review workload (Fig.~\ref{fig:prospective_malignant}j, k, l).
In H7, 82.2\% of cases were assigned to low- and high-probability regions using locked retrospective thresholds, but safety-qualified acceleration was limited to the rule-in subset (100\% PPV), corresponding to a possible 18.5\% review reduction (Fig.~\ref{fig:prospective_malignant}k). 
Although the low-probability region was not used for direct rule-out triage, its high NPV of 98.6\% (Extended Data Table~\ref{tab:lock_t}) suggests value as a secondary prioritization signal, helping pathologists prioritize higher-risk or uncertain biopsies for review.

Second, for postoperative tumor characterization, GRACE may reduce secondary-review workload by separating high-confidence predictions.
This was most pronounced for Lauren classification in H7, where 28.8\% of cases achieved 100\% intestinal-type PPV, qualifying for reduced review (Fig.~\ref{fig:prospective_malignant}j, k, m). 
Safety-qualified fractions were smaller for histological grading (9.7\% at H7) and PNI (5.3\% at H8), a finding consistent with the focal heterogeneity of these features (Fig.~\ref{fig:prospective_malignant}k). However, GRACE still provided valuable prioritization signals, achieving 96.3\% PPV for G3 rule-in (Fig.~\ref{fig:prospective_malignant}n) and 92.9\% NPV for intestinal-type rule-out, thereby directing pathologists toward ambiguous cases requiring closer review.

Third, GRACE could offer a practical approach to prioritize MMR-IHC follow-up by identifying cases that require immediate confirmation while moving low-risk cases out of the urgent testing queue.
It identified a 100\% NPV rule-out subset comprising 46.8\% of the prospective cohort (Fig.~\ref{fig:prospective_malignant}j, k, o). 
Routing nearly half of all surgical cases to deferred or batched confirmation, without missing MMR-deficient tumors in this cohort, could effectively reduce urgent IHC workload and concentrate rapid reflex testing on high-probability cases.

Together, these prospective results support the feasibility of evaluating GRACE as a safety-constrained triage tool for task-specific workflow optimization in gastric cancer pathology. 
Under strict safety constraints, GRACE may reduce the malignancy review burden by up to 69.6\%, decrease secondary reviews for intestinal-type Lauren rule-in by 28.8\%, and defer 46.8\% of urgent MMR-IHC tests. More broadly, this framework illustrates a clinically grounded role for AI in pathology, reducing routine workload while preserving expert judgment for cases marked by uncertainty, urgency, or major treatment implications.

\subsection*{Randomized crossover reader study for pathologist-AI collaboration}
To obtain direct evidence of clinical translation in a diagnostic workflow, we conducted a prospective randomized crossover reader study with 8 practicing pathologists. The reader study focused on three biopsy-based mucosal tasks, including acute inflammatory activity grading, chronic atrophic gastritis diagnosis, and IM diagnosis. The enrollment and analysis flow followed CONSORT guidance\cite{turner2012consolidated} (Extended Data Figs.~\ref{fig:consort}a and Extended Data Table~\ref{tab:RCT_data}).
Participants were stratified by experience level and randomized to one of two crossover sequences (without AI then with AI, or vice versa), separated by a four-week washout period (Fig.~\ref{fig:RCT_1}a).
This design enabled a direct assessment of GRACE's impact on diagnostic accuracy, efficiency, confidence, and inter-reader consistency (Extended Data Table~\ref{tab:overall_ai_vs_noai}).

\paragraph{Balanced diagnostic reliability in AI-assisted reading.}
GRACE enhanced diagnostic reliability across tasks and reader experience levels, increasing overall accuracy while reducing missed positive cases and false-positive interpretations.
AI assistance increased overall accuracy from 82.0\% to 89.9\%, corresponding to a 7.8 percentage-point absolute gain and an approximately 43.6\% relative reduction in diagnostic error ($P < 0.05$, Fig.~\ref{fig:RCT_1}b). Multivariable generalized estimating equation (GEE) models, adjusting for pathologist experience, crossover sequence, and diagnostic task, confirmed that AI assistance was associated with nearly doubled odds of a correct diagnosis (adjusted odds ratio (OR) 1.987, $P < 0.05$, Extended Data Table~\ref{tab:gee_bootstrap_all}).
The improvement in accuracy was observed across all three diagnostic tasks, with the largest absolute gain in acute activity severity grading (+9.1 percentage points, Extended Data Table~\ref{tab:performance}; adjusted OR 2.086, $P < 0.05$), followed by diagnosis of chronic atrophic gastritis (+8.5 percentage points, adjusted OR 1.841, $P < 0.05$). 
In addition, the benefit was particularly pronounced among junior pathologists, whose accuracy increased by 10.3 percentage points compared with their performance without GRACE assistance (adjusted OR 2.010, $P < 0.05$). 
Crucially, these gains were not accompanied by an apparent trade-off toward either overcalling or undercalling.
AI assistance concurrently improved sensitivity (80.9\% to 88.3\%) and specificity (85.7\% to 92.6\%, both $P < 0.05$), indicating improved positive-case detection and false-positive avoidance.
Junior pathologists showed particularly large improvements, with sensitivity and specificity increasing by 9.5 and 8.8 percentage points, respectively (adjusted OR 1.846 and 2.100, respectively, both $P < 0.05$). 
Among individual tasks, sensitivity improved most for chronic atrophic gastritis diagnosis (+10.2 percentage points, adjusted OR 2.017, $P < 0.05$).

\paragraph{Diagnostic confidence and review efficiency.} 
GRACE assistance improved pathologists' diagnostic confidence for gastric mucosal assessments. 
On the 10-point confidence scale, mean diagnostic confidence increased from 8.5 without AI to 9.2 with AI assistance, corresponding to a 9.0\% relative increase ($P < 0.05$, Fig.~\ref{fig:RCT_1}c, Extended Data Table~\ref{tab:confidence}). Overall, pathologists' confidence increased in 55.5\% of reader assessments (Extended Data Fig.~\ref{fig:AI_impact_cat_line}a). 
The confidence gain was most pronounced among junior pathologists, who showed an adjusted increase of +1.184 points ($P < 0.05$). Task-by-experience analysis further showed consistently larger median confidence improvements for junior than senior readers across diagnostic tasks (Extended Data Fig.~\ref{fig:AI_impact_cat_line}c), suggesting that GRACE is particularly valuable for supporting less-experienced interpretation while continuing to benefit expert practice.
Beyond improving diagnostic performance, GRACE accelerated interpretation, significantly reducing diagnostic time across all tasks and experience levels compared with unaided review (Extended Data Table~\ref{tab:time}). Geometric mean diagnostic time decreased from 45.0 seconds without AI to 38.3 seconds with AI, and 81.9\% of reader assessments were completed faster under model-supported review (Extended Data Fig.~\ref{fig:AI_impact_cat_line}b). 
Consistently, paired comparisons of reading time for the same cases showed a tendency toward shorter AI-assisted reading times for both junior and senior pathologists, with most points falling below the diagonal where AI-assisted and unaided reading times are equal (Fig.~\ref{fig:RCT_1}d-f). In adjusted GEE analysis, AI assistance was associated with a 14.9\% shorter diagnostic time (adjusted time ratio 0.851, $P < 0.05$) with an even larger reduction among junior pathologists (17.7\%, adjusted time ratio 0.823, $P < 0.05$). 

\begin{figure}
\captionsetup{labelformat=empty}
\centering
\includegraphics[width=0.95\textwidth]{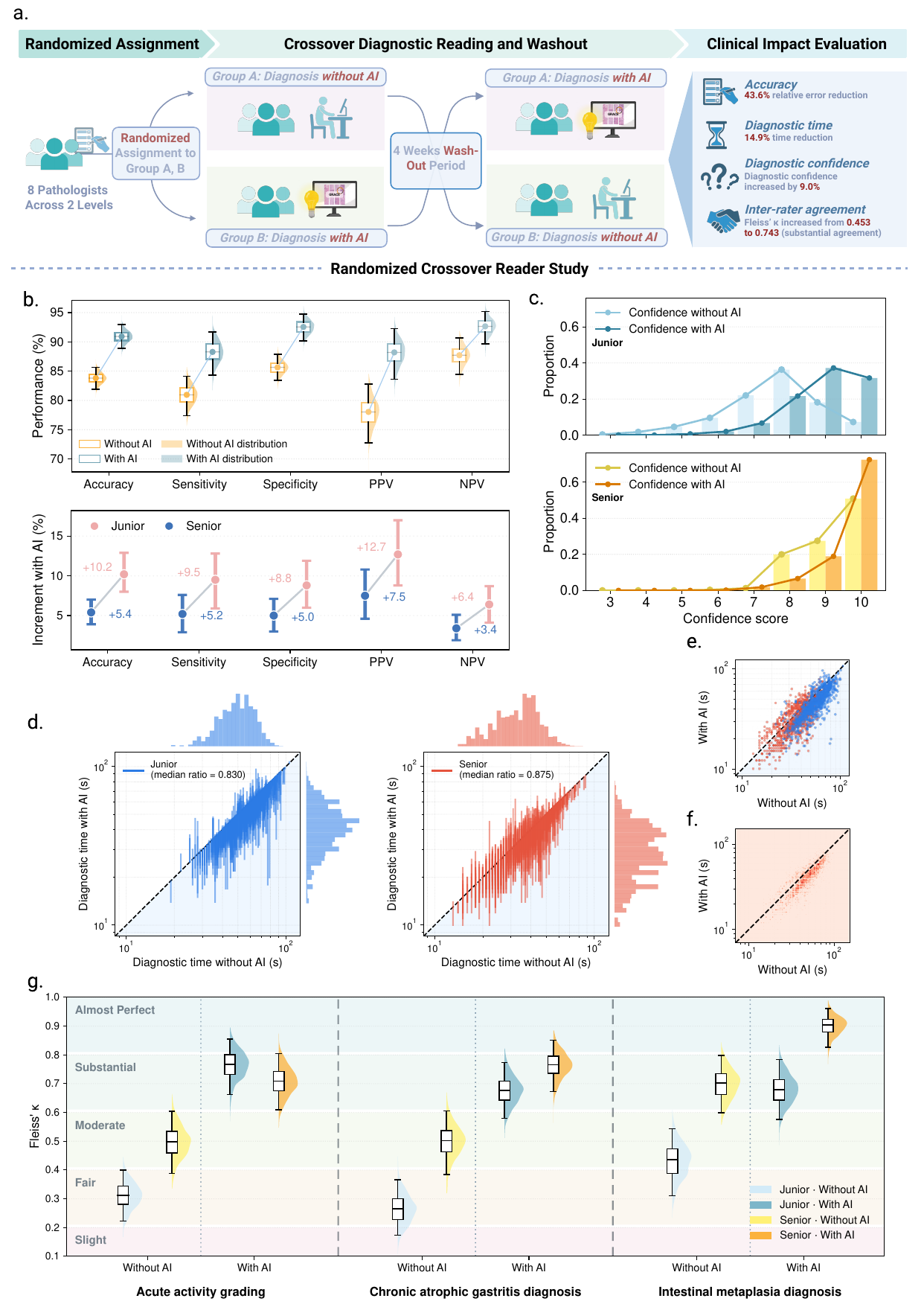}
\caption{}
\end{figure}
\begin{figure}
\ContinuedFloat
\caption{%
    \textbf{Analysis of pathologist-AI interaction in the randomized crossover reader study.}
    (\textbf{a}) Randomized crossover workflow with AI-assisted and unassisted reading separated by a four-week washout.
    (\textbf{b}) Diagnostic performance with unassisted and AI-assisted reading (top), and AI-associated gains by reader experience (bottom).
    (\textbf{c}) Distribution of self-reported diagnostic confidence during unassisted and AI-assisted reading, stratified by reader experience. Panels share the same x-axis.
    (\textbf{d}) Paired diagnostic time for AI-assisted versus unassisted reading, stratified by reader experience. Each vertical segment represents one paired reading and shows the change in time with AI assistance. Axes are logarithmic, and values below the dashed diagonal indicate shorter diagnostic time with AI assistance.
    (\textbf{e}) Paired diagnostic time shown as individual case-level points, with unassisted time on the x-axis and AI-assisted time on the y-axis. Axes are logarithmic, and points below the dashed diagonal indicate shorter diagnostic time with AI assistance. Blue and red indicate junior and senior readers, respectively.
    (\textbf{f}) Two-dimensional density map of the paired diagnostic time data in (\textbf{e}). Darker contours indicate higher density. The dashed diagonal marks equal reading time, and density below the diagonal indicates shorter diagnostic time with AI assistance.
    (\textbf{g}) Inter-rater agreement measured by Fleiss' $\kappa$, stratified by task, reader experience, and reading condition. Half-violins show the $\kappa$ distribution, boxplots summarize the median and 95\% CI, and shaded bands indicate agreement categories from slight to almost perfect\cite{landis1977measurement}.
    }
    \label{fig:RCT_1}
    \end{figure}
    
\paragraph{Diagnostic consistency through improved inter-rater agreement.}
GRACE effectively mitigated subjective variability and promoted diagnostic consensus among pathologists.
Fleiss' $\kappa$ increased from 0.453 without AI to 0.743 with AI, reflecting a significant improvement ($P < 0.05$, Extended Data Table~\ref{tab:fleiss}). 
Overall, AI-assisted agreement reached the substantial range and, among senior pathologists evaluating IM diagnosis, reached the almost-perfect range\cite{landis1977measurement} (Fig.~\ref{fig:RCT_1}g).
Notably, the effect was strongest among junior pathologists ($\Delta\kappa = +0.367$, $P < 0.05$), indicating that GRACE helped narrow the experience gap by increasing interobserver agreement and improving diagnostic accuracy, confidence, and efficiency.

\paragraph{The role of GRACE in pathologists' diagnoses.}
We characterized the impact of AI assistance on pathologists' diagnoses by classifying each diagnostic decision into one of five mutually exclusive pathologist-AI outcome categories (Extended Data Table~\ref{tab:ai_impact_categories}, see Methods: \nameref{methods:ai_impact_categorization}).
\textbf{AI correction} occurred in 9.63\% of diagnostic decisions overall, capturing instances in which AI assistance changed an initially incorrect interpretation into a correct diagnosis (Extended Data Table~\ref{tab:task_level_outcomes}). Correction events were more frequent in tasks with greater subjectivity, including acute activity severity grading (10.88\%) and atrophy diagnosis (10.50\%). 
Moreover, \textbf{AI reinforcement} represented the dominant interaction pattern, accounting for 74.67\% of reader diagnoses overall and 83.75\% in IM diagnosis. Together with \textbf{AI correction}, these patterns suggest that GRACE supported readers through two complementary mechanisms: reinforcing correct interpretations and correcting initially erroneous ones.
In comparison, \textbf{AI distraction} was uncommon across tasks, accounting for only 2.13\% of diagnoses overall and remaining lower than \textbf{AI correction} within each diagnostic task.
Additionally, \textbf{human safeguard} represented 5.58\% of diagnoses, indicating that pathologists sometimes rejected discordant AI input rather than accepting it uncritically.

Distributions of diagnostic time and confidence shifts across the five pathologist-AI outcome categories showed that diagnostic workflow improvements were concentrated in categories where AI either reinforced or corrected the reader's diagnosis. Assessments classified as \textbf{AI correction} or \textbf{AI reinforcement} were more often completed faster with AI assistance (86.1\% and 84.3\%, respectively), and 42.0\% of \textbf{AI correction} diagnostic decisions achieved a reading-time reduction greater than 20\% (Extended Data Fig.~\ref{fig:AI_impact_cat_line}f). Confidence was also generally maintained or increased in these categories (non-negative confidence shift: 94.8\% for \textbf{AI correction} and 96.6\% for \textbf{AI reinforcement}), with most changes concentrated between 0 and +3 points (Extended Data Fig.~\ref{fig:AI_impact_cat_line}e).
By contrast, less favorable categories showed weaker workflow facilitation. While \textbf{AI distraction} and \textbf{shared failure} represented smaller subsets of diagnostic decisions, slower reading occurred more often in these groups (27.5\% and 25.5\%, respectively), suggesting that these interactions sometimes required additional deliberation rather than accelerating interpretation.

Taken together, these findings answer a practical question about AI’s role in clinical settings. GRACE was most useful as an assistive diagnostic partner, reinforcing correct human judgments and correcting a measurable subset of initial errors. This pattern suggests that the value of AI lies not in replacing expert interpretation, but in improving the reliability, efficiency, and consistency of human diagnostic decision-making.

\paragraph{Pathologist-referenced review triage analysis.}
Having established that GRACE improved assisted reader performance, we next evaluated whether GRACE could extend AI support to workflow-level, expert-referenced review prioritization in high-volume gastric mucosal lesion diagnosis.
GRACE rule-out and rule-in thresholds were accepted only when their NPV and PPV met or exceeded conservative senior-pathologist benchmarks from the reader study (see Methods: \nameref{methods:expert_referenced_review_triage}). 
At the selected non-inferiority-constrained operating point, GRACE identified substantial triage-eligible subsets, including 60.7\% of cases for atrophy diagnosis and 82.7\% for IM diagnosis (Extended Data Fig.~\ref{fig:AI_impact_cat_line}g, h). Under this framework, GRACE could stratify review priority, flagging ruled-out cases as candidates for abbreviated concordance checking and ruled-in cases for prioritized targeted confirmation and reporting. These findings indicate that GRACE supports a more efficient collaborative review workflow for high-volume mucosal lesion assessment, concentrating pathologist expertise where clinical judgment is most consequential.

\section*{Discussion}
We introduce GRACE, a gastric-specialized PFM that delivers top-ranked performance, with translational credibility established through multi-institutional prospective validation and reader-study evidence of effective pathologist-AI collaboration.
What distinguishes GRACE from pancancer PFMs is its breadth across the gastric pathology workflow and its ability to support progressively more complex clinical decisions.
GRACE extends from mucosal lesion diagnosis and tumor characterization to biomarker prescreening and survival-risk stratification, demonstrating that routine gastric H\&E morphology contains decision-relevant signals beyond conventional histological classification.
These signals include morphology-associated correlates of molecular state and patient outcome, as reflected by GRACE’s leading molecular prediction performance, with the highest mean macro-AUC (0.8682), and by survival analyses that stratified patients into distinct DFS and OS risk groups.
Together, these results position GRACE as a disease-focused foundation model  for gastric precision pathology, converting routine H\&E morphology into clinically actionable evidence for precision triage, patient stratification, and care planning.

The advantage of GRACE over generalized pancancer PFMs highlights a shift from coarse pan-tissue pattern recognition toward a biologically grounded encoding of gastric pathology.
General-purpose PFMs, trained across diverse organ systems, are well positioned to capture broad cross-tissue contrasts.
Continued pretraining on gastric data tuned GRACE toward subtler intra-disease variation, such as heterogeneous histological patterns and inflammatory changes that are associated with different gastric tissue states and complicate pathological interpretation.
Consequently, rather than relying primarily on coarse visual similarity, GRACE may encode a potential biological architecture that helps resolve complex diagnostic ambiguities in gastric pathology.
Supportive spatial transcriptomic analyses (Extended Data Fig.~\ref{fig:ST}) showed that GRACE-derived feature partitions were broadly concordant with pathologist annotations and marker-defined tissue compartments, suggesting that the learned feature space reflects biologically organized tissue compartments. 
Epithelial-associated markers such as EPCAM\cite{wenqi2009epcam} were enriched in glandular regions, whereas MKI67\cite{xiong_ki-67mki67_2019} highlighted more restricted proliferative subregions overlapping tumor-associated areas, and CD68 marked immune cell-rich regions adjacent to epithelial compartments. 
These spatial-molecular alignments suggest that GRACE-derived features capture tissue compartment organization and biologically relevant variation within the gastric microenvironment.

Crucially, GRACE addresses a central translational barrier for PFMs by anchoring benchmark performance in clinically grounded evidence across prospective generalizability, safety-constrained workflow analysis, and pathologist-AI collaboration.
Multi-institutional prospective validation first supported GRACE's robustness across institutions and tissue-processing protocols, establishing external credibility beyond retrospective benchmarks. 
Safety-constrained analyses using retrospectively locked thresholds then addressed concrete workflow bottlenecks, providing evidence that GRACE could potentially accelerate malignancy triage, reduce review burden for tumor characterization, and prioritize confirmatory IHC work-up. 
Progressing from locked-threshold workflow simulations to real-world pathologist-AI evaluation, the randomized crossover reader study tested GRACE in realistic collaboration settings, showing that GRACE strengthened diagnostic performance by stabilizing correct decisions, recovering errors, and concurrently improving efficiency, confidence, and inter-reader consistency.
We then evaluated whether these reader-level gains could scale into workflow benefit. GRACE could enable substantial review triage in high-volume mucosal-lesion diagnosis while maintaining performance non-inferior to the senior pathologist reference. Together, these findings move GRACE beyond a high-performing PFM by defining an evidence-supported translational pathway for a deployable gastric pathology AI framework.

This study also has some limitations.
Although GRACE was pretrained on geographically mixed datasets, its clinical validation cohorts were predominantly Asian, reflecting the high burden of gastric cancer in Asia \cite{shin_updated_2023}.
Evaluation of GRACE in racially, ethnically, or geographically diverse populations may help to establish its broader generalizability. 
Beyond population diversity, another boundary of the present study is the single-modality scope.
Considering that GC care is intrinsically multimodal, a future gastric-disease foundation model integrating endoscopy, radiology, and longitudinal clinical records could contextualize pathology-derived signals and enable more clinically faithful risk stratification.
From a clinical deployment perspective, GRACE offers rigorous pathology evidence but it does not yet provide the stepwise coordination required of agentic AI systems. Future multi-agent systems could query GRACE for verifiable histology-based evidence, enabling its outputs to be contextualized, communicated, and audited across the longitudinal care pathway.

To our knowledge, GRACE is among the first gastric disease-specialized PFMs evaluated across multi-institutional prospective cohorts and a randomized crossover pathologist-AI collaboration reader study within realistic diagnostic workflows. By demonstrating workflow-embedded support with measurable improvements in accuracy, efficiency, confidence, and inter-reader consistency, GRACE moves beyond benchmark-level model evaluation toward a clinically grounded gastric PFM for scalable pathology decision support.

\clearpage

\section*{Dataset and methods}
\subsection*{Pretraining datasets}

During pretraining, we included private biopsy and surgical WSIs from Hospital H7 (23,708 slides). We further incorporated publicly available datasets, including HistAI\cite{nechaev2025histai} (\url{https://github.com/HistAI/HISTAI}), The Cancer
Genome Atlas (TCGA) (\url{https://www.cancer.gov/tcga}), Comprehensive Assessment of Chronic Gastritis on WSI Data\cite{xia2025gastritismil}(\url{https://www.scidb.cn/detail?dataSetId=83ee1521074742cdaae997cf0b46a7b1}), Gastric Cancer Lymph Node Dataset\cite{Chen2020}(\url{https://figshare.com/articles/dataset/Gastric_cancer_lymph_node_data_set/13065986}), and KBSMC Gastric Cancer Grading Dataset (\url{https://github.com/colin19950703/KBSMC_gastric_cancer_grading_dataset}). A detailed summary of the pretraining dataset is included in Extended Data Table~\ref{tab:pretrain_dataset}.

\subsection*{Downstream evaluation datasets}
Downstream model evaluation was conducted on multicenter in-house cohorts with no patient- or slide-level overlap with the pretraining dataset.
For downstream evaluation, GRACE was assessed using 22,645 WSIs from 11,774 patients collected from 9 hospitals. To standardize reporting across endpoints and validation settings, we defined three core benchmarking units: task, cohort, and case (Extended Data Table~\ref{tab:defs_cohort_task_case}).
These datasets were collected across hospitals with diverse clinical settings, including differences in patient populations, scanner vendors and platforms (e.g., Leica Aperio, 3DHISTECH, KFBIO, and TEKSQRAY), data formats (e.g., SDPC, KFB, SVS, MRXS, and TIF), and staining protocols.
The datasets were divided into internal cohorts, which supported downstream model training, validation, and internal testing, and separate external cohorts that were reserved strictly for evaluation with the final fixed model weights obtained during internal validation. For each individual task, the internal, external, and prospective cohorts were mutually exclusive. Detailed case and slide distribution for downstream task evaluation is provided in Extended Data Table~\ref{tab:hospital_dataset}.

\subsubsection*{Abnormal gastric biopsy tissue diagnosis}
Abnormal biopsy samples comprised multiple precancerous or clinically abnormal conditions collected from Hospital H5, including \textit{H. pylori}-associated chronic gastritis (HPACG), autoimmune chronic gastritis with \textit{H. pylori} (ACGHP), gastric polyps, gastric ulcers, and intestinal metaplasia (IM). The normal tissue category was additionally expanded to include chronic gastritis without \textit{H. pylori} infection (CGxHP). To evaluate GRACE’s ability to distinguish abnormal tissue from normal tissue or CGxHP, we defined a binary abnormality-detection task comparing these abnormal biopsy categories against the combined normal/CGxHP group. We curated a cohort of 1,327 patients and 2,374 slides from Hospital H5, comprising 1,083 abnormal cases (1,964 slides) and 244 normal/CGxHP cases (410 slides). The dataset was split into training, validation, and test sets in a 7:1:2 ratio, unless otherwise specified. Comprehensive performance metrics are provided in Extended Data Table~\ref{tab:Normal_and_Abnormal_Tissue_Diagnosis}.

\subsubsection*{Diagnosis of \textit{H. pylori} infection}
To evaluate GRACE's performance in detecting \textit{Helicobacter pylori} (\textit{H. pylori}) infection, we conducted a binary classification across multiple cohorts. The Hospital H2 surgical cohort served as the internal validation cohort with 172 cases and 1,007 slides, comprising 60 \textit{H. pylori}-positive cases (368 slides) and 112 \textit{H. pylori}-negative cases (639 slides). To examine cross-cohort generalization, we included a retrospective external validation cohort from Hospital H1 with 130 cases (131 slides), encompassing 30 \textit{H. pylori}-positive cases (31 slides) and 100 \textit{H. pylori}-negative cases (100 slides). Comprehensive performance metrics are summarized in Extended Data Tables~\ref{tab:HP_Infection_Diagnosis} and \ref{tab:H_pylori_Infection_Diagnosis_External_H1}.

\subsubsection*{Diagnosis of autoimmune chronic gastritis with \textit{H. pylori}}
To assess GRACE's performance in diagnosing ACGHP, we constructed a dataset from the Hospital H5 biopsy cohort consisting of 1,327 cases (2,545 slides), including 149 ACGHP cases (185 slides) and 1,178 non-ACGHP cases (2,360 slides). Comprehensive performance metrics are presented in Extended Data Table~\ref{tab:Autoimmune_Chronic_Gastritis_with_HP_Diagnosis}.

\subsubsection*{Diagnosis of \textit{H. pylori}-associated chronic gastritis}
To evaluate the ability of GRACE to identify HPACG, we curated a dataset from the Hospital H5 biopsy cohort consisting of 1,327 cases (2,632 slides), including 135 HPACG cases (223 slides) and 1,192 non-HPACG cases (2,409 slides). Comprehensive performance metrics are summarized in Extended Data Table~\ref{tab:HP-Associated_Chronic_Gastritis_Diagnosis}.

\subsubsection*{Diagnosis of chronic gastritis}
Chronic gastritis (CG) is a persistent inflammation of the gastric mucosa, which can result in mucosal injury\cite{sipponen2015chronic}. To evaluate GRACE's performance in diagnosing CG, we assembled a gastric biopsy dataset for CG classification. The internal cohort comprised 5,519 cases (5,765 slides) from Hospital H1, with 3,633 CG cases (3,780 slides) and 1,886 non-CG cases (1,985 slides). For prospective validation on external cohorts, Hospital H1 provided 950 cases (971 slides), including 499 CG (513 slides) and 451 non-CG cases (458 slides). An additional prospective validation cohort from Hospital H9 was included, totaling 169 cases (272 slides) with 132 CG cases (232 slides) and 37 non-CG cases (40 slides). Comprehensive performance metrics are summarized in Extended Data Tables~\ref{tab:Chronic_Gastritis_Diagnosis}, \ref{tab:Chronic_Gastritis_Diagnosis_Prospective_H1} and \ref{tab:Chronic_Gastritis_Diagnosis_Prospective_H9}.

\subsubsection*{Chronic gastritis severity grading}
Given the established association between CG severity and subsequent gastric cancer (GC) risk, we formulated a three-class severity grading task using a subcohort from Hospital H1, stratifying cases into mild, moderate, and severe categories. The internal dataset comprised 3,632 patients (3,779 slides), including 1,379 mild (1,428 slides), 2,004 moderate (2,093 slides), and 249 severe cases (258 slides). Prospective validation was conducted using two independent cohorts. A prospective cohort from Hospital H1 included 180 cases (186 slides), consisting of 80 mild (85 slides), 80 moderate (80 slides), and 20 severe cases (21 slides). An additional prospective cohort from Hospital H2 comprised 347 cases (348 slides), spanning 251 mild (252 slides), 94 moderate (94 slides), and 2 severe cases (2 slides). Comprehensive performance metrics are summarized in Extended Data Tables~\ref{tab:Chronic_Gastritis_Severity_Grading}, \ref{tab:Chronic_Gastritis_Severity_Grading_Prospective_H1} and \ref{tab:Chronic_Gastritis_Severity_Grading_Prospective_H2}.

\subsubsection*{Diagnosis of acute activity in chronic gastritis}

To evaluate GRACE for detecting acute inflammatory activity in CG, we constructed an internal biopsy validation cohort from Hospital H1 comprising 2,211 cases (2,303 slides), including 2,000 cases with acute activity (2,082 slides) and 211 cases without acute activity (221 slides). Prospective validation was performed across three independent cohorts. The first cohort from Hospital H7 included 332 cases (578 slides), with 198 cases exhibiting acute activity (332 slides) and 134 cases without acute activity (246 slides). The second cohort from Hospital H8 comprised 29 cases (30 slides), including 13 cases with acute activity (14 slides) and 16 cases without acute activity (16 slides). The third cohort from Hospital H9 contained 120 cases (204 slides), including 35 cases with acute activity (59 slides) and 85 cases without acute activity (145 slides). Detailed performance metrics are provided in Extended Data Tables~\ref{tab:diagnosis_of_Acute_Activity_in_Chronic_Gastritis}, \ref{tab:Diagnosis_of_Acute_Activity_in_Chronic_Gastritis_Prospective_H7}, \ref{tab:Diagnosis_of_Acute_Activity_in_Chronic_Gastritis_Prospective_H8}, and \ref{tab:Diagnosis_of_Acute_Activity_in_Chronic_Gastritis_Prospective_H9}.

\subsubsection*{Acute activity severity grading in chronic gastritis}
We next examined GRACE’s ability to grade the severity of acute inflammatory activity in CG. In the Hospital H1 biopsy cohort, 3,142 patients (3,273 slides) were categorized as mild (2,092 cases, 2,180 slides), moderate (901 cases, 936 slides), or severe (149 cases, 157 slides). Prospective evaluation was performed using two independent cohorts. A separate cohort from Hospital H1 included 404 cases (414 slides), comprising 286 mild (294 slides), 103 moderate (104 slides), and 15 severe cases (16 slides). An additional prospective cohort from Hospital H7 consisted of 165 cases (280 slides), with 106 mild (185 slides), 52 moderate (86 slides), and 7 severe cases (9 slides). Detailed performance metrics are provided in Extended Data Tables~\ref{tab:Acute_Activity_Severity_Grading}, \ref{tab:Acute_Activity_Severity_Grading_Prospective_H1}, and \ref{tab:Acute_Activity_Severity_Grading_Prospective_H7}.

\subsubsection*{Atrophy diagnosis in chronic gastritis}
To assess GRACE’s ability to detect atrophic changes in CG, we assembled a biopsy cohort of 3,363 patients (3,503 slides) from Hospital H1, including 1,961 cases with chronic atrophic gastritis (2,036 slides) and 1,402 without chronic atrophic gastritis (1,467 slides). Prospective validation was conducted across four independent cohorts. A prospective validation cohort from Hospital H1 included 474 cases (487 slides), with 238 atrophic cases (244 slides) and 236 non-atrophic cases (243 slides). The Hospital H2 cohort consisted of 300 cases (300 slides), including 44 atrophic cases (44 slides) and 256 non-atrophic cases (256 slides). The Hospital H8 cohort comprised 30 cases (31 slides), including 11 atrophic cases (12 slides) and 19 non-atrophic cases (19 slides). The Hospital H9 cohort included 126 cases (219 slides), with 43 atrophic cases (76 slides) and 83 non-atrophic cases (143 slides). Detailed performance metrics are reported in Extended Data Tables~\ref{tab:diagnosis_of_Atrophy_in_Chronic_Gastritis}, \ref{tab:Diagnosis_of_Atrophy_in_Chronic_Gastritis_Prospective_H1}, \ref{tab:Diagnosis_of_Atrophy_in_Chronic_Gastritis_Prospective_H2}, \ref{tab:Diagnosis_of_Atrophy_in_Chronic_Gastritis_Prospective_H8}, and \ref{tab:Diagnosis_of_Atrophy_in_Chronic_Gastritis_Prospective_H9}.

\subsubsection*{Atrophy severity grading in chronic gastritis}
Beyond the diagnostic classification of atrophy in patients with CG, GRACE was further evaluated for its ability to classify atrophy severity into two classes. The dataset from the Hospital H1 biopsy cohort included 685 cases (720 slides), comprising 500 mild cases (528 slides) and 185 moderate cases (192 slides). The experimental results are shown in Extended Data Table~\ref{tab:Atrophy_Severity_Grading}. 

\subsubsection*{Gastric polyp diagnosis}
To evaluate GRACE for gastric polyp classification, we constructed a dataset comprising 5,410 cases (5,621 slides) from Hospital H1, including 1,639 polyp cases (1,687 slides) and 3,771 non-polyp cases (3,934 slides). Prospective external validation was performed using three biopsy cohorts. The Hospital H1 cohort included 611 cases (618 slides), with 411 polyp cases (415 slides) and 200 non-polyp cases (203 slides). The Hospital H7 cohort consisted of 401 cases (696 slides), including 42 polyp cases (68 slides) and 359 non-polyp cases (628 slides). An additional cohort from Hospital H8 comprised 48 cases (49 slides), spanning 15 polyp cases (15 slides) and 33 non-polyp cases (34 slides). Detailed experimental results are presented in Extended Data Tables~\ref{tab:Polyp_Diagnosis}, \ref{tab:Polyp_Diagnosis_Prospective_H1}, \ref{tab:Polyp_Diagnosis_Prospective_H7}, and \ref{tab:Polyp_Diagnosis_Prospective_H8}.

\subsubsection*{Gastric ulcer diagnosis}
To evaluate GRACE’s performance in detecting gastric ulcers, we assembled a biopsy dataset comprising 5,410 cases (5,621 slides) from Hospital H1, including 140 ulcer cases (145 slides) and 5,270 non-ulcer cases (5,476 slides). Generalization was further assessed using a retrospective external biopsy cohort from Hospital H5. The Hospital H5 cohort included 366 cases (627 slides), with 123 ulcer cases (131 slides) and 243 non-ulcer cases (496 slides). Detailed results are presented in Extended Data Tables~\ref{tab:Ulcer_Diagnosis} and \ref{tab:Ulcer_Diagnosis_External_H5}.

\subsubsection*{Gastric intestinal metaplasia diagnosis}
GRACE was evaluated for detecting intestinal metaplasia (IM) in gastric biopsy specimens. The internal dataset included 3,386 patients (3,527 slides) from the Hospital H1 biopsy cohort, with 1,532 IM cases (1,601 slides) and 1,854 non-IM cases (1,926 slides). External validation was performed across both retrospective and prospective cohorts. The retrospective Hospital H5 cohort included 1,327 cases (2,496 slides), consisting of 42 IM cases (67 slides) and 1,285 non-IM cases (2,429 slides). The prospective Hospital H1 cohort included 479 cases (492 slides), including 198 IM cases (205 slides) and 281 non-IM cases (287 slides). The prospective Hospital H2 cohort included 372 cases (373 slides), including 126 IM cases (126 slides) and 246 non-IM cases (247 slides). The prospective Hospital H8 cohort included 34 cases (35 slides), including 16 IM cases (17 slides) and 18 non-IM cases (18 slides). The prospective Hospital H9 cohort included 131 cases (230 slides), including 48 IM cases (88 slides) and 83 non-IM cases (142 slides). Detailed experimental results are presented in Extended Data Tables \ref{tab:Intestinal_Metaplasia_Diagnosis}, \ref{tab:Intestinal_Metaplasia_Diagnosis_External_H5}, \ref{tab:Intestinal_Metaplasia_Diagnosis_Prospective_H1}, \ref{tab:Intestinal_Metaplasia_Diagnosis_Prospective_H2}, \ref{tab:Intestinal_Metaplasia_Diagnosis_Prospective_H8}, and \ref{tab:Intestinal_Metaplasia_Diagnosis_Prospective_H9}.

\subsubsection*{Malignancy diagnosis}
To assess GRACE's performance in distinguishing malignant from non-malignant cases using gastric biopsy slides, we analyzed the Hospital H1 biopsy cohort which included 5,519 cases (5,765 slides), comprising 109 malignant cases (144 slides) and 5,410 non-malignant cases (5,621 slides). A prospective validation cohort from Hospital H7 was included with 546 cases (931 slides), spanning 131 malignant cases (244 slides) and 415 non-malignant cases (687 slides). Another prospective validation cohort from Hospital H8 comprised 102 cases (127 slides), including 47 malignant cases (71 slides) and 55 non-malignant cases (56 slides). Detailed experimental results are presented in Extended Data Tables~\ref{tab:Malignancy_Diagnosis}, \ref{tab:Malignancy_Diagnosis_Prospective_H7} and \ref{tab:Malignancy_Diagnosis_Prospective_H8}.

\subsubsection*{Pathology subtyping}
We developed a three-class pathology subtyping task using 4,060 surgically resected slides from 682 patients at Hospital H2, distinguishing signet-ring cell carcinoma (SRC), tubular adenocarcinoma, and non-specified adenocarcinoma. The cohort included 81 SRC cases (507 slides), 269 tubular adenocarcinoma cases (1,571 slides), and 332 non-specified adenocarcinoma cases (1,982 slides). Prospective external validation was performed using two independent cohorts. The Hospital H1 cohort comprised 33 cases (38 slides), including 14 SRC cases (14 slides), 15 tubular adenocarcinoma cases (20 slides), and 4 non-specified adenocarcinoma cases (4 slides). An additional prospective cohort from Hospital H7 included 179 cases (4,728 slides), spanning 3 SRC cases (90 slides), 143 tubular adenocarcinoma cases (4,119 slides), and 33 non-specified adenocarcinoma cases (519 slides). Detailed experimental results are summarized in Extended Data Tables~\ref{tab:Pathology_Subtyping}, \ref{tab:Pathology_Subtyping_Prospective_H1} and \ref{tab:Pathology_Subtyping_Prospective_H7}.

\subsubsection*{Lauren classification}
Lauren classification provided a complementary histopathologic classification framework. We developed a binary classification task, distinguishing intestinal-type and diffuse-type gastric adenocarcinomas. The internal cohort was constructed from a pooled dataset across Hospital H1, H3, and H4 with 602 cases (604 slides), including 223 intestinal-type cases (224 slides) and 379 diffuse-type cases (380 slides). Prospective validation was performed using a Hospital H7 cohort with 73 cases (2,067 slides), including 37 intestinal-type cases (1,010 slides) and 36 diffuse-type cases (1,057 slides). Detailed experimental results are summarized in Extended Data Tables~\ref{tab:Lauren_Classification} and \ref{tab:Lauren_Classification_Prospective_H7}.

\subsubsection*{Histologic differentiation grading}
We evaluated GRACE for histologic differentiation grading by formulating a binary classification task distinguishing poorly differentiated tumors (G3) from well- and moderately differentiated tumors (G1+G2). The internal development cohort comprised 684 cases with 4,072 surgically resected slides from Hospital H2, including 523 G3 cases (3,138 slides) and 161 G1+G2 cases (934 slides). Model generalizability was assessed using two independent external retrospective cohorts. Hospital H3 included 217 cases (217 slides), consisting of 168 G3 cases (168 slides) and 49 G1+G2 cases (49 slides). Hospital H4 included 118 cases (118 slides), with 59 G3 cases (59 slides) and 59 G1+G2 cases (59 slides). Prospective validation was further conducted across three independent cohorts. The prospective cohort from Hospital H2 comprised 92 cases (135 slides), including 50 G3 cases (78 slides) and 42 G1+G2 cases (57 slides). Hospital H7 contributed 145 cases (4,044 slides), with 100 G3 cases (2,908 slides) and 45 G1+G2 cases (1,136 slides). Hospital H8 included 43 cases (43 slides), consisting of 6 G3 cases (6 slides) and 37 G1+G2 cases (37 slides). Detailed experimental results are summarized in Extended Data Tables~\ref{tab:Tumor_Differentiation_Grading}, \ref{tab:Tumor_Differentiation_Grading_External_H3}, \ref{tab:Tumor_Differentiation_Grading_External_H4}, \ref{tab:Tumor_Differentiation_Grading_Prospective_H2}, \ref{tab:Tumor_Differentiation_Grading_Prospective_H7}, and \ref{tab:Tumor_Differentiation_Grading_Prospective_H8}.

\subsubsection*{T-staging classification}
To evaluate GRACE's performance on pathological T-staging, we constructed a cohort from Hospital H2 with 583 cases (3,431 slides), including 193 T1 cases (1,149 slides), 98 T2 cases (550 slides), 134 T3 cases (810 slides), and 158 T4 cases (922 slides). Model generalizability was evaluated using an independent external retrospective cohort from Hospital H6 with 186 cases (348 slides), consisting of 40 T1 cases (74 slides), 24 T3 cases (41 slides), 35 T2 cases (75 slides), and 87 T4 cases (158 slides).
Prospective validation was further performed on an additional cohort from Hospital H2 comprising 94 cases (138 slides), including 40 T1 cases (55 slides), 13 T2 cases (17 slides), 23 T3 cases (35 slides), and 18 T4 cases (31 slides). Detailed experimental results are summarized in Extended Data Tables~\ref{tab:T-Staging_Internal_H2}, \ref{tab:T-Staging_External_H6} and \ref{tab:T-Staging_Prospective_H2}.

\subsubsection*{N-staging prediction}
N-staging represents a critical determinant of prognosis and treatment planning in GC, influencing surgical strategy, the extent of lymphadenectomy, and the need for adjuvant therapy\cite{ji2023lymph}. GRACE predicts nodal involvement based on morphologic features of the primary tumor. Accordingly, we formulated a binary N-staging prediction task distinguishing node-negative disease (N0) from node-positive disease (N+). Using surgically resected gastric tumor WSIs from Hospital H2, we constructed a binary N-staging prediction task with 694 cases (4,114 slides), including 361 N0 cases (2,160 slides) and 333 N+ cases (1,954 slides). For retrospective external validation, an independent cohort from Hospital H1 with 398 cases (400 slides) was used, including 212 N+ cases (212 slides) and 186 N0 cases (188 slides). External prospective validation was conducted within an independent cohort from Hospital H1 with 52 cases (60 slides), spanning 20 N+ cases (22 slides) and 32 N0 cases (38 slides). Detailed experimental results are summarized in Extended Data Tables~\ref{tab:Lymph_Node_Metastasis_Prediction}, \ref{tab:Lymph_Node_Metastasis_Prediction_External_H1}, and \ref{tab:Lymph_Node_Metastasis_Prediction_Prospective_H1}.

\subsubsection*{Perineural invasion classification}
We formulated an internal cohort from Hospital H1 with 396 cases (398 slides) for perineural invasion (PNI) classification, including 255 PNI-positive cases (256 slides) and 141 PNI-negative cases (142 slides). External validation was performed using an independent retrospective cohort from Hospital H6 with 233 cases (437 slides), consisting of 71 PNI-positive cases (131 slides) and 162 PNI-negative cases (306 slides). Prospective validation was conducted across two cohorts. The prospective cohort from Hospital H1 included 60 cases (69 slides), with 30 PNI-positive cases (32 slides) and 30 PNI-negative cases (37 slides). The prospective cohort from Hospital H8 contributed 38 cases (38 slides), comprising 25 PNI-positive cases (25 slides) and 13 PNI-negative cases (13 slides). Detailed experimental results are summarized in Extended Data Tables~\ref{tab:Perineural_Invasion_}, \ref{tab:Perineural_Invasion_External_H6}, \ref{tab:Perineural_Invasion_Prospective_H1} and \ref{tab:Perineural_Invasion_Prospective_H8}.

\subsubsection*{Vascular invasion classification}
For vascular invasion (VI), data from three hospitals (H1, H3, and H4) were pooled into a single evaluation cohort with 944 cases and 946 slides, including 459 VI-positive cases (460 slides) and 485 VI-negative cases (486 slides). Detailed results are provided in Extended Data Table~\ref{tab:Vascular_Invasion}.

\subsubsection*{Tumor regression grade classification}
The tumor regression grade (TRG) evaluates the degree of tumor regression after cytotoxic treatment. According to the American Joint Committee on Cancer (AJCC) cancer staging system\cite{langer2018tumor}, cases with TRG 0-1 were grouped as low residual tumor cases, while those with TRG 2-3 were grouped as high residual tumor cases. To evaluate GRACE's ability to classify TRG, we formulated a two-class classification task using 1,078 surgical resection slides from 165 patients at Hospital H2. The dataset comprised 28 low residual tumor cases (194 slides) and 137 high residual tumor cases (884 slides). Detailed experimental results are summarized in Extended Data Table~\ref{tab:Response_to_Chemotherapy_Diagnosis}.

\subsubsection*{HER2 status prediction}
A subset of GCs exhibits HER2 amplification, and aberrant HER2 expression serves as a targetable molecular driver\cite{malla_her-2_2024}. To evaluate GRACE's performance in predicting HER2 expression status, tumor cases were categorized according to CAP's template for gastric HER2 biomarker reporting\cite{bang2010trastuzumab}: IHC scores of 0 and 1+ were considered HER2-negative, 3+ was considered HER2-positive, and 2+ cases were treated as equivocal and excluded pending FISH confirmation. The internal dataset comprised 265 cases (1,598 slides) from the Hospital H2 surgical cohort, including 27 3+ expression cases (152 slides) and 238 0+ and 1+ expression cases (1,446 slides). Detailed experimental results are summarized in Extended Data Table~\ref{tab:HER2_Expression_Prediction}.

\subsubsection*{Mismatch repair status prediction}
Mismatch repair (MMR) status was assessed by IHC for MLH1, MSH2, MSH6, and PMS2. Tumors with retained nuclear expression of all four proteins were classified as proficient MMR (pMMR). Tumors with loss of nuclear expression of one or more MMR proteins were classified as deficient MMR (dMMR)\cite{park2023standardized}. The internal Hospital H2 cohort included 53 dMMR cases (310 slides) and 350 pMMR cases (2,101 slides). External retrospective validation was performed using a cohort with 119 cases from Hospital H1, including 19 dMMR and 100 pMMR cases, with one slide available per case. Prospective validation was conducted independently at Hospital H2 using 79 cases (116 slides), including 8 dMMR cases (11 slides) and 71 pMMR cases (105 slides). Detailed experimental results are summarized in Extended Data Tables~\ref{tab:MMR_Prediction}, \ref{tab:MMR_Prediction_External_H1} and \ref{tab:MMR_Prediction_Prospective_H2}.

\subsubsection*{Ki-67 proliferation index expression prediction}
GRACE incorporated Ki-67 expression prediction, and Ki-67 expression level was categorized using standard clinical thresholds as low ($\leq 10\%$ positive cells), medium ($10\% < x \leq 50\%$), and high ($>50\%$)\cite{xiong_ki-67mki67_2019}. Data from three hospitals (H1, H3, and H4) were combined into a single pooled cohort for evaluation. The combined surgical cohort comprised 705 patients, each represented by one corresponding WSI. Among them, 368 exhibited high Ki-67 expression, 258 showed medium expression, and 79 demonstrated low expression. Detailed experimental results are summarized in Extended Data Table~\ref{tab:Ki-67_Prediction}.

\subsubsection*{EBV status prediction}
TCGA recognizes EBV-positive GC as a distinct molecular subtype\cite{cancer2014comprehensive}. GRACE incorporated an EBV status prediction task using the surgical cohort from Hospital H2 with 1,854 slides from 313 patients, comprising 1,771 slides from 300 EBV-negative cases and 83 slides from 13 EBV-positive cases. Detailed experimental results are summarized in Extended Data Table~\ref{tab:EBV_Status_Prediction}.

\subsubsection*{Disease-free survival prediction}
Using surgically resected specimens, GRACE performed disease-free survival (DFS) prediction using a pooled internal cohort from Hospitals H3 and H4. This cohort included 448 disease-free cases (448 slides) and 132 cases with recurrence or progression (132 slides). The model was trained and validated using five-fold cross-validation. For retrospective external validation, we applied the trained model to an unseen dataset from Hospital H1, comprising 368 disease-free cases (472 slides) and 57 cases with recurrence or progression (69 slides). Detailed experimental results are summarized in Extended Data Tables \ref{tab:Survival_DFS_OS_prediction}  and \ref{tab:Survival_DFS_OS_prediction_external}.

\subsubsection*{Overall survival prediction}
GRACE performed overall survival (OS) prediction from a pooled internal cohort from Hospitals H3 and H4, consisting of 374 living cases (374 slides) and 206 deceased cases (206 slides). The model was trained and validated using five-fold cross-validation. For external validation, we applied the trained model to the Hospital H1 cohort, comprising 203 living cases (258 slides) and 42 deceased cases (50 slides). Detailed experimental results are summarized in Extended Data Tables~\ref{tab:Survival_DFS_OS_prediction} and \ref{tab:Survival_DFS_OS_prediction_external}.

\subsection*{Method}
\subsubsection*{Data curation and quality control}
\paragraph{Standardized preprocessing.}
All WSIs were processed with a common preprocessing pipeline to reduce technical variation across cohorts. Tissue detection was first applied to isolate foreground tissue and remove background regions, using a fixed parameter set for reproducibility. Patch extraction was performed from the native scanning level (level 0). WSIs were scanned at either 20$\times$ or 40$\times$ magnification, and patches were resized to ensure a consistent effective resolution of 0.25~$\mu$m/pixel. To reduce overrepresentation of large resections or slides with extensive tissue area, extraction was capped at 5,000 patches per slide.
\paragraph{Quality control.}
WSIs and corresponding clinical reports were reviewed through a multi-step quality-control process before model development. After digitization, slides were screened for image-quality problems that could affect pathological assessment, including out-of-focus regions, tissue folds, incomplete tissue sections, tissue loss, pen markings, air bubbles, and major staining defects. Cases with insufficient image quality or incomplete clinical information were flagged for adjudication before inclusion.
Gastrointestinal pathologists reviewed cases from each cohort to confirm diagnostic annotations, pathology labels, and relevant clinical metadata. Baseline diagnoses were independently checked by primary and senior pathologists. Discrepant cases were resolved by consensus discussion, and confirmed labeling errors were corrected before analysis.

\paragraph{Data imbalance in clinical cohorts.}
Several tasks in the benchmark involve clinical cohorts with naturally imbalanced class distributions, where the target condition may have low prevalence and some histological subtypes are rare. These distributions reflect the clinical evaluation cohorts and were not artificially balanced during dataset construction.

\paragraph{Spatial transcriptomics analysis.}
For the spatial transcriptomics (ST) case study shown in Extended Data Fig.~\ref{fig:ST}, we used data from a human gastric cancer specimen profiled with the 10x Xenium platform. The data were obtained from a publicly available dataset reported by Schroeder et al.\cite{schroeder2025scaling} (\url{https://zenodo.org/records/15164980}).
For downstream analysis, the ST data were spatially aggregated into 100~\textmu m $\times$ 100~\textmu m grid bins using the HEST-1k toolkit. GRACE features were extracted from the corresponding histology image and aggregated at the same grid-bin resolution. K-means clustering was then applied to the image-derived GRACE features to generate feature clusters. Gene-level expression maps were generated by projecting log-transformed expression values onto the tissue coordinates. These maps were used for qualitative comparison between GRACE image-feature clusters, pathologist annotation, and spatial patterns of selected marker-gene expression.

\subsubsection*{Model design}
Many recent computational PFMs have been developed through self-supervised pretraining, particularly using frameworks such as DINO\cite{DINO}. Following this paradigm, we initialized the backbone with Virchow2\cite{virchow2} weights and performed gastric-specific continued self-supervised pretraining using the DINO objective. Low-rank adaptation (LoRA) was used to make this continued pretraining parameter-efficient.

\paragraph{LoRA-based gastric-specific adaptation.}
In the GRACE implementation, LoRA was configured through PEFT with rank $r=8$, scaling factor $\alpha=16$, and dropout 0.1, and applied to attention projection layers. Under the standard PEFT LoRA formulation, each adapted linear projection can be represented as
\[
\widetilde{W}=W+\Delta W,\qquad
\Delta W=\frac{\alpha}{r}BA.
\]

\paragraph{Optimization and training.}
Pretraining used AdamW with the default PyTorch momentum parameters
$(\beta_1=0.9,\beta_2=0.999)$, mixed-precision training with automatic loss scaling, and cosine schedules for the learning rate, weight decay, and teacher momentum. The learning rate was linearly scaled by the effective global batch size and annealed to $10^{-6}$ after warm-up. Weight decay was cosine-annealed from 0.04 to 0.4. Distributed training was performed on one node equipped with $8\times80$\,GB NVIDIA H800 GPUs, using distributed data parallelism.

\paragraph{Patient-level classification with ABMIL.}
For downstream classification, pre-extracted patch features were used as fixed inputs to an ABMIL~\cite{ilse2018attention} classifier. Each case or patient $p$ was represented as a bag
\[
\mathbf{X}_p=\{\mathbf{x}_{pi}\}_{i=1}^{N_p},
\]
where $\mathbf{x}_{pi}\in\mathbb{R}^{D}$ denotes a patch-level feature. If multiple WSIs were assigned to the same case, their feature tensors were concatenated into the same bag.

Each instance was projected into a latent space by
\[
\mathbf{h}_{pi}
=
\phi(\mathbf{W}_f\mathbf{x}_{pi}+\mathbf{b}_f),
\qquad
\mathbf{h}_{pi}\in\mathbb{R}^{512},
\]
where $\phi(\cdot)$ is ReLU. The implemented feature block also applies dropout with probability 0.25 after this activation.

Single-head attention was computed as
\[
e_{pi}
=
\mathbf{w}^{\top}
\tanh(\mathbf{V}\mathbf{h}_{pi}+\mathbf{b}_v)+b,
\]
where $\mathbf{V}\in\mathbb{R}^{128\times512}$ and
$\mathbf{w}\in\mathbb{R}^{128}$. Attention weights were normalized over all instances in the bag:
\[
a_{pi}
=
\frac{\exp(e_{pi})}{\sum_{j=1}^{N_p}\exp(e_{pj})}.
\]
The case-level representation was then
\[
\mathbf{z}_p
=
\sum_{i=1}^{N_p}a_{pi}\mathbf{h}_{pi},
\qquad
\mathbf{z}_p\in\mathbb{R}^{512},
\]
followed by a linear classifier
\[
\hat{\mathbf{y}}_p=\mathbf{W}_c\mathbf{z}_p+\mathbf{b}_c,
\qquad
\mathbf{p}_p=\mathrm{softmax}(\hat{\mathbf{y}}_p).
\]

ABMIL was trained with patient/case-level cross-entropy on the logits:
\[
\mathcal{L}_{\mathrm{cls}}
=
-\sum_{c=1}^{C}y_{pc}\log p_{pc},
\]
with losses averaged over training cases. Each downstream task was trained independently on pre-extracted features. Performance was evaluated at the case/patient level, and uncertainty was estimated using 1,000 bootstrap resamples as implemented in the evaluation metrics.

\subsubsection*{Rationale for LoRA-based continued pretraining}
\paragraph{LoRA vs. full fine-tuning.}
We adopted LoRA because it reduces the number of trainable parameters during adaptation without compromising performance relative to full fine-tuning (FFT).
In our implementation, LoRA optimizes 26.2M trainable parameters, corresponding to 3.978\% of the 655M-parameter backbone, whereas FFT updates all 655M backbone parameters. This represents an approximately $25.1\times$ reduction in the number of trainable parameters.

Ablations across representative clinical endpoints and validation settings (Extended Data Tables~\ref{tab:ablation_malignancy_diagnosis}, \ref{tab:ablation_diagnosis_of_grading_chronic_gastritis}, \ref{tab:ablation_atrophy_severity_grading}) supported comparable performance between LoRA and FFT. LoRA was numerically higher in most comparisons, while FFT achieved slightly higher specificity at fixed sensitivity for the internal CG grading task. However, most differences were small relative to bootstrap uncertainty intervals.
Thus, LoRA provides parameter-efficient adaptation for gastric pathology tasks while reducing optimization cost and memory use.

\subsubsection*{Data split}
For internal model development, internal cohorts were partitioned at the patient level into training, validation, and internal test subsets using a 7:1:2 ratio. All slides from the same patient were assigned to the same subset. External and prospective cohorts were not used for model fitting, hyperparameter tuning, threshold selection, or model selection, and were evaluated only after the final model weights and thresholds had been fixed using the internal validation data.
For survival prediction, five-fold cross-validation was used for internal model development and validation.

\subsubsection*{Prospective validation deployment}
To evaluate the clinical robustness and real-world generalizability of GRACE, we conducted a multicenter, prospective observational study of consecutive patients undergoing routine biopsy or resection. Using prospectively accrued real-world clinical data, we assessed model performance across natural variation in case mix, tissue quality, and workflow conditions. Study design, reporting, and evaluation procedures were aligned with the DECIDE-AI guidelines\cite{vasey2022reporting} for early-stage clinical assessment of AI-based decision support systems. The human subjects review protocol was approved by the Medical Ethics Committee of Nanfang Hospital of Southern Medical University (approval No. NFEC-2025-403) on August 14, 2025. The study protocol was registered on ClinicalTrials.gov (ID: NCT07157618). The study was sponsored by Nanfang Hospital, Southern Medical University. 

\paragraph{Model locking.}
The GRACE feature extractor was frozen before downstream task development. The ABMIL model was trained exclusively on internal cohort data, and all model weights were locked before initiation of the retrospective and prospective external evaluations. No further training, hyperparameter tuning, threshold recalibration, or model selection was performed using any held-out evaluation cohort.

\paragraph{Eligibility criteria.}
Eligible patients were adults ($\geq 18$ years) who underwent biopsy or surgical resection for suspected or confirmed gastric lesions, had available H\&E-stained slides suitable for digitization with acceptable staining quality, and had complete clinical follow-up with a definitive pathological diagnosis. Exclusion criteria included insufficient slide quality for digital analysis, absence of a final pathological diagnosis, referral for external consultation without diagnostic consensus, and withdrawal of patient consent.

\paragraph{Study timeframe.} The prospective validation study was conducted across participating centers within an overall enrollment window from August 14, 2025, to February 1, 2026. At each center, patients were enrolled consecutively during the center's active enrollment period. All eligible cases were included according to the eligibility criteria, with no manual filtering or enrichment, to reduce selection bias and approximate routine diagnostic practice.
For each case, WSIs were processed automatically by the inference pipeline immediately after digitization and entry into the study dataset. 

\paragraph{Non-interventional use of GRACE.}
The prospective evaluation was observational and non-interventional. Model outputs were not made available to clinicians for routine clinical decision-making and therefore did not influence patient care. 

\paragraph{Safety-constrained tumor assessment using retrospectively locked thresholds.}
\label{methods:safety_constrained_tumor}
To evaluate whether GRACE could speed up malignancy diagnosis, reduce secondary-review workload, and prioritize IHC, we defined safety-constrained rule-out and rule-in analysis for tumor assessment. Thresholds were derived exclusively from the internal retrospective cohort and were selected to achieve 100\% NPV for rule-out and 100\% PPV for rule-in. These thresholds were locked before prospective validation and then applied unchanged to the prospective cohort. Cases falling within the locked rule-out or rule-in regions were considered actionable only if the corresponding prospective subset retained 100\% NPV and/or 100\% PPV. All remaining cases were considered non-actionable and would be retained for standard diagnostic review. Results are summarized in Extended Data Table~\ref{tab:lock_t}.

\subsubsection*{Randomized crossover reader study for pathologist-AI collaboration assessment}
To determine whether GRACE improves diagnostic decision-making, we conducted a prospective, randomized crossover reader study in gastric pathology. Pathologists interpreted cases both independently and with GRACE assistance, enabling direct assessment of pathologist-AI collaboration across diagnostic accuracy, efficiency, confidence, and interobserver agreement. The human subjects review protocol was approved by the Medical Ethics Committee of Nanfang Hospital, Southern Medical University (approval No. NFEC-2025-653) on December 5, 2025. The protocol was registered on ClinicalTrials.gov (ID: NCT07291362).

\paragraph{Participants and randomization.}
Eight eligible pathologists participated in the prospective, randomized, two-period crossover reader study. The enrollment and analysis flow of the reader study are shown in Extended Data Fig.~\ref{fig:consort}a, following CONSORT reporting guidelines\cite{turner2012consolidated}. Participants were stratified by subspecialty experience level as junior ($\leq$5 years, $n=4$) or senior ($>5$ years, $n=4$), and then randomized 1:1 to one of two reading sequences, ensuring that each sequence included two junior and two senior pathologists. Group A (\(n=4\)) interpreted cases without AI assistance in Period 1, followed by a 4-week washout period and GRACE-assisted diagnosis in Period 2. Group B (\(n=4\)) completed the reverse sequence, beginning with GRACE-assisted diagnosis in Period 1, followed by the same 4-week washout period and unassisted diagnosis in Period 2. All randomized pathologists received the assigned sequence, completed both study periods and the washout interval, and were included in the primary analysis. No participants withdrew, and no data were excluded because of protocol deviations. All participants provided written informed consent and were blinded to the study hypotheses and aggregate outcomes.

\paragraph{Diagnostic tasks and case selection.}
Pathologist performance was evaluated across three clinically challenging and consequential diagnostic tasks relevant to diagnosis of gastric mucosal lesions. Reference-standard labels were established before analysis through consensus review. Cases were selected by stratified random sampling so that the distribution of diagnostic categories matched those of the prospective validation cohorts. For each diagnostic task, 100 cases were randomly selected, each with a unique WSI. All performance evaluations were conducted at the case level. 

The evaluated tasks comprised: (1) grading of acute inflammatory activity in patients with chronic gastritis, categorized as mild (70 cases), moderate (28 cases), and severe (2 cases) grades; (2) diagnosis of chronic atrophic gastritis, classified as non-atrophic (46 cases) or atrophic (54 cases); and (3) diagnosis of IM, classified as non-IM (68 cases) or IM (32 cases) (Extended Data Table~\ref{tab:RCT_data}). 
For reader-study sensitivity, specificity, PPV, and NPV, positive classes were prespecified as moderate/severe acute activity versus mild acute activity, atrophy versus non-atrophy, and IM versus non-IM.

\paragraph{Study procedures and reading protocol.}
Pathologists reviewed the WSIs using SmartPath (\url{https://smartpathology.top/}), a web-based WSI review and scoring platform designed to approximate routine digital pathology review while enabling controlled presentation of GRACE decision support. SmartPath included a case-management home page, an interactive WSI review workspace, and a structured scoring panel for recording diagnostic assessments. The review workspace supported standard digital pathology functions, including slide navigation, zooming, panning, and annotation, with no time limits imposed during case review (Extended Data Fig.~\ref{fig:consort}b, c).

After WSI upload, SmartPath automatically performed tissue detection, patch-level tiling, and inference using the GRACE model, enabling standardized processing. During the unassisted period, pathologists reviewed cases without access to algorithmic outputs. During the GRACE-assisted period, the same viewing interface additionally displayed the GRACE-predicted category, class-specific prediction probabilities, and attention-based heatmaps (Extended Data Fig.~\ref{fig:consort}c).

Before the formal study, all participating pathologists completed hands-on training covering the study workflow, task-specific label sets, and how GRACE outputs would be displayed in the interface. They were informed that GRACE outputs were intended for decision support, could be correct or incorrect, and would not be accompanied by feedback on prediction correctness during case review. Case order was randomized within each period to minimize ordering effects, and final diagnostic responsibility remained with the reader. For each case, pathologists selected a final diagnosis from predefined task-specific categories and reported diagnostic confidence on a 10-point scale, ranging from 1 (minimal confidence) to 10 (maximal confidence). Reading time was automatically recorded from slide opening to case submission.

\paragraph{Robustness checks: order effects in the crossover design.}
To evaluate whether AI-assisted diagnostic performance differed by randomized reading sequence, we fitted a population-averaged logistic regression model using GEE with an exchangeable correlation structure clustered by pathologist.
Let $Y_{ir}$ denote the AI-assisted diagnostic correctness for case $i$ evaluated by reader $r$. 
The sequence indicator $\text{Group}_{r}$ encoded whether reader $r$ received GRACE assistance in Period 1 or Period 2.
The diagnostic task was denoted by $\text{Task}_{i}$, and pathologist experience level was denoted by $\text{Level}_{r}$. 
The primary term of interest was $\text{Group}_{r}$. Covariates $\text{Task}_{i}$ and $\text{Level}_{r}$ were included to adjust for variation in task difficulty and pathologist experience. Interaction terms $\text{Group}_{r} \times \text{Task}_{i}$ and $\text{Group}_{r} \times \text{Level}_{r}$ were then added to assess whether any sequence effect varied by diagnostic task or pathologist experience level.

\begin{equation}
\text{logit}\!\left\{ \Pr\!\left( Y_{ir} = 1 \right) \right\}
=
\beta_0
+ \beta_1 \,\text{Group}_{r}
+ \beta_2 \,\text{Task}_{i}
+ \beta_3 \,\text{Level}_{r}
+ \beta_4 \,(\text{Group}_{r} \times \text{Task}_{i})
+ \beta_5 \,(\text{Group}_{r} \times \text{Level}_{r})
\end{equation}

As shown in Extended Data Table~\ref{tab:crossover_regression}, no statistically significant association was observed between reading sequence and AI-assisted diagnostic correctness (P = 0.298). In addition, no significant interactions were detected between sequence and diagnostic task (all $P \geq 0.674$) or between sequence and reader experience level ($P = 0.216$). These results provided no evidence that AI-assisted diagnostic performance differed according to whether GRACE assistance was received in Period 1 or Period 2. However, because reading sequence was part of the randomized crossover design, $\text{Group}_{r}$ was still included as an adjustment variable in the primary analyses.

\paragraph{Diagnostic performance and evaluation metrics.}
Diagnostic performance metrics included accuracy, sensitivity, specificity, NPV, and PPV (Extended Data Tables~\ref{tab:overall_ai_vs_noai} and \ref{tab:performance}). Unadjusted estimates were reported separately for the unassisted and AI-assisted conditions. Uncertainty was assessed using nonparametric bootstrap resampling at the case-slide-task level (1,000 resamples). 

Adjusted effects of AI assistance were estimated using GEEs with an exchangeable working correlation structure and pathologist as the grouping variable (Extended Data Table~\ref{tab:gee_bootstrap_all}). Accuracy, sensitivity, specificity, PPV, and NPV were modeled using binomial logit GEEs and reported as adjusted ORs. Overall models adjusted for reader experience level, crossover sequence, and diagnostic task. Task-stratified and experience-stratified models omitted the corresponding stratification variable. Adjusted 95\% CIs and two-sided bootstrap $P$ values were obtained by refitting the same GEE model in each bootstrap resample.

\paragraph{Interpretation efficiency and diagnostic confidence.}
Diagnostic confidence was scored on a 1--10 scale and summarized using the mean (Extended Data Table~\ref{tab:confidence}). Interpretation time was quantified as reading time per case in seconds. Descriptive interpretation time was summarized using the median, mean, and geometric mean (Extended Data Table~\ref{tab:time}). 

Diagnostic confidence scores were analyzed using Gaussian GEE models and reported as adjusted mean differences, with positive values indicating higher confidence with AI assistance. 
Log-transformed interpretation time was analyzed using Gaussian GEE models, and exponentiated coefficients are reported as adjusted geometric mean time ratios, with values below 1 indicating shorter reading time with AI assistance. 

\paragraph{Inter-rater agreement.}
Inter-rater agreement was assessed using Fleiss' $\kappa$, calculated separately for unassisted and AI-assisted diagnoses at the task-specific case-slide level. Agreement was summarized overall, by diagnostic task, and by reader experience level (Extended Data Table~\ref{tab:fleiss}).
$\kappa$ values were interpreted according to the Landis and Koch criteria\cite{landis1977measurement}: slight (0.00-0.20), fair (0.21-0.40), moderate (0.41-0.60), substantial (0.61-0.80), and almost perfect (0.81-1.00) agreement.

\paragraph{AI impact categorization.}
\label{methods:ai_impact_categorization}
To characterize how AI assistance influences pathologist decision-making, we developed an AI impact analysis that explicitly models the interaction between human judgment and AI assistance. For each case, a category was assigned considering three variables: (i) the correctness of the unassisted pathologist diagnosis ($s$), (ii) the correctness of the AI model prediction ($m$), and (iii) the correctness of the assisted diagnosis ($a$). This formulation yields five mutually exclusive AI impact categories (Extended Data Table~\ref{tab:ai_impact_categories}).

\textbf{AI correction} was defined as cases in which the unassisted human diagnosis was incorrect ($s=0$), the AI prediction was correct ($m=1$), and the AI-assisted diagnosis was correct ($a=1$).
\textbf{AI reinforcement} denoted cases in which the unassisted human diagnosis, AI prediction, and AI-assisted diagnosis were all correct ($s=1, m=1, a=1$).
\textbf{Human safeguard} comprised cases in which the AI-assisted diagnosis was correct despite an incorrect AI prediction, including: (i) a correct unassisted human diagnosis with an incorrect AI prediction ($s=1, m=0, a=1$), or (ii) an incorrect unassisted human diagnosis with an incorrect AI prediction but correct AI-assisted diagnosis ($s=0, m=0, a=1$).
\textbf{AI distraction} denoted cases in which the unassisted human diagnosis was correct but the AI-assisted diagnosis was incorrect, including: (i) an incorrect AI prediction ($s=1, m=0, a=0$); or (ii) a correct AI prediction ($s=1, m=1, a=0$).
\textbf{Shared failure} denoted cases in which both the unassisted human diagnosis and AI-assisted diagnosis were incorrect ($s=0, a=0$), including cases where the AI prediction was also incorrect ($m=0$) and cases where the AI prediction was correct ($m=1$).

\paragraph{Pathologist-referenced review triage analysis.}
\label{methods:expert_referenced_review_triage}
We further performed a pathologist-referenced triage analysis for two binary tasks: atrophic chronic gastritis diagnosis (\(N=474\), positive rate = 0.502) and intestinal metaplasia diagnosis (\(N=479\), positive rate = 0.413). Unlike the retrospectively locked tumor-assessment thresholds, this analysis used senior pathologist performance in the reader study as the safety benchmark. Senior pathologist NPV and PPV were estimated by case-level cluster bootstrap resampling with 1,000 iterations, retaining all senior-reader observations for each sampled case. The upper bounds of the resulting 95\% CIs were used as task-specific benchmark values.

For each task, paired GRACE rule-out and rule-in thresholds \((t_{\mathrm{NPV}}, t_{\mathrm{PPV}})\) were evaluated. A threshold pair was considered acceptable only if the bootstrap mean GRACE rule-out NPV and rule-in PPV met or exceeded the corresponding senior-pathologist benchmark values. Among acceptable threshold pairs, the operating point maximizing the proportion of cases assigned to either high-confidence region was selected and reported as the triage-eligible fraction under the expert-referenced safety criterion.

\paragraph{Task-specific correction of mucosal diagnostic errors.}
To move beyond aggregate reader-study accuracy, we examined cases in which initially incorrect unassisted diagnoses became correct after GRACE assistance. This analysis allowed us to identify the specific error patterns that were most responsive to AI-assisted revision, with the true label shown against the original unassisted diagnosis in Extended Data Table~\ref{tab:task_true_label_predicted}.
The corrected errors showed clear task-specific structure. In acute activity grading, revisions most often involved movement between adjacent severity categories, consistent with GRACE helping readers resolve borderline inflammatory grades. In atrophy diagnosis, corrected cases included both initially missed atrophy and initial overcalls, suggesting that AI assistance reduced inconsistency in assessing glandular loss. By contrast, in IM diagnosis, corrected errors were dominated by initially positive calls in truly negative cases, indicating that GRACE primarily reduced unnecessary IM overcalls while still correcting some missed lesions.

Together, these findings show that GRACE improved reader performance not simply by increasing aggregate accuracy, but by addressing distinct sources of mucosal diagnostic error: grade-boundary ambiguity in acute activity, inconsistent atrophy assessment and IM overcalling. This corrected-error analysis provides interpretable evidence for how AI assistance changed reader decisions in clinically challenging mucosal diagnoses.
These findings suggest that future gastric pathology AI systems should be evaluated not only by standalone performance, but also by their ability to deliver task-specific, interpretable correction in clinically challenging diagnostic scenarios.

\subsubsection*{Metrics and statistical analysis}
\paragraph{Performance metrics for binary classification tasks.}
For binary tasks, model discrimination was assessed using the area under the receiver operating characteristic curve (AUROC)\cite{nahm2022receiver} with corresponding 95\% confidence intervals (CIs)\cite{bland1996transformations}. Classification accuracy was summarized using macro-accuracy (macro-ACC). Default-threshold classification metrics, including confusion matrices, positive predictive value (PPV), and negative predictive value (NPV) were calculated from binary class assignments using a positive-class probability threshold of 0.5. Receiver operating characteristic (ROC) curves\cite{zweig1993receiver} were generated by sweeping the decision threshold over the predicted probability of the positive class. 

To summarize performance under a high-sensitivity operating condition, we also reported specificity at 90\% sensitivity\cite{lalkhen2008clinical}. This quantity was estimated by interpolation along the ROC curve, where specificity corresponds to $1-\mathrm{FPR}$ at a sensitivity level of 90\%. When confusion matrices were shown with ROC panels, they were generated separately at the highest discrete score threshold that achieved at least 90\% empirical sensitivity.

\paragraph{Performance metrics for multi-class classification tasks.}
For multiclass classification tasks, discrimination was evaluated using the macro-averaged area under the receiver operating characteristic curve (macro-AUC), calculated in a one-versus-rest manner. Classification accuracy was summarized using macro-ACC. 
For this analysis, one-versus-rest ROC curves were computed for each class, class-specific specificities were identified at the operating point achieving at least 90\% sensitivity, and the resulting specificities were macro-averaged across classes.

\paragraph{Positive-class definitions.}
For binary and one-versus-rest analyses, the positive classes were prespecified as follows: ACGHP-positive status for autoimmune chronic gastritis with \textit{H. pylori} diagnosis; HPACG-positive status for \textit{H. pylori}-associated chronic gastritis diagnosis; CG-positive status for chronic gastritis diagnosis; moderate/severe CG for chronic gastritis severity grading; acute-activity-positive status for acute activity diagnosis; moderate/severe acute activity for acute activity severity grading; atrophy-positive status for atrophy diagnosis; moderate atrophy for atrophy severity grading; ulcer-positive status for ulcer diagnosis; polyp-positive status for polyp diagnosis; IM-positive status for intestinal metaplasia diagnosis; malignancy-positive status for malignancy diagnosis; SRC-positive status for pathology subtyping; intestinal type for Lauren classification; G3 for histologic differentiation grading; T3/T4 for T-staging; N+ for N-staging; PNI-positive status for perineural invasion; VI-positive status for vascular invasion; high residual tumor for tumor regression grade; HER2 3+ for HER2 expression prediction; dMMR for MMR prediction; high Ki-67 expression for Ki-67 prediction; and EBV-positive status for EBV prediction.

\paragraph{Uncertainty estimation.}
Across all tasks, 95\% CIs were estimated using nonparametric bootstrap resampling with 1{,}000 replicates. In each replicate, case-level prediction-label pairs were resampled with replacement and all metrics were recomputed to generate empirical distributions. 95\% CIs were defined by the 2.5th and 97.5th percentiles. For visualization, bar plots display the bootstrap mean macro-AUC for each task, with error bars indicating one standard deviation of the bootstrap distribution.

\paragraph{Paired statistical testing for within-task model comparison.}
To assess within-task statistical significance, we performed stratified paired bootstrap resampling at the patient level. The same bootstrap resamples were used for both models, yielding paired estimates of each evaluated metric. For each task and metric, a one-sided Wilcoxon signed-rank test\cite{wilcoxon1992individual} was applied to the paired bootstrap estimates to test whether the target model exceeded the baseline model. The resulting pairwise comparisons are reported in Extended Data Figs.~\ref{fig:bar-1}, \ref{fig:bar-2}, \ref{fig:bar-3}.

\paragraph{Cross-task comparative analysis and model ranking.} For cross-task comparative analyses, the statistical unit was the task-cohort combination. 
For each task-cohort combination, one cohort-level AUC estimate was calculated for each model, and paired comparisons were performed across the same set of task-cohort combinations. 
We first compared GRACE and the comparator models jointly using the Friedman test, treating task-cohort combinations as repeated blocks. Friedman results were reported as $\chi^2$ statistics with 3 degrees of freedom, paired-task sample size $n$, exact \textit{P} values and Kendall's $W$, where $W$ quantified concordance of model rankings across tasks from 0 (no rank agreement) to 1 (perfect rank agreement). The results were summarized in Extended Data Table~\ref{tab:friedman_results}.
After a significant Friedman test, paired one-sided Wilcoxon signed-rank tests\cite{wilcoxon1992individual} were applied across task-cohort combinations to compare GRACE with each comparator model. The resulting \textit{P} values were adjusted using Holm's method across the comparisons, with significance defined as Holm-adjusted $P < 0.001$. The results were summarized in Extended Data Table~\ref{tab:wilcoxon_results}.

We additionally summarized overall FM performance using average ranks across task-cohort combinations. Within each task-cohort combination, rank 1 was assigned to the highest AUC and rank 4 to the lowest AUC. Across 72 task-cohort combinations, GRACE achieved the best mean rank (1.03), followed by Virchow2 (2.57), UNI (3.10), and CONCH (3.31). Given a critical difference of 0.5528 at $\alpha=0.05$, the rank separation between GRACE and each comparator exceeded the significance threshold, indicating that GRACE significantly outperformed all baseline models\cite{demvsar2006statistical,antonelli2022medical}. The results were summarized in Fig.~\ref{Fig:performance}c.

\paragraph{Survival prediction definitions and evaluations.} 
For survival analyses, DFS was defined as the interval from the date of surgical resection to the first documented disease recurrence or death from any cause. Patients without documented recurrence or death were censored at the date of last disease-free assessment. OS was defined as the interval from the date of diagnosis to death from any cause. Patients alive at last follow-up were censored at the date they were last known to be alive.

Survival prediction was performed using an ABMIL-based model trained on surgically resected WSI features under a patient-level five-fold cross-validation framework. Model outputs were interpreted as discrete-time hazard estimates and converted into survival probabilities. A continuous risk score was computed as the negative sum of predicted survival probabilities across the model-defined intervals, with higher values indicating higher predicted risk.
Internal performance was evaluated using out-of-fold predictions. Each patient contributed one risk score from the model trained without that patient. Fold-wise C-indices were computed using comparable patient pairs under right censoring and summarized across folds. CIs were estimated by patient-level bootstrap resampling of fixed validation predictions.
External validation did not involve model retraining or fine-tuning on the external cohort. Each external patient was evaluated independently by the five fold-specific survival models trained during internal cross-validation, yielding five continuous risk scores per patient. For discrimination analysis, the five risk scores were averaged to obtain a continuous ensemble risk score, which was used to calculate the external C-index. CIs were estimated by patient-level bootstrap resampling of the fixed external predictions with 1,000 iterations.

Kaplan-Meier curves were used to visualize survival by consensus risk group. The five model-derived binary risk assignments were combined by majority vote: patients classified as high risk by at least three of the five models were assigned to the final high-risk group, and the remaining patients were assigned to the final low-risk group. Survival differences between groups were assessed using the log-rank test.

\paragraph{Modeling AI-associated correction and diagnostic harm conditional on initial human judgment.}
To examine whether AI-associated outcome transitions varied by diagnostic task or reader characteristics, we fitted two exploratory conditional GEE logistic regression models with an exchangeable working correlation structure clustered by pathologist. Analyses were performed at the reader-case level.

The AI-associated correction model was restricted to observations in which the initial human diagnosis was incorrect \((Y_i^{(0)}=0)\), and estimated the probability that the observation was classified as \textbf{AI correction}:
\[
\Pr\!\left(y_i^{(C)}=1 \mid Y_i^{(0)}=0, \mathbf{X}_i\right)
=
\operatorname{logit}^{-1}\!\left(\beta_0^{(C)}+\boldsymbol{\beta}^{(C)\top}\mathbf{X}_i\right).
\]

The AI-associated diagnostic harm model was restricted to observations in which the initial human diagnosis was correct \((Y_i^{(0)}=1)\), and estimated the probability that the observation resulted in an incorrect AI-assisted diagnosis:
\[
\Pr\!\left(y_i^{(D)}=1 \mid Y_i^{(0)}=1, \mathbf{X}_i\right)
=
\operatorname{logit}^{-1}\!\left(\beta_0^{(D)}+\boldsymbol{\beta}^{(D)\top}\mathbf{X}_i\right).
\]

In both models, \(\mathbf{X}_i\) included diagnostic task, pathologist experience level, and crossover sequence group. Diagnostic task was encoded using indicator variables, with acute activity severity grading as the reference category. Effect sizes are reported as odds ratios with 95\% CIs (Extended Data Table~\ref{tab:transition_models}).

In the AI-associated correction model, no clear associations were observed for reader experience, crossover sequence, or diagnostic task, suggesting that AI-associated correction among initially incorrect diagnoses was not confined to a specific reader group or diagnostic context. In the AI-associated diagnostic harm model, senior pathologist experience was associated with a lower probability of diagnostic harm among initially correct diagnoses (OR = 0.36, $P=0.015$). Together, these exploratory models suggest that AI-associated correction was broadly distributed across reader and task strata, whereas diagnostic harm was not task-specific and was less likely among senior readers.

\subsubsection*{Computing hardware and software}
GRACE pretraining was performed on one compute node, equipped with 8 NVIDIA H800 GPUs with 80 GB memory. The released pretraining environment used Python 3.10, PyTorch\cite{paszke2019pytorch} v2.5.1 with CUDA 12.1, torchvision v0.20.1, timm\cite{rw2019timm} v1.0.24, PEFT v0.16.0, loralib, h5py v3.15.1, NumPy\cite{harris2020array} v2.2.6, and Pillow v12.1.0. 
WSI preprocessing, patch extraction, HDF5 dataset generation, and downstream feature extraction were performed using the PrePATH workflow \url{https://github.com/birkhoffkiki/PrePATH}.
Downstream classification was run with PyTorch v2.6.0 with CUDA 11.8, NumPy, pandas\cite{mckinney2010data}, openpyxl, tqdm, torchmetrics\cite{detlefsen2022torchmetrics}, and TensorBoardX. Survival prediction used PyTorch, NumPy, pandas, tqdm, openpyxl, and scikit-survival for concordance-index computation.

\section*{Code and data availability}
The source code developed for this work will be made publicly available upon acceptance. Publicly available datasets used in this study include the HistAI\cite{nechaev2025histai} (\url{https://github.com/HistAI/HISTAI}), The Cancer
Genome Atlas (TCGA) (\url{https://www.cancer.gov/tcga}), Comprehensive Assessment of Chronic Gastritis on WSI Data\cite{xia2025gastritismil}(\url{https://www.scidb.cn/detail?dataSetId=83ee1521074742cdaae997cf0b46a7b1}), Gastric Cancer Lymph Node Dataset\cite{Chen2020}(\url{https://figshare.com/articles/dataset/Gastric_cancer_lymph_node_data_set/13065986}), and KBSMC Gastric Cancer Grading Dataset (\url{https://github.com/colin19950703/KBSMC_gastric_cancer_grading_dataset}). These datasets are publicly accessible through their respective repositories or publications.
WSIs and clinical data from the proprietary cohorts used for model development and evaluation are not publicly available owing to institutional policies, data-use agreements, and patient privacy regulations. Access for non-commercial academic research may be requested from the corresponding author and will require institutional approval and appropriate data-sharing agreements.

\section*{Contributions}
    Ling L. conceived the study and developed its methodology; designed, executed, and analyzed the experiments; carried out gastric data cleaning, preprocessing, feature extraction; and prepared the manuscript, code, and supplementary materials.
    J.M. conceived the study and developed its methodology, designed the experiments, prepared the manuscript, and preprocessed the gastric pathology data.  
    Z.Z. collected and curated the gastric pathology datasets, coordinated prospective validation studies and the reader study, and provided clinical annotation and pathological expertise.
    F.Z. conceived the methodology, contributed to the study design, preprocessed and proposed the analysis of the gastric transcriptomic data. 
    Y.X. and Yihui W. contributed to the study design, preprocessed the gastric data, and extracted features from whole-slide images.
    C.J. and C.L. developed an online digital pathology image review and scoring system for the reader study.
    Z.G., Y.C., and H.W. participated in discussions on benchmark development and contributed to the study design.
    O.K.T., Zhen W., J.Z., S.S., S.L., Yu W., Z.L., R.C.K.C., X.Z., and Zhe W. provided the gastric pathology datasets and oversaw the data collection and quality control processes.
    Z.C., Q.X., C.Z., F.W., and Y.Y. participated in the reader study as pathologists and collected the gastric pathology datasets.
    H.C. supervised the research and the development of the AI model, coordinated data collection and provided expertise for task design.
    Li L. supervised the research, coordinated prospective validation studies and the reader study, provided clinical annotation and pathological expertise for task design.

\section*{Acknowledgement}
    This work was supported by the Chongqing Technology innovation and application development special major project (CSTB2024TIAD-STX0003), Shenzhen Science and Technology Innovation Committee Fund (Project No.~\seqsplit{KCXFZ20230731094059008}), Innovation and Technology Commission (Project No. MHP/002/22 and ITCPD/17-9), Research Grants Council of the Hong Kong Special Administrative Region, China (Project No. T45-401/22-N, R6003-22 and C4024-22GF), Frontier Technology Research for Joint Institutes with Industry Scheme (Project No. OKT24EG01), and National Key R\&D Program of China (Project No. 2023YFE0204000). 
    We thank HKUST SuperPod for providing the GPU platform for foundation model pretraining. We thank Ying Tan and Weihao Qiu for their participation as pathologists in the randomized crossover reader study.

\section*{Ethics declarations}
    This project has been reviewed and approved by the Human and Artifacts Research Ethics Committee (HAREC) of Hong Kong University of Science and Technology (protocol no. HREP-2025-0341). The prospective study protocol was approved by the Medical Ethics Committee of Nanfang Hospital, Southern Medical University (protocol no. NFEC-2025-403). The corresponding protocol was registered on ClinicalTrials.gov (ID: NCT07157618). The randomized crossover reader study was approved by the Medical Ethics Committee of Nanfang Hospital, Southern Medical University (approval No. NFEC-2025-653). The clinical application protocol of the randomized crossover reader study was registered on ClinicalTrials.gov (ID: NCT07291362).

\FloatBarrier

\begingroup
\raggedbottom
\bibliography{reference_updated}
\endgroup

\FloatBarrier
\section*{Extended data}
\setcounter{figure}{0}
\renewcommand{\figurename}{Extended Data Figure}
\setcounter{table}{0}
\renewcommand{\tablename}{Extended Data Table}

\begin{figure}[H]
    \centering
    \includegraphics[width=0.9\textwidth]{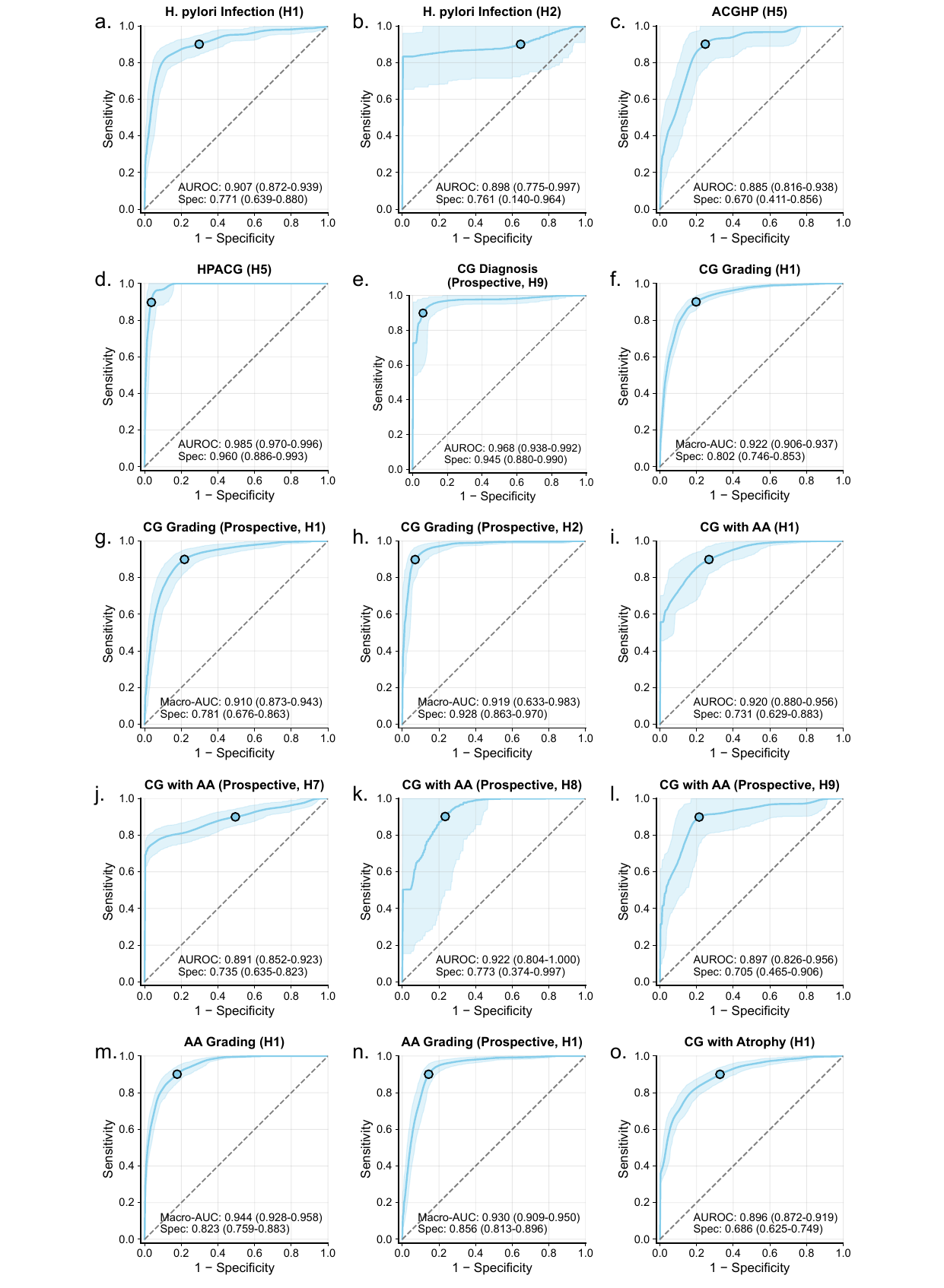}
    \caption{\textbf{Receiver operating characteristic (ROC) curves for gastric pathology tasks.}
    (\textbf{a}-\textbf{o}) ROC curves are shown with 95\% bootstrap confidence intervals (CIs). The legends report area under the ROC curve (AUROC) with 95\% CI for binary classification tasks, macro-averaged AUC (macro-AUC) with 95\% CI for multiclass classification tasks, and specificity at 90\% sensitivity (95\% CI).}
    \label{fig:AUROC-1}
\end{figure}

\FloatBarrier
\begin{figure}
    \centering
    \includegraphics[width=0.9\textwidth]{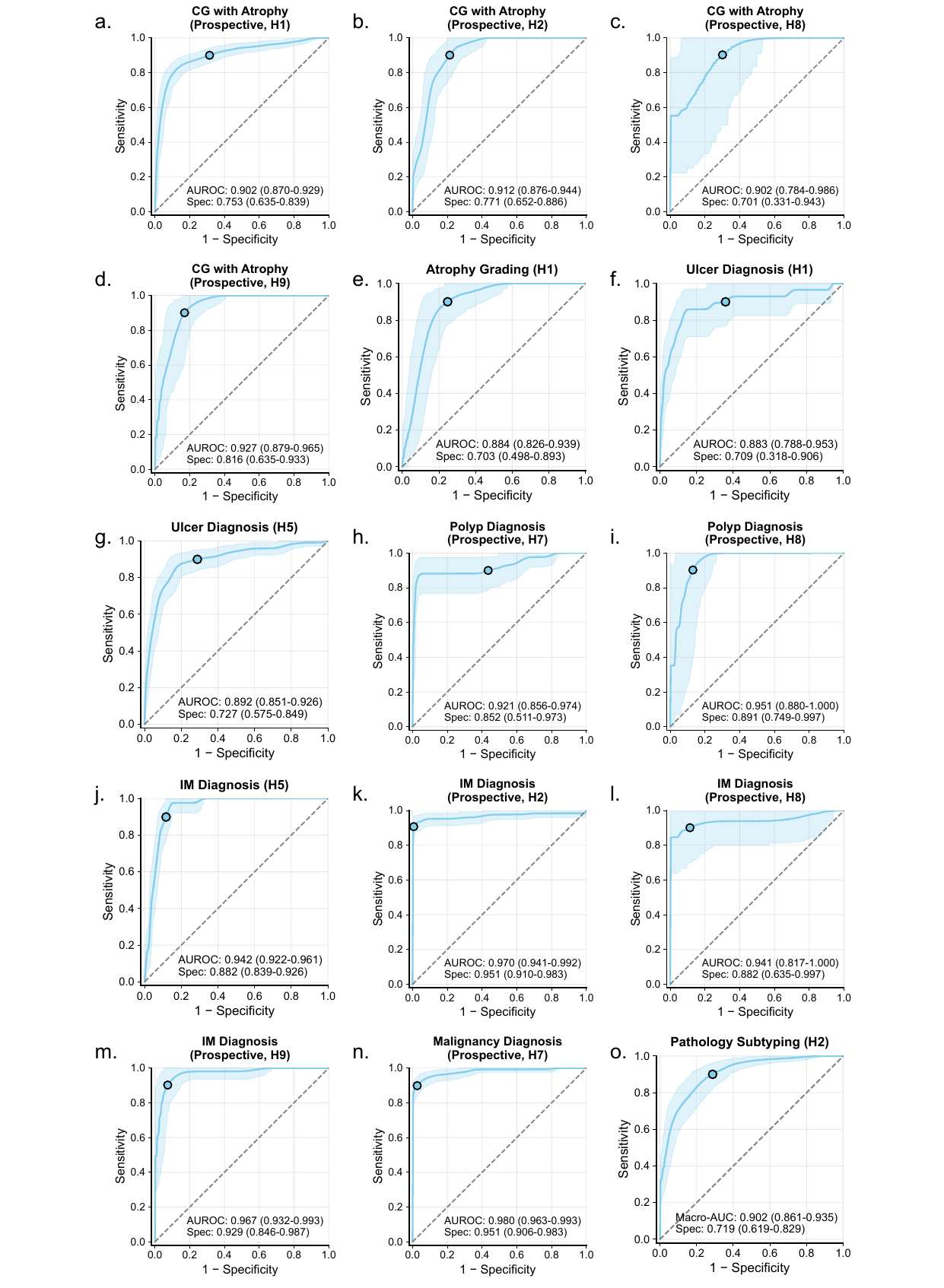}
    \caption{\textbf{ROC curves for gastric pathology tasks (continued).}
        (\textbf{a}-\textbf{o}) ROC curves are shown with 95\% bootstrap CIs. The legends report AUROC (95\% CI) for binary classification tasks, macro-AUC (95\% CI) for multiclass classification tasks, and specificity at 90\% sensitivity (95\% CI). }
    \label{fig:AUROC-2}
\end{figure}

\FloatBarrier
\begin{figure}
    \centering
    \includegraphics[width=0.9\textwidth]{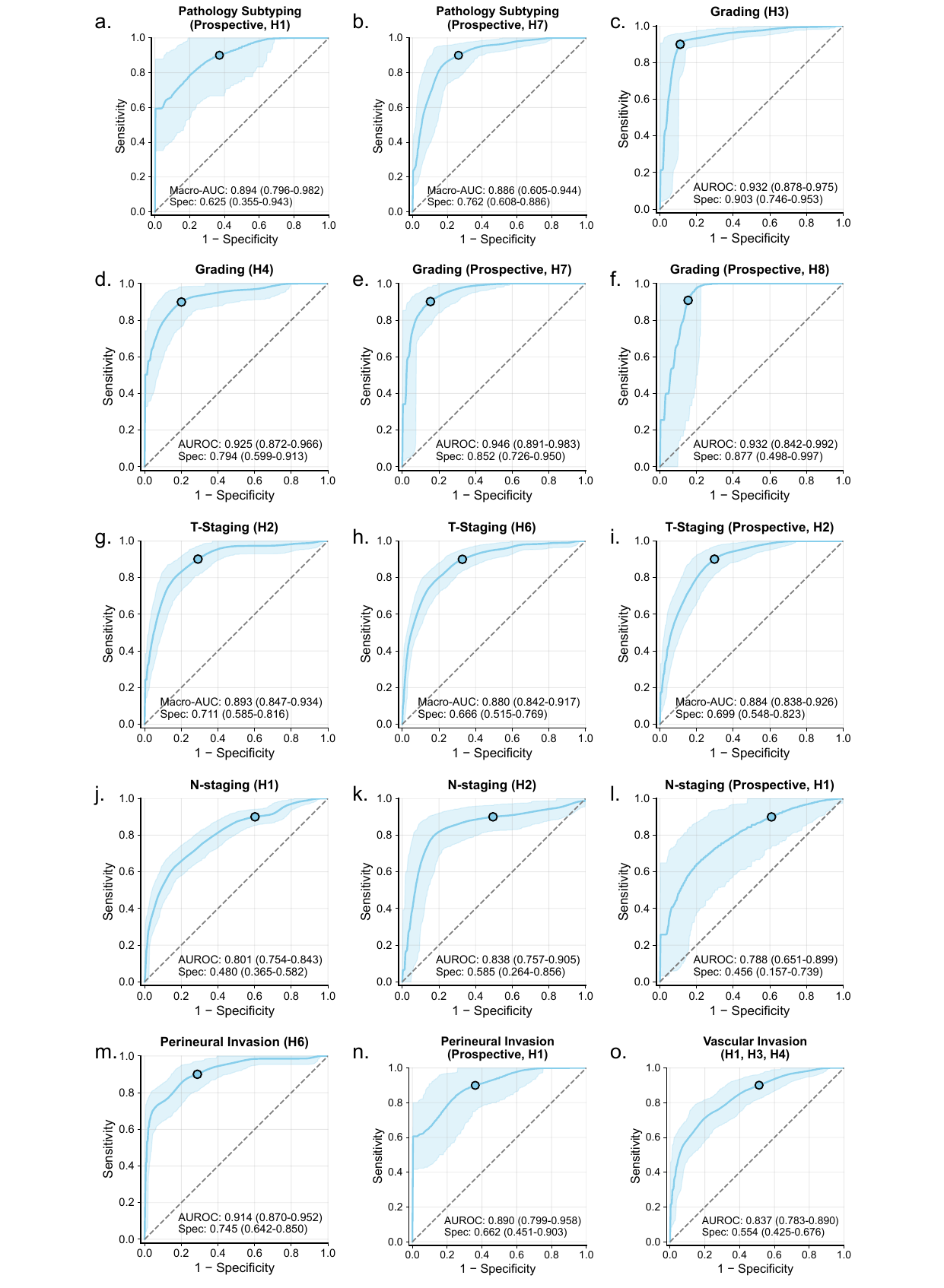}
    \caption{\textbf{ROC curves for gastric pathology tasks (continued).}
        (\textbf{a}-\textbf{o}) ROC curves are shown with 95\% bootstrap CIs. The legends report AUROC (95\% CI) for binary classification tasks, macro-AUC (95\% CI) for multiclass classification tasks, and specificity at 90\% sensitivity (95\% CI). }
    \label{fig:AUROC-3}
\end{figure}

\FloatBarrier

\begin{figure}
    \centering
    \includegraphics[width=0.9\textwidth]{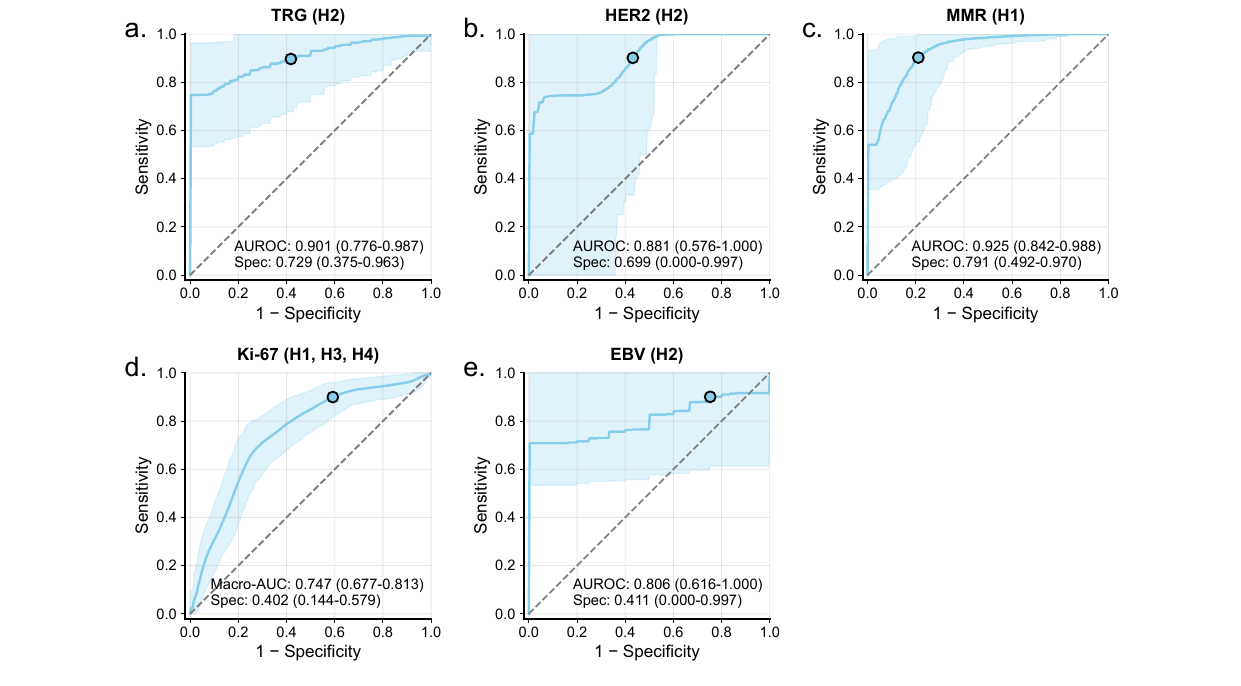}
    \caption{\textbf{ROC curves for gastric pathology tasks (continued).}
        (\textbf{a}-\textbf{e}) ROC curves are shown with 95\% bootstrap CIs. The legends report AUROC (95\% CI) for binary classification tasks, macro-AUC (95\% CI) for multiclass classification tasks, and specificity at 90\% sensitivity (95\% CI). }
    \label{fig:AUROC-4}
\end{figure}

\FloatBarrier
\begin{figure}
    \centering
    \includegraphics[width=\textwidth]{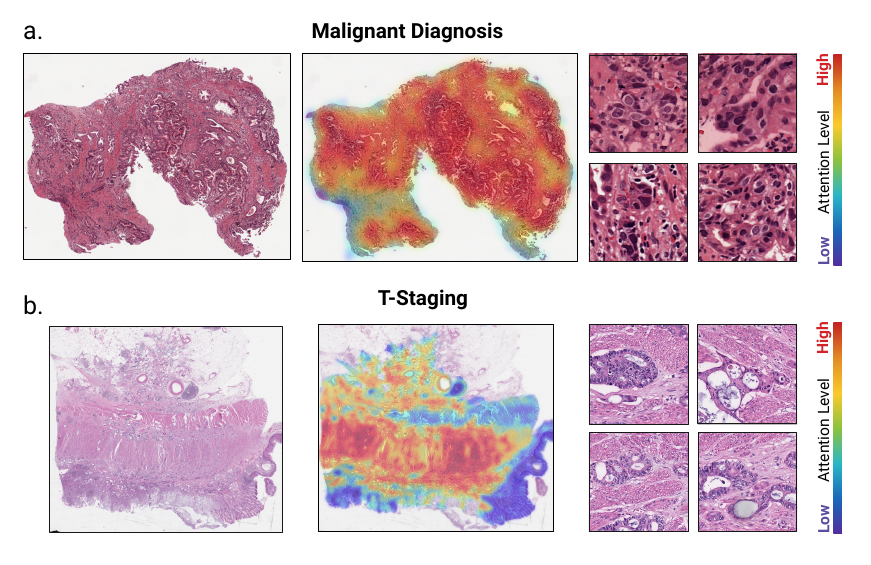}
    \caption{\textbf{Heatmap visualization of GRACE.}
    (\textbf{a}-\textbf{b}) Whole-slide image (WSI) attention-based heatmaps with top-ranked patches illustrating regions contributing most to predictions for malignant diagnosis (\textbf{a}) and T-staging (\textbf{b}). }
    
    \label{fig:heatmap}
\end{figure}

\FloatBarrier
\begin{figure}[!htbp]
    \centering
    \includegraphics[width=\textwidth]{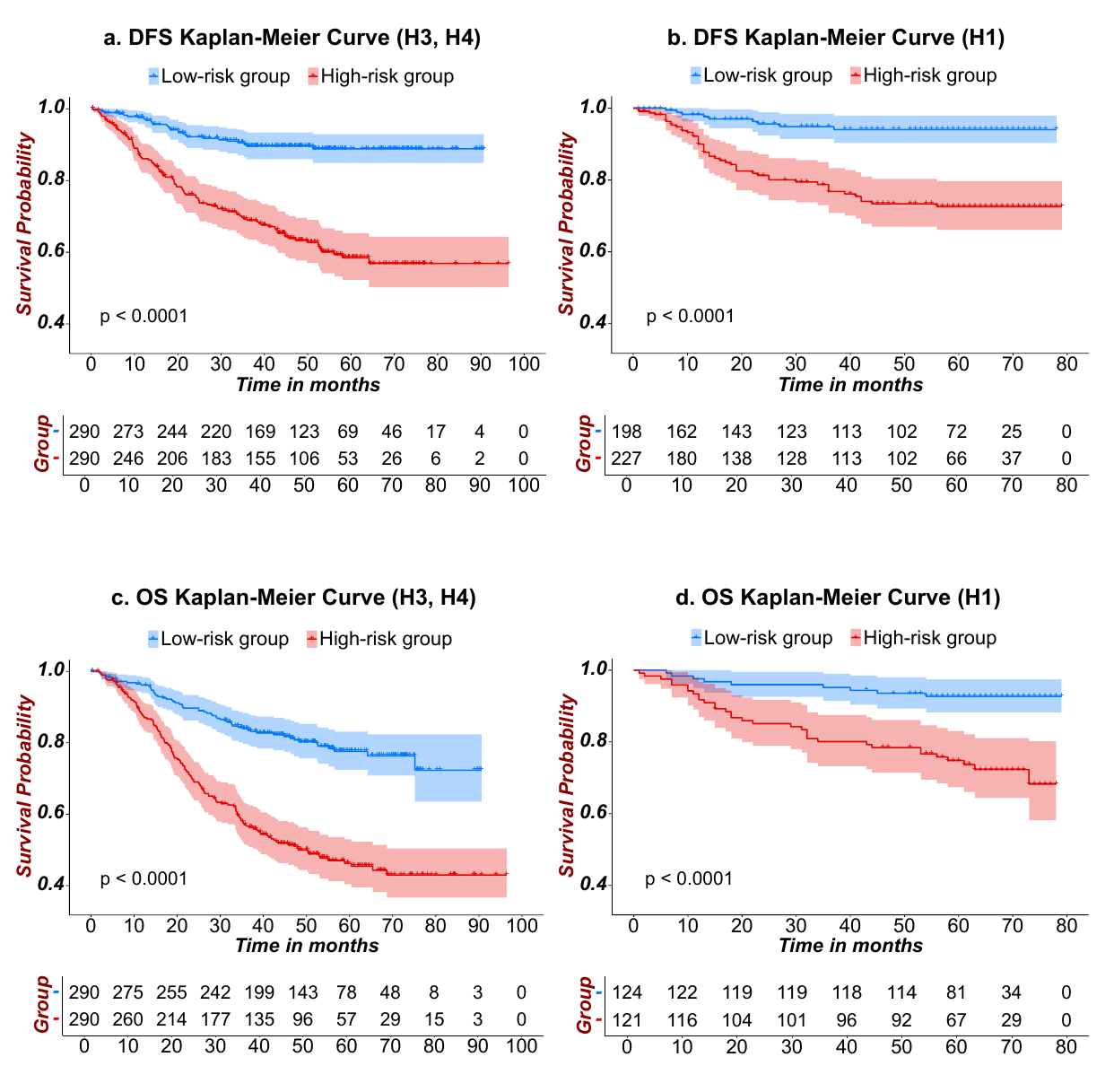}
    \caption{\textbf{Performance of GRACE on survival prediction tasks.} (\textbf{a}-\textbf{b}) Kaplan-Meier curves for disease-free survival (DFS) prediction. (\textbf{c}-\textbf{d}) Kaplan-Meier curves for overall survival (OS) prediction.}
    \label{fig:KM}
\end{figure}

\begin{table}[t]
\centering
\footnotesize
\setlength{\tabcolsep}{1.6pt}
\renewcommand{\arraystretch}{1.25}
\begin{threeparttable}
\captionsetup{justification=raggedright,singlelinecheck=false,width=\linewidth}
\caption{\textbf{Safety-constrained tumor assessment using retrospectively locked thresholds.} $T_{\mathrm{out}}$ refers to the threshold for rule-out, and $T_{\mathrm{in}}$ refers to the threshold for rule-in.}
\label{tab:lock_t}
\begin{tabular}{p{4.2cm} c c c c c c c c c}
\toprule
\textbf{Tasks} & \textbf{$T_{\mathrm{out}}$} & \textbf{\shortstack{Low-threshold\\NPV}} & \textbf{\shortstack{Low-threshold\\Coverage}} & \textbf{\shortstack{Low-threshold\\Cases}} & \textbf{$T_{\mathrm{in}}$} & \textbf{\shortstack{High-threshold\\PPV}} & \textbf{\shortstack{High-threshold\\Coverage}} & \textbf{\shortstack{High-threshold\\Cases}} & \textbf{\shortstack{Total\\Cases}} \\
\midrule
\addlinespace
\multicolumn{10}{l}{\textbf{Malignancy diagnosis}} \\
\quad Internal, H1 & 0.001 & 1.000 & 95.4\% & 1055 & 0.856 & 1.000 & 1.8\% & 20 & 1106 \\
\quad Prospective, H8 & 0.001 & \textbf{1.000} & 33.3\% & 34 & 0.856 &\textbf{1.000} & 36.3\% & 37 & 102 \\
\quad Prospective, H7 & 0.001 & 0.986 & 63.7\% & 348 & 0.856 & \textbf{1.000} & 18.5\% & 101 & 546 \\
\addlinespace
\multicolumn{10}{l}{\textbf{Histologic differentiation grading}} \\
\quad Internal, H2 & 0.205 & 1.000 & 8.0\% & 11 & 0.492 & 1.000 & 62.3\% & 86 & 138 \\
\quad Prospective, H7 & 0.205 & \textbf{1.000} & 9.7\% & 14 & 0.492 & 0.963 & 55.2\% & 80 & 145 \\
\quad Prospective, H2 & 0.205 & \textbf{1.000} & 7.6\% & 7 & 0.492 & 0.807 & 62.0\% & 57 & 92 \\
\quad Prospective, H8 & 0.205 & \textbf{1.000} & 4.7\% & 2 & 0.492 & 0.300 & 46.5\% & 20 & 43 \\
\addlinespace
\multicolumn{10}{l}{\textbf{Lauren classification}} \\
\quad Internal, H1, H3, H4 & 0.042 & 1.000 & 29.8\% & 36 & 0.970 & 1.000 & 16.5\% & 20 & 121 \\
\quad Prospective, H7 & 0.042 & 0.929 & 19.2\% & 14 & 0.970 & \textbf{1.000} & 28.8\% & 21 & 73 \\
\addlinespace
\multicolumn{10}{l}{\textbf{Perineural invasion}} \\
\quad Internal, H1 & 0.062 & 1.000 & 22.5\% & 18 & 0.994 & 1.000 & 23.8\% & 19 & 80 \\
\quad Prospective, H8 & 0.062 & -- & 0.0\% & 0 & 0.994 & \textbf{1.000} & 5.3\% & 2 & 38 \\
\quad Prospective, H1 & 0.062 & -- & 0.0\% & 0 & 0.994 & \textbf{1.000} & 5.0\% & 3 & 60 \\
\addlinespace
\multicolumn{10}{l}{\textbf{MMR prediction}} \\
\quad Internal, H2 & 0.041 & 1.000 & 66.7\% & 54 & 0.839 & 1.000 & 4.9\% & 4 & 81 \\
\quad Prospective, H2 & 0.041 &\textbf{1.000}& 46.8\% & 37 & 0.839 & 0.500 & 10.1\% & 8 & 79 \\
\bottomrule
\end{tabular}
\begin{tablenotes}
\footnotesize
\item Abbreviations: NPV, negative predictive value; PPV, positive predictive value; MMR, mismatch repair.
\item A dash for NPV indicates that no cases were selected by the low-threshold rule ($n_{\mathrm{out}}=0$), so NPV is undefined.
\end{tablenotes}
\end{threeparttable}
\end{table}

\FloatBarrier
\begin{figure}
    \centering
    \includegraphics[width=\textwidth]{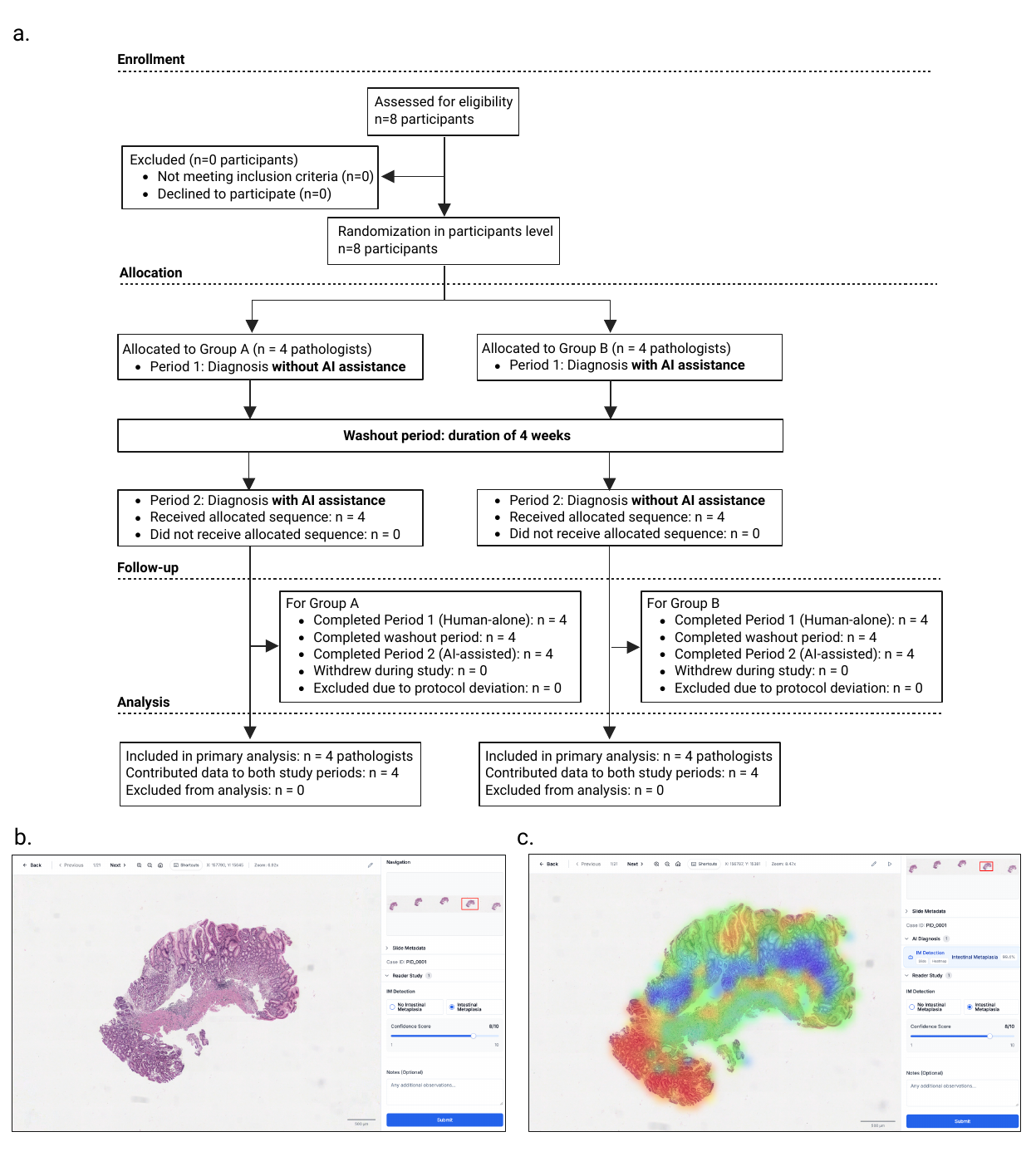}
    \captionof{figure}{\textbf{Randomized crossover reader study design and the reading environment.}
    (\textbf{a}) CONSORT flow diagram\cite{turner2012consolidated} for the randomized two-period crossover reader study. Eight eligible pathologists were randomized to unassisted diagnosis followed by AI-assisted diagnosis after a 4-week washout, or to the reverse sequence.
    (\textbf{b}-\textbf{c}) Representative WSI review interface during unassisted reading (\textbf{b}) and AI-assisted reading (\textbf{c}), shown using intestinal metaplasia (IM) diagnosis as an example. In the AI-assisted phase, GRACE displayed the predicted category, prediction probabilities, and WSI attention heatmaps.}
    \label{fig:consort}
\end{figure}

\FloatBarrier
\begin{table}[t]
\centering
\renewcommand{\arraystretch}{1.25}
\begin{threeparttable}
\captionsetup{justification=raggedright,singlelinecheck=false,width=\linewidth}
\caption{\textbf{Characteristics of cases across the three tasks included in the randomized crossover reader study.}}
\label{tab:RCT_data}
\begin{tabular}{p{10.5cm} c}
\toprule
\textbf{Characteristic} & \textbf{Number} \\
\midrule

\addlinespace
\multicolumn{2}{l}{\textbf{Overall}} \\
\quad Total unique cases$^{1}$ & 245 \\
\quad Total unique WSIs$^{1}$ & 245 \\
\quad Pathologists in crossover Group A & 4 \\
\quad Pathologists in crossover Group B & 4 \\
\quad Total task-specific case endpoints$^{2}$ & 300 \\
\quad Paired reader-task observations$^{3}$ & 2,400 \\
\quad Condition-specific reading events$^{4}$ & 4,800 \\

\addlinespace
\multicolumn{2}{l}{\textbf{Task 1: Acute activity severity grading}} \\
\quad \textit{Dataset size} \\
\qquad Cases & 100 \\
\qquad WSIs & 100 \\
\quad \textit{Class distribution (cases)} \\
\qquad Mild & 70 \\
\qquad Moderate & 28 \\
\qquad Severe & 2 \\

\addlinespace
\multicolumn{2}{l}{\textbf{Task 2: Atrophy diagnosis in chronic gastritis}} \\
\quad \textit{Dataset size} \\
\qquad Cases & 100 \\
\qquad WSIs & 100 \\
\quad \textit{Class distribution (cases)} \\
\qquad No atrophy & 46 \\
\qquad Atrophy & 54 \\

\addlinespace
\multicolumn{2}{l}{\textbf{Task 3: Intestinal metaplasia diagnosis}} \\
\quad \textit{Dataset size} \\
\qquad Cases & 100 \\
\qquad WSIs & 100 \\
\quad \textit{Class distribution (cases)} \\
\qquad No intestinal metaplasia & 68 \\
\qquad Intestinal metaplasia & 32 \\

\bottomrule
\end{tabular}
\begin{tablenotes}
\footnotesize
\item Abbreviations: WSI, whole-slide image.
 \item $^{1}$ Some WSIs were used in more than one diagnostic task. Therefore, task-specific counts do not sum to the number of unique WSIs.

\item $^{2}$ The reader study included 300 task-specific endpoints: 100 for acute activity severity grading, 100 for atrophy diagnosis, and 100 for intestinal metaplasia diagnosis, derived from 245 unique WSIs.

\item $^{3}$ Paired reader-task observations were calculated as 8 pathologists $\times$ 300 task-specific endpoints. Each observation includes matched unassisted and AI-assisted diagnoses for the same reader-task-case unit and was used as the denominator for paired performance and pathologist-AI impact analyses.

\item $^{4}$ Condition-specific reading events counted the two crossover conditions separately: 2,400 paired reader-task observations $\times$ 2 conditions = 4,800 reading events.
\end{tablenotes}
\end{threeparttable}
\end{table}

\FloatBarrier
\begin{table}[htbp]
\centering

\caption{\textbf{Overall comparisons of diagnostic performance, time, confidence, and inter-rater agreement with and without AI assistance.}}
\label{tab:overall_ai_vs_noai}
\begin{threeparttable}
\begin{tabular}{lcccc}
\toprule
\textbf{Metric$^{1}$} & \textbf{Without AI (95\% CI)} & \textbf{With AI (95\% CI)} & \textbf{Effect (95\% CI)$^{2}$} & \textbf{$P$ value$^{3}$} \\
\midrule

\multicolumn{5}{l}{\textbf{Diagnostic performance}} \\
\quad Accuracy     & 0.820 (0.802--0.839) & 0.899 (0.877--0.920) & +0.078$^{4}$ (+0.062 -- +0.095) & $<\!0.05$ \\
\quad Sensitivity & 0.809 (0.778--0.842) & 0.883 (0.845--0.918) & +0.073 (+0.052 -- +0.095)& $<\!0.05$ \\
\quad Specificity & 0.857 (0.835--0.878) & 0.926 (0.902--0.948) & +0.069 (+0.049 -- +0.089)& $<\!0.05$ \\
\quad PPV & 0.781 (0.722--0.829) & 0.883 (0.836--0.923) & +0.102 (+0.072 -- +0.135)& $<\!0.05$\\ 
\quad NPV & 0.877 (0.847--0.907) & 0.926 (0.899--0.952) & +0.049 (+0.034 -- +0.067)& $<\!0.05$\\ 

\addlinespace
\multicolumn{5}{l}{\textbf{Self-reported confidence (1--10)}} \\
\quad Mean & 8.482 (8.440--8.525) & 9.246 (9.209--9.286) & +0.764 (+0.728 -- +0.802) & $<\!0.05$ \\

\addlinespace
\multicolumn{5}{l}{\textbf{Diagnostic time (seconds)}} \\
\quad Median & 46.000 (45.000--47.000) & 40.000 (39.000--41.000) & -6.000 (-7.000 -- -5.000)& $<\!0.05$ \\
\quad Mean     & 47.710 (47.251--48.159) & 40.959 (40.511--41.429) & -6.750 (-7.059 -- -6.421) & $<\!0.05$ \\
\quad Geometric mean & 44.959 (44.529--45.374) & 38.264 (37.831--38.698) & 0.851 (0.844 -- 0.859) & $<\!0.05$ \\

\addlinespace
\multicolumn{5}{l}{\textbf{Inter-rater agreement}} \\
\quad Fleiss' $\kappa$ & 0.453 (0.408--0.493) & 0.743 (0.695--0.784) & +0.290 (+0.250 -- +0.328) & $<\!0.05$ \\

\bottomrule
\end{tabular}

\begin{tablenotes}[flushleft]
\footnotesize{

\item Abbreviations: AI, artificial intelligence; CI, confidence interval; NPV, negative predictive value; PPV, positive predictive value.
\item$^{1}$ Metric estimates are reported with 95\% CIs. Sensitivity, specificity, PPV, and NPV were calculated using predefined positive classes: moderate/severe acute activity for acute activity grading, atrophy for atrophy diagnosis, and intestinal metaplasia for intestinal metaplasia diagnosis. 
\item[$^{2}$] Effect represents the estimated difference between the AI-assisted condition and the independent (unassisted) condition for the same metric. For accuracy, sensitivity, specificity, PPV, NPV, median and mean diagnostic time, mean confidence, and Fleiss' $\kappa$, the effect is expressed as the absolute difference (AI-assisted $-$ independent). For geometric mean diagnostic time, the effect is expressed as the ratio of geometric means (AI-assisted / independent). Corresponding 95\% CIs were obtained by paired bootstrap resampling at the task-specific case-slide level (1{,}000 iterations).
\item$^{3}$ For metrics reported as absolute differences, the null hypothesis is $\Delta = 0$. For the geometric mean time ratio, inference is based on the bootstrap distribution of the log-ratio, corresponding to a null hypothesis of ratio $= 1$.
\item$^{4}$ Effects were calculated from unrounded pooled accuracy values derived from raw correct-diagnosis counts: 1969/2400 correct without AI assistance and 2157/2400 correct with AI assistance for overall accuracy.
}
\end{tablenotes}

\end{threeparttable}

\end{table}

\FloatBarrier
\begingroup

\begin{longtable}{lccc}
\caption{\textbf{Adjusted effects of AI assistance estimated using generalized estimating equation (GEE) models.}}
\label{tab:gee_bootstrap_all} \\
\toprule
\textbf{Metric} & \textbf{N Cases} & \textbf{Adjusted effect$^{1}$ (95\% bootstrap CI)} & \textbf{Bootstrap $P$ value$^{2}$} \\
\midrule
\endfirsthead

\multicolumn{4}{l}{\textit{\tablename~\thetable\ continued}}\\
\toprule
\textbf{Metric} & \textbf{N Cases} & \textbf{Adjusted effect$^{1}$ (95\% bootstrap CI)} & \textbf{Bootstrap $P$ value$^{2}$} \\
\midrule
\endhead

\midrule
\multicolumn{4}{r}{\textit{Continued on next page}} \\
\endfoot

\endlastfoot

\multicolumn{4}{l}{\textbf{Overall}} \\
\quad Accuracy & 300 & 1.987 (1.679--2.387) & $<\!0.05$ \\
\quad Sensitivity & 300 & 1.820 (1.478--2.387) & $<\!0.05$ \\
\quad Specificity & 300 & 2.119 (1.689--2.867) & $<\!0.05$ \\
\quad PPV & 300 & 2.095 (1.708--2.796) & $<\!0.05$ \\
\quad NPV & 300 & 1.779 (1.478--2.309) & $<\!0.05$ \\
\quad Diagnostic time & 300 & 0.851 (0.844--0.859) & $<\!0.05$ \\
\quad Confidence & 300 & +0.764 (+0.722 -- +0.803) & $<\!0.05$ \\
\midrule
\multicolumn{4}{l}{\textbf{\textit{By task}}} \\
\multicolumn{4}{l}{\quad \textbf{AA grading}} \\
\quad\quad Accuracy & 100 & 2.086 (1.549--3.050) & $<\!0.05$ \\
\quad\quad Sensitivity & 100 & 1.423 (0.958--2.711) & 0.102 \\
\quad\quad Specificity & 100 & 2.731 (1.729--5.956) & $<\!0.05$ \\
\quad\quad PPV & 100 & 2.639 (1.718--5.474) & $<\!0.05$ \\
\quad\quad NPV & 100 & 1.475 (1.050--2.661) & $<\!0.05$ \\
\quad\quad Diagnostic time & 100 & 0.844 (0.833--0.856) & $<\!0.05$ \\
\quad\quad Confidence & 100 & +0.704 (+0.641 -- +0.763) & $<\!0.05$ \\

\multicolumn{4}{l}{\quad \textbf{Atrophy diagnosis}} \\
\quad\quad Accuracy & 100 & 1.841 (1.443--2.555) & $<\!0.05$ \\
\quad\quad Sensitivity & 100 & 2.017 (1.523--2.910) & $<\!0.05$ \\
\quad\quad Specificity & 100 & 1.660 (1.113--2.889) & $<\!0.05$ \\
\quad\quad PPV & 100 & 1.753 (1.237--2.962) & $<\!0.05$ \\
\quad\quad NPV & 100 & 1.885 (1.465--2.662) & $<\!0.05$ \\
\quad\quad Diagnostic time & 100 & 0.865 (0.851--0.878) & $<\!0.05$ \\
\quad\quad Confidence & 100 & +0.810 (+0.746 -- +0.879) & $<\!0.05$ \\

\multicolumn{4}{l}{\quad \textbf{IM diagnosis}} \\
\quad\quad Accuracy & 100 & 2.120 (1.604--3.149) & $<\!0.05$ \\
\quad\quad Sensitivity & 100 & 1.988 (1.250--4.658) & $<\!0.05$ \\
\quad\quad Specificity & 100 & 2.189 (1.566--3.491) & $<\!0.05$ \\
\quad\quad PPV & 100 & 2.143 (1.577--3.402) & $<\!0.05$ \\
\quad\quad NPV & 100 & 2.025 (1.322--4.565) & $<\!0.05$ \\
\quad\quad Diagnostic time & 100 & 0.845 (0.835--0.855) & $<\!0.05$ \\
\quad\quad Confidence & 100 & +0.777 (+0.701 -- +0.848) & $<\!0.05$ \\

\midrule
\multicolumn{4}{l}{\textbf{\textit{By experience level}}} \\
\multicolumn{4}{l}{\quad \textbf{Junior}} \\
\quad\quad Accuracy & 300 & 2.010 (1.667--2.472) & $<\!0.05$ \\
\quad\quad Sensitivity & 300 & 1.846 (1.465--2.496) & $<\!0.05$ \\
\quad\quad Specificity & 300 & 2.100 (1.594--2.867) & $<\!0.05$ \\
\quad\quad PPV & 300 & 2.063 (1.623--2.758) & $<\!0.05$ \\
\quad\quad NPV & 300 & 1.792 (1.460--2.370) & $<\!0.05$ \\
\quad\quad Diagnostic time & 300 & 0.823 (0.815--0.831) & $<\!0.05$ \\
\quad\quad Confidence & 300 & +1.184 (+1.119 -- +1.240) & $<\!0.05$ \\

\multicolumn{4}{l}{\quad \textbf{Senior}} \\
\quad\quad Accuracy & 300 & 1.951 (1.568--2.504) & $<\!0.05$ \\
\quad\quad Sensitivity & 300 & 1.789 (1.329--2.563) & $<\!0.05$ \\
\quad\quad Specificity & 300 & 2.176 (1.569--3.490) & $<\!0.05$ \\
\quad\quad PPV & 300 & 2.141 (1.585--3.375) & $<\!0.05$ \\
\quad\quad NPV & 300 & 1.768 (1.347--2.475) & $<\!0.05$ \\
\quad\quad Diagnostic time & 300 & 0.880 (0.868--0.892) & $<\!0.05$ \\
\quad\quad Confidence & 300 & +0.343 (+0.295 -- +0.389) & $<\!0.05$ \\

\midrule
\multicolumn{4}{p{\dimexpr\textwidth-2\tabcolsep\relax}}{\footnotesize
Abbreviations: GEE, generalized estimating equation; N cases, number of cases; CI, confidence interval; NPV, negative predictive value; PPV, positive predictive value; AA, acute activity; IM, intestinal metaplasia; OR, odds ratio.}\\
\multicolumn{4}{p{\dimexpr\textwidth-2\tabcolsep\relax}}{\footnotesize
$^{1}$ Adjusted effects are reported as ORs for accuracy, sensitivity, specificity, PPV and NPV; as adjusted ratios of geometric mean diagnostic time for diagnostic time; and as adjusted mean differences for confidence. Diagnostic time was modeled on the log scale. Time ratios below 1 indicate shorter reading time with AI assistance, and positive mean differences indicate higher confidence with AI assistance.}\\
\multicolumn{4}{p{\dimexpr\textwidth-2\tabcolsep\relax}}{\footnotesize
$^{2}$ For adjusted analyses, the corresponding GEE model was refitted in each bootstrap replicate, and percentile-based 95\% CIs and two-sided bootstrap $P$ values were derived from the bootstrap distribution of the adjusted effect estimate.}\\
\multicolumn{4}{p{\dimexpr\textwidth-2\tabcolsep\relax}}{\footnotesize
Overall models adjusted for experience level, crossover sequence, and task; stratified models omitted the stratification variable.}\\

\end{longtable}
\endgroup

\FloatBarrier
\begin{longtable}{lccccc}
\caption{\textbf{Diagnostic performance with and without AI assistance, shown overall, by task, and by reader level.}}
\label{tab:performance} \\

\toprule
\textbf{Metrics$^{1}$}
& \textbf{N Cases}
& \textbf{Without AI (95\% CI)}
& \textbf{With AI (95\% CI)}
& \textbf{Effect (95\% CI)$^{2}$}
& \textbf{$P$ value$^{3}$} \\
\midrule
\endfirsthead

\multicolumn{6}{l}{\textit{\tablename~\thetable\ continued}} \\
\toprule
\textbf{Metrics$^{1}$}
& \textbf{N Cases}
& \textbf{Without AI (95\% CI)}
& \textbf{With AI (95\% CI)}
& \textbf{Effect (95\% CI)$^{2}$}
& \textbf{$P$ value$^{3}$} \\
\midrule
\endhead

\midrule
\multicolumn{6}{r}{\textit{Continued on next page}} \\
\endfoot

\bottomrule
\endlastfoot

\multicolumn{6}{l}{\textbf{Overall}} \\
\quad Accuracy & 300 & 0.820 (0.802--0.839) & 0.899 (0.877--0.920) & +0.078 (+0.062 -- +0.095) & $<\!0.05$ \\
\quad Sensitivity & 300 & 0.809 (0.778--0.842) & 0.883 (0.845--0.918) & +0.073 (+0.052 -- +0.095) & $<\!0.05$ \\
\quad Specificity & 300 & 0.857 (0.835--0.878) & 0.926 (0.902--0.948) & +0.069 (+0.049 -- +0.089) & $<\!0.05$ \\
\quad PPV & 300 & 0.781 (0.722--0.829) & 0.883 (0.836--0.923) & +0.102 (+0.072 -- +0.135) & $<\!0.05$ \\
\quad NPV & 300 & 0.877 (0.847--0.907) & 0.926 (0.899--0.952) & +0.049 (+0.034 -- +0.067) & $<\!0.05$ \\

\midrule
\multicolumn{6}{l}{\textbf{\textit{By task}}} \\

\quad \textit{AA grading} & & & & & \\
\qquad Accuracy & 100 & 0.800 (0.761--0.838) & 0.891 (0.849--0.930) & +0.091 (+0.059 -- +0.126) & $<\!0.05$ \\
\qquad Sensitivity & 100 & 0.821 (0.746--0.886) & 0.867 (0.783--0.946) & +0.046 (-0.004 -- +0.094) & 0.086 \\
\qquad Specificity & 100 & 0.868 (0.829--0.904) & 0.946 (0.908--0.976) & +0.079 (+0.045 -- +0.114) & $<\!0.05$ \\
\qquad PPV & 100 & 0.727 (0.605--0.817) & 0.874 (0.774--0.945) & +0.147 (+0.084 -- +0.226) & $<\!0.05$ \\
\qquad NPV & 100 & 0.919 (0.873--0.956) & 0.943 (0.902--0.978) & +0.024 (+0.004 -- +0.049) & $<\!0.05$ \\

\quad \textit{Atrophy diagnosis} & & & & & \\
\qquad Accuracy & 100 & 0.781 (0.744--0.814) & 0.866 (0.820--0.906) & +0.085 (+0.056 -- +0.111) & $<\!0.05$ \\
\qquad Sensitivity & 100 & 0.755 (0.706--0.798) & 0.856 (0.797--0.910) & +0.102 (+0.068 -- +0.135) & $<\!0.05$ \\
\qquad Specificity & 100 & 0.812 (0.764--0.858) & 0.878 (0.812--0.936) & +0.065 (+0.020 -- +0.110)& $<\!0.05$\\
\qquad PPV & 100 & 0.825 (0.756--0.885) & 0.892 (0.825--0.947) & +0.066 (+0.030 -- +0.110) & $<\!0.05$ \\
\qquad NPV & 100 & 0.738 (0.644--0.813) & 0.839 (0.757--0.907) & +0.101 (+0.063 -- +0.144) & $<\!0.05$ \\

\quad \textit{IM diagnosis} & & & & & \\
\qquad Accuracy & 100 & 0.880 (0.854--0.905) & 0.939 (0.915--0.961) & +0.059 (+0.040 -- +0.080) & $<\!0.05$ \\
\qquad Sensitivity & 100 & 0.891 (0.838--0.935) & 0.941 (0.882--0.979) & +0.051 (+0.019 -- +0.083) & $<\!0.05$ \\
\qquad Specificity & 100 & 0.875 (0.846--0.901) & 0.938 (0.908--0.962) & +0.062 (+0.037 -- +0.090) & $<\!0.05$ \\
\qquad PPV & 100 & 0.770 (0.669--0.846) & 0.876 (0.793--0.933) & +0.106 (+0.059 -- +0.160) & $<\!0.05$ \\
\qquad NPV & 100 & 0.944 (0.908--0.972) & 0.971 (0.937--0.991) & +0.027 (+0.011 -- +0.047) & $<\!0.05$ \\
\midrule
\multicolumn{6}{l}{\textbf{\textit{By experience level}}} \\

\quad \textit{Junior} & & & & & \\
\qquad Accuracy & 300 & 0.762 (0.734--0.791) & 0.864 (0.834--0.893) & +0.103 (+0.080 -- +0.129) & $<\!0.05$ \\
\qquad Sensitivity & 300 & 0.748 (0.702--0.797) & 0.843 (0.788--0.891) & +0.095 (+0.059 -- +0.128) & $<\!0.05$ \\
\qquad Specificity & 300 & 0.812 (0.781--0.844) & 0.901 (0.866--0.933) & +0.088 (+0.060 -- +0.119) & $<\!0.05$ \\
\qquad PPV & 300 & 0.715 (0.650--0.770) & 0.843 (0.777--0.893) & +0.127 (+0.088 -- +0.170) & $<\!0.05$ \\
\qquad NPV & 300 & 0.836 (0.795--0.874) & 0.901 (0.864--0.934) & +0.064 (+0.041 -- +0.087) & $<\!0.05$ \\

\quad \textit{Senior} & & & & & \\
\qquad Accuracy & 300 & 0.879 (0.861--0.898) & 0.933 (0.916--0.952) & +0.054 (+0.039 -- +0.070) & $<\!0.05$ \\
\qquad Sensitivity & 300 & 0.871 (0.836--0.904) & 0.922 (0.888--0.954) & +0.052 (+0.029 -- +0.076) & $<\!0.05$ \\
\qquad Specificity & 300 & 0.901 (0.878--0.922) & 0.951 (0.930--0.970) & +0.050 (+0.030 -- +0.071) & $<\!0.05$ \\
\qquad PPV & 300 & 0.847 (0.801--0.888) & 0.922 (0.882--0.956) & +0.075 (+0.046 -- +0.108) & $<\!0.05$ \\
\qquad NPV & 300 & 0.917 (0.891--0.943) & 0.951 (0.928--0.971) & +0.034 (+0.019 -- +0.051) & $<\!0.05$ \\

\end{longtable}

\begin{minipage}{\textwidth}
\footnotesize
Abbreviations: N cases, number of cases; AI, artificial intelligence; CI, confidence interval; PPV, positive predictive value; NPV, negative predictive value; AA, acute activity; IM, intestinal metaplasia.

$^{1}$ Reported values are the metric estimate with 95\% CI under each condition. Sensitivity, specificity, PPV, and NPV were computed using a predefined positive class. 

$^{2}$ Effect indicates the absolute difference between AI-assisted and independent conditions (AI-assisted $-$ independent). Corresponding 95\% CIs were obtained by paired bootstrap resampling at the task-specific case-slide level (1{,}000 iterations).

$^{3}$ Two-sided bootstrap $P$ values test $H_0: \Delta = 0$.

\end{minipage}

\begin{table}[htbp]
\centering
\begin{threeparttable}
\caption{\textbf{Diagnostic confidence (1--10) with and without AI assistance, shown overall, by task, and by reader level. }}
\label{tab:confidence}

\begin{tabular}{lccccc}
\toprule
\textbf{Metrics$^{1}$}
& \textbf{N Cases}
& \textbf{Without AI (95\% CI)}
& \textbf{With AI (95\% CI)}
& \textbf{$\Delta$ (95\% CI)$^{2}$}
& \textbf{$P$ value$^{3}$} \\
\midrule
\textbf{Overall} & 300 & 8.482 (8.440--8.525) & 9.246 (9.209--9.286) & +0.764 (+0.728 -- +0.802) & $<\!0.05$ \\

\midrule
\multicolumn{6}{l}{\textbf{\textit{By task}}} \\

\quad AA grading & 100 & 8.424 (8.359--8.481) & 9.127 (9.064--9.189) & +0.704 (+0.642 -- +0.764) & $<\!0.05$ \\
\quad Atrophy diagnosis & 100 & 8.384 (8.310--8.456) & 9.194 (9.124--9.260) & +0.810 (+0.738 -- +0.876) & $<\!0.05$ \\
\quad IM diagnosis & 100 & 8.640 (8.571--8.700) & 9.418 (9.355--9.471) & +0.777 (+0.708 -- +0.849) & $<\!0.05$ \\

\midrule
\multicolumn{6}{l}{\textbf{\textit{By experience level}}} \\

\quad Junior & 300 & 7.691 (7.621--7.763) & 8.875 (8.817--8.934) & +1.184 (+1.116 -- +1.246) & $<\!0.05$ \\
\quad Senior & 300 & 9.274 (9.236--9.312) & 9.617 (9.578--9.660) & +0.343 (+0.301 -- +0.388) & $<\!0.05$ \\

\bottomrule
\end{tabular}

\begin{tablenotes}
\footnotesize
\item Abbreviations: N cases, number of cases; AI, artificial intelligence; CI, confidence interval; AA, acute activity; IM, intestinal metaplasia.
\item$^{1}$ Reported values are the mean confidence scores with 95\% CIs under each condition.
\item$^{2}$ $\Delta$ indicates the difference between AI-assisted and independent conditions. 95\% CIs were obtained using paired bootstrap resampling at the task-specific case-slide level (1{,}000 iterations).
\item$^{3}$ Two-sided $P$ values test $H_0: \Delta = 0$.
\end{tablenotes}

\end{threeparttable}
\end{table}

\FloatBarrier
\begin{figure}[p]
\captionsetup{labelformat=empty}
\centering
\includegraphics[width=\textwidth]{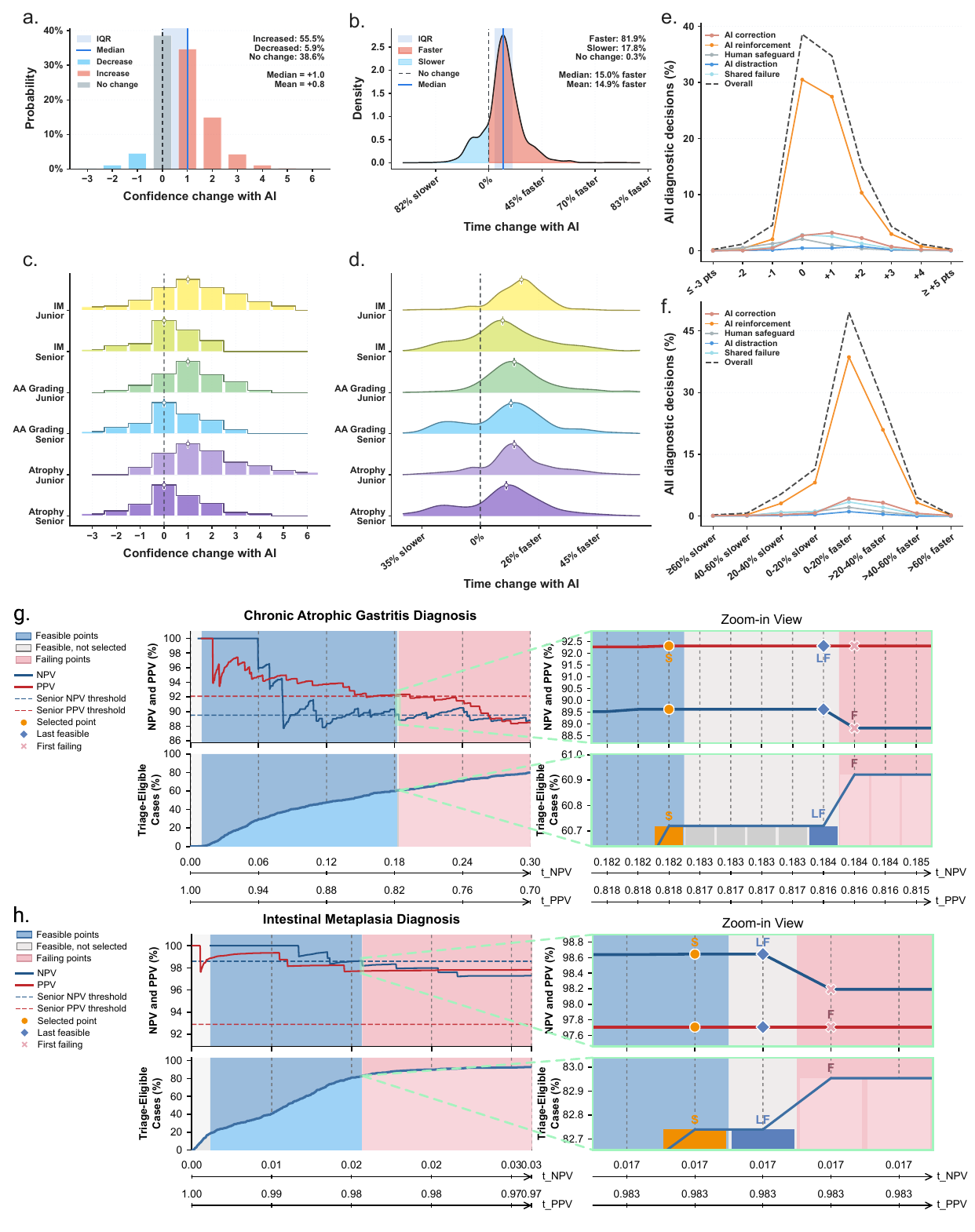}
\caption{}
\end{figure}
\begin{figure}
\ContinuedFloat
\caption{%
\textbf{Impact of AI assistance on reader behavior and pathologist-referenced triage.}
(\textbf{a}-\textbf{b}) Overall distributions of AI-associated changes in diagnostic reading time and confidence, measured relative to the unassisted condition. 
(\textbf{c}-\textbf{d}) Task- and experience-stratified ridgeline distributions of changes in confidence and reading time. Each ridge represents one task-experience subgroup, with height reflecting relative frequency. White vertical markers indicate subgroup medians.
(\textbf{e}-\textbf{f}) Distributions of diagnostic decisions across confidence shift (AI-assisted minus unassisted) and time-efficiency shift (log ratio of unassisted to AI-assisted reading time), stratified by pathologist-AI interaction category. The y-axis represents the percentage of all diagnostic decisions.
(\textbf{g}-\textbf{h}) Pathologist-referenced review triage analysis. Paired rule-out and rule-in thresholds \((t_{\mathrm{NPV}}, t_{\mathrm{PPV}})\) were tested to identify the largest triage-eligible set of AI-resolved diagnostic endpoints whose NPV and PPV remained above senior-pathologist benchmark values. The upper panels show NPV and PPV across threshold pairs, and the lower panels show the triage-eligible proportion. The selected operating point (S) maximizes the triage-eligible proportion among threshold pairs satisfying both senior-pathologist upper-bound benchmarks from case-level bootstrap: NPV 98.6\% and PPV 92.9\% for IM diagnosis, and NPV 89.5\% and PPV 92.1\% for chronic atrophic gastritis diagnosis. Zoom-in panels mark the selected point (S), the last feasible point (LF), and the first PPV-failing point (F).
}
    \label{fig:AI_impact_cat_line}
    \end{figure}

\begin{table}[!htbp]
\centering
\begin{threeparttable}
\caption{\textbf{Diagnostic time (seconds) with and without AI assistance, shown overall, by task, and by reader level.}}
\label{tab:time}
\begin{tabular}{lccccc}
\toprule
\textbf{Metrics$^{1}$}
& \textbf{N Cases}
& \textbf{Without AI (95\% CI)}
& \textbf{With AI (95\% CI)}
& \textbf{Effect (95\% CI)$^{2}$ }
& \textbf{$P$ value$^{3}$} \\
\midrule

\multicolumn{6}{l}{\textbf{Overall}} \\
\quad Median & 300 & 46.000 (45.000--47.000) & 40.000 (39.000--41.000) & -6.000 (-7.000 -- -5.000) & $<\!0.05$  \\
\quad Mean & 300 & 47.710 (47.251--48.159) & 40.959 (40.511--41.429) & -6.750 (-7.059 -- -6.421) & $<\!0.05$ \\
\quad Geometric mean & 300 & 44.959 (44.529--45.374) & 38.264 (37.831--38.698) & 0.851 (0.844--0.859) & $<\!0.05$ \\

\midrule
\multicolumn{6}{l}{\textbf{\textit{By task}}} \\

\quad \textit{AA grading} & & & & & \\
\qquad Median & 100 & 46.000 (45.000--47.000) & 40.000 (39.000--41.000) & -6.000 (-7.000 -- -5.000)&$<\!0.05$  \\
\qquad Mean & 100 & 49.265 (48.684--49.878) & 42.265 (41.622--42.965) & -7.000 (-7.503 -- -6.489) & $<\!0.05$ \\
\qquad Geometric mean & 100 & 47.208 (46.608--47.835) & 39.827 (39.140--40.532) & 0.844 (0.832--0.855) & $<\!0.05$ \\
\quad \textit{Atrophy diagnosis} & & & & & \\
\qquad Median & 100 & 49.000 (48.000--50.000) & 43.000 (42.000--44.000) & -6.000 (-8.000 -- -5.000)&$<\!0.05$  \\
\qquad Mean & 100 & 48.981 (48.305--49.644) & 42.414 (41.616--43.195) & -6.567 (-7.168 -- -5.979) & $<\!0.05$ \\
\qquad Geometric mean & 100 & 45.053 (44.462--45.652) & 38.956 (38.184--39.689) & 0.865 (0.852--0.877) & $<\!0.05$ \\
\quad \textit{IM diagnosis} & & & & & \\
\qquad Median & 100 & 44.000 (43.000--45.000) & 38.000 (37.000--39.000) & -6.000 (-7.000 -- -5.000)&$<\!0.05$  \\
\qquad Mean & 100 & 44.883 (44.155--45.654) & 38.199 (37.551--38.871) & -6.684 (-7.156 -- -6.231) & $<\!0.05$ \\
\qquad Geometric mean & 100 & 42.728 (42.056--43.408) & 36.109 (35.493--36.719) & 0.845 (0.834--0.855) & $<\!0.05$ \\

\midrule
\multicolumn{6}{l}{\textbf{\textit{By experience level}}} \\

\quad \textit{Junior} & & & & & \\
\qquad Median & 300 & 56.000 (55.000--57.000) & 46.000 (45.000--47.000) & -10.000 (-11.000 -- -9.000)&$<\!0.05$  \\
\qquad Mean & 300 & 56.692 (55.961--57.432) & 47.355 (46.706--48.076) & -9.337 (-9.790 -- -8.909) & $<\!0.05$ \\
\qquad Geometric mean & 300 & 54.911 (54.174--55.650) & 45.207 (44.573--45.913) & 0.823 (0.816--0.831) & $<\!0.05$ \\
\quad \textit{Senior} & & & & & \\
\qquad Median & 300 & 39.000 (38.000--39.000) & 34.000 (33.000--35.000) & -5.000 (-6.000 -- -4.000)&$<\!0.05$  \\
\qquad Mean & 300 & 38.727 (38.183--39.220) & 34.563 (33.990--35.110) & -4.164 (-4.576 -- -3.749) & $<\!0.05$ \\
\qquad Geometric mean & 300 & 36.811 (36.325--37.261) & 32.387 (31.862--32.917) & 0.880 (0.869--0.891) & $<\!0.05$ \\

\bottomrule
\end{tabular}

\begin{tablenotes}
\footnotesize
\item Abbreviations: N cases, number of cases; AI, artificial intelligence; CI, confidence interval; AA, acute activity; IM, intestinal metaplasia; median, median diagnostic time; mean, mean diagnostic time; geometric mean, geometric mean diagnostic time.
\item$^{1}$ Reported values are the estimated diagnostic time under each condition, with 95\% CIs. Depending on the metric, values in the "Without AI" and "With AI" columns represent the median, arithmetic mean, or geometric mean reading time.
\item$^{2}$ Effect indicates the estimated difference or ratio comparing the AI-assisted and independent conditions for the same metric. For median and mean diagnostic time, the effect is expressed as the absolute difference (AI-assisted $-$ independent). For geometric mean diagnostic time, the effect is expressed as the ratio of geometric means (AI-assisted divided by independent). Corresponding 95\% CIs were obtained by paired bootstrap resampling at the task-specific case-slide level (1{,}000 iterations).
\item$^{3}$ For metrics reported as absolute differences, the null hypothesis was $\Delta = 0$. For the geometric mean time ratio, inference was based on the bootstrap distribution of the log-ratio, corresponding to a null hypothesis of ratio $= 1$.
\end{tablenotes}

\end{threeparttable}
\end{table}

\begin{table}[!htbp]
\centering
\begin{threeparttable}
\caption{\textbf{Inter-rater agreement with and without AI assistance, shown overall, by task, and by reader level.}}
\label{tab:fleiss}

\begin{tabular}{lccccc}
\toprule
\textbf{Metrics$^{1}$}
& \textbf{N Cases}
& \textbf{Without AI $\kappa$ (95\% CI)}
& \textbf{With AI $\kappa$ (95\% CI)}
& \textbf{$\Delta\kappa$ (95\% CI)$^{2}$ }
& \textbf{$P$ value$^{3}$} \\
\midrule

\textbf{Overall} & 300 & 0.453 (0.408--0.493) & 0.743 (0.695--0.784) & +0.290 (+0.250 -- +0.328) & $<\!0.05$ \\

\midrule
\multicolumn{6}{l}{\textbf{\textit{By task}}} \\

\quad AA grading 
& 100 
& 0.406 (0.331--0.470) & 0.723 (0.632--0.802) & +0.318 (+0.239 -- +0.407)
& $<\!0.05$ \\

\quad Atrophy diagnosis
& 100 
& 0.356 (0.278--0.428) & 0.687 (0.600--0.763) & +0.331 (+0.265 -- +0.399)
& $<\!0.05$ \\

\quad IM diagnosis 
& 100 
&0.559 (0.465--0.638) & 0.785 (0.713--0.848) & +0.227 (+0.159 -- +0.297)
& $<\!0.05$ \\

\midrule
\multicolumn{6}{l}{\textbf{\textit{By experience level}}} \\

\quad Junior 
& 300 & 0.350 (0.287--0.406) & 0.717 (0.659--0.770) & +0.367 (+0.310 -- +0.429) & $<\!0.05$ \\

\quad Senior 
& 300 & 0.583 (0.520--0.644) & 0.806 (0.754--0.849)  & +0.223 (+0.169 -- +0.274) & $<\!0.05$ \\

\bottomrule
\end{tabular}

\begin{tablenotes}
\footnotesize
\item Abbreviations: N cases, number of cases; AI, artificial intelligence; CI, confidence interval; 
AA, acute activity; IM, intestinal metaplasia.
\item$^{1}$ Reported values are Fleiss' $\kappa$ estimates with 95\% CIs. Agreement levels were interpreted according to Landis and Koch\cite{landis1977measurement}: slight (0.00--0.20), fair (0.21--0.40), moderate (0.41--0.60), substantial (0.61--0.80), and almost perfect (0.81--1.00).
\item$^{2}$ $\Delta\kappa$ indicates the difference in Fleiss' $\kappa$ between AI-assisted and independent conditions. 95\% CIs were obtained using paired bootstrap resampling at the task-specific case-slide level (1{,}000 iterations). 
\item$^{3}$ Two-sided $P$ values test $H_0: \Delta\kappa = 0$.
\end{tablenotes}

\end{threeparttable}
\end{table}

\begin{table}[htbp]
\centering
\setlength{\tabcolsep}{4pt}
\renewcommand{\arraystretch}{1.15}
\caption{\textbf{Definition of AI impact categories based on diagnostic correctness across human-only, AI-only, and AI-assisted conditions.}}
\label{tab:ai_impact_categories}
\begin{tabular}{lcccc}
\toprule
\textbf{Category} 
& \textbf{Human-alone correctness ($s$)}
& \textbf{AI model correctness ($m$)}
& \textbf{Correctness with AI ($a$)}
& \textbf{Label changed} \\
\midrule
AI correction      
& $\times$ 
& $\checkmark$  
& $\checkmark$ 
& $\checkmark$ \\

AI reinforcement   
& $\checkmark$ 
& $\checkmark$ 
& $\checkmark$ 
& $\times$ \\

Human safeguard    
& $\checkmark$ 
& $\times$ 
& $\checkmark$ 
& $\times$ \\

Human safeguard    
& $\times$ 
& $\times$ 
& $\checkmark$ 
& $\checkmark$ \\

AI distraction     
& $\checkmark$ 
& $\times$ 
& $\times$ 
& $\checkmark$ \\

AI distraction     
& $\checkmark$ 
& $\checkmark$ 
& $\times$ 
& $\checkmark$ \\

Shared failure     
& $\times$ 
& $\times$ 
& $\times$ 
& $\times$ or $\checkmark$ \\

Shared failure     
& $\times$ 
& $\checkmark$ 
& $\times$ 
& $\times$ or $\checkmark$ \\
\bottomrule
\end{tabular}
\vspace{2pt}
\end{table}

\begin{table}[!htbp]
\centering
\setlength{\tabcolsep}{4pt}
\renewcommand{\arraystretch}{1.15}
\begin{threeparttable}
\caption{\textbf{Pathologist-AI interaction outcomes by task and pathologist experience level.}}
\label{tab:task_level_outcomes}
\begin{tabular}{p{4.2cm} c ccc cc}
\toprule
\multirow[c]{2}{*}{\textbf{Metric}} &
\multirow[c]{2}{*}{\textbf{Overall}} &
\multicolumn{3}{c}{\textbf{Task}} &
\multicolumn{2}{c}{\textbf{Level}} \\
\cmidrule(lr){3-5} \cmidrule(lr){6-7}
& & \textbf{AA} & \textbf{Atrophy} & \textbf{IM} & \textbf{Junior} & \textbf{Senior} \\
\midrule
AI model accuracy, \%
& 87.33 & 84.00 & 83.00 & 95.00 & 87.33 & 87.33 \\

Diagnosis change rate\tnote{1}, \%
& 12.21 & 14.00 & 13.25 & 9.38 & 16.50 & 7.92 \\

\midrule
\multicolumn{7}{l}{\textbf{\textit{AI impact category, n (\%)}}} \\

\hspace{0.5em}AI correction
& 231 (9.63\%) & 87 (10.88\%) & 84 (10.50\%) & 60 (7.50\%) & 154 (12.83\%) & 77 (6.42\%) \\

\hspace{0.5em}AI reinforcement
& 1,792 (74.67\%) & 571 (71.38\%) & 551 (68.88\%) & 670 (83.75\%) & 840 (70.00\%) & 952 (79.33\%) \\

\hspace{0.5em}Human safeguard
& 134 (5.58\%) & 55 (6.88\%) & 58 (7.25\%) & 21 (2.62\%) & 43 (3.58\%) & 91 (7.58\%) \\

\hspace{0.5em}AI distraction
& 51 (2.13\%) & 18 (2.25\%) & 19 (2.38\%) & 14 (1.75\%) & 36 (3.00\%) & 15 (1.25\%) \\

\hspace{0.5em}Shared failure
& 192 (8.00\%) & 69 (8.62\%) & 88 (11.00\%) & 35 (4.38\%) & 127 (10.58\%) & 65 (5.42\%) \\

\bottomrule
\end{tabular}
\begin{tablenotes}
\footnotesize
\item Abbreviations: AA, acute activity severity grading; 
IM, intestinal metaplasia diagnosis; AI, artificial intelligence. 
\item $^{1}$ Diagnosis change rate denotes the proportion of diagnostic observations in which the initial diagnosis was altered after AI assistance.

\end{tablenotes}

\end{threeparttable}
\end{table}

\FloatBarrier

\begin{figure}
    \centering
    \includegraphics[width=\textwidth]{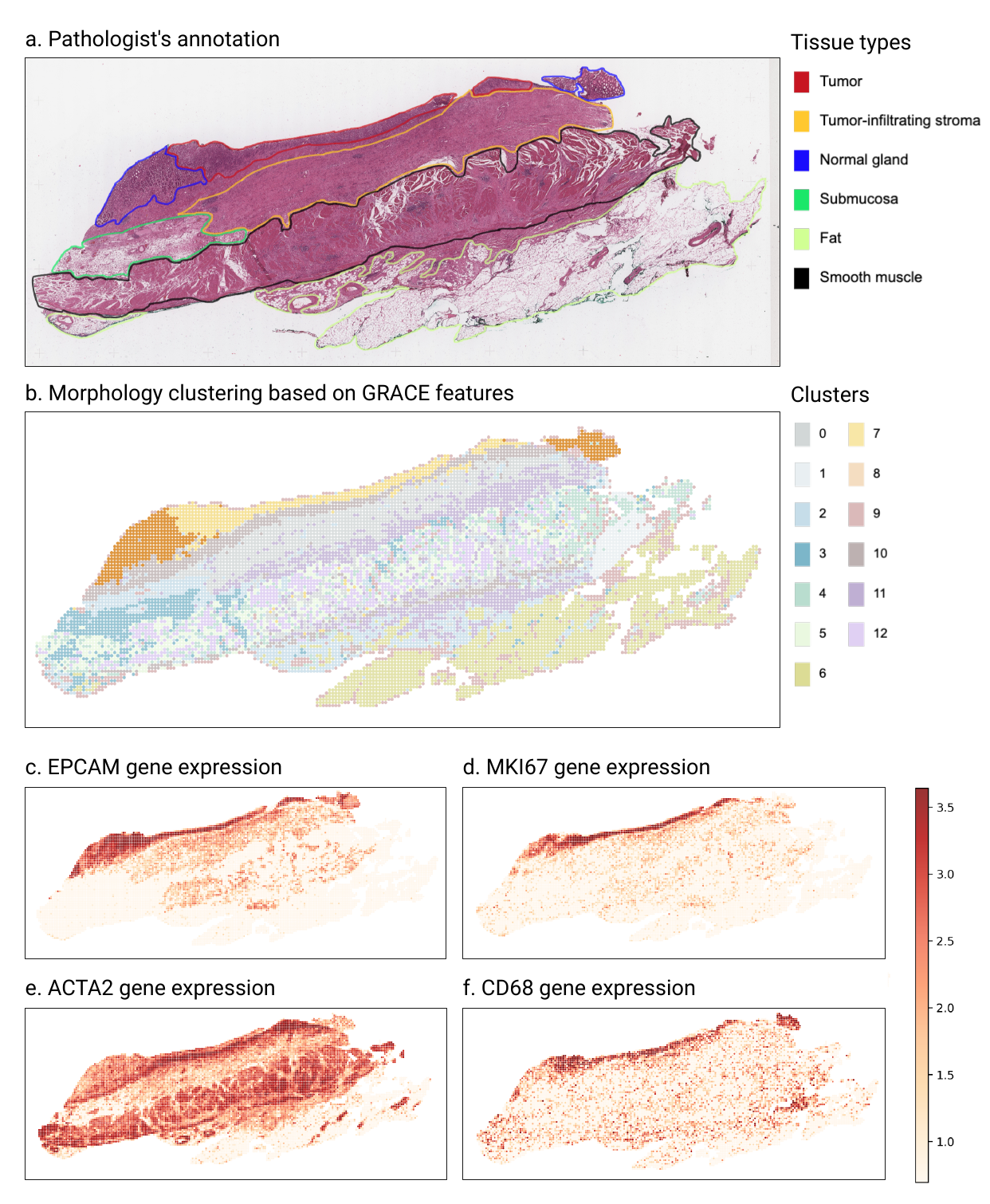}
    \caption{\textbf{Comparison of pathologist-annotated tissue compartments, GRACE feature-based clusters, and marker gene expression patterns.}
    (\textbf{a}) Pathologist annotation of the tissue section, delineating the major histological compartments, including tumor, tumor-infiltrating stroma, normal glands, submucosa, fat, and smooth muscle.
    (\textbf{b}) Unsupervised clustering of GRACE features, showing the spatial distribution of 13 feature-defined clusters across the same tissue section.
    (\textbf{c}-\textbf{f}) Spatial expression maps of selected marker genes, including EPCAM, MKI67, ACTA2, and CD68, displayed on the aligned tissue section. Greater color intensity indicates higher log-transformed expression.}
    \label{fig:ST}
\end{figure}

\FloatBarrier
\renewcommand{\arraystretch}{1.5}
\begin{table}[ht]
\caption{\textbf{Dataset composition for pretraining.}}
\centering
\begin{tabularx}{\textwidth}{p{4cm} c c Y}
\toprule
\textbf{Dataset} & \textbf{Total Slides} & \textbf{Stain Type} & \textbf{Link} \\
\midrule
Hospital H7 & 23,708 & H\&E & Private data \\

HistAI & 202 & H\&E and other stains &
\url{https://github.com/HistAI/HISTAI} \\

The Cancer
Genome Atlas (TCGA) & 419 & H\&E &
\url{https://www.cancer.gov/tcga} \\

Comprehensive Assessment of Chronic Gastritis on WSI Data & 792 & H\&E &
\url{https://www.scidb.cn/detail?dataSetId=83ee1521074742cdaae997cf0b46a7b1} \\

Gastric Cancer Lymph Node Dataset & 500 & H\&E &
\url{https://figshare.com/articles/dataset/Gastric_cancer_lymph_node_data_set/13065986} \\

KBSMC Gastric Cancer Grading Dataset & 98 & H\&E & \url{https://github.com/colin19950703/KBSMC_gastric_cancer_grading_dataset} \\
\bottomrule
\end{tabularx}
\label{tab:pretrain_dataset}
\end{table}

\begin{table}[H]
\centering
\caption{\textbf{Definitions of core benchmarking units used in this study.}}
\label{tab:defs_cohort_task_case}
\setlength{\tabcolsep}{6pt}
\renewcommand{\arraystretch}{1.25}
\begin{tabular}{p{2.2cm} p{10.2cm}}
\toprule
\textbf{Term} & \textbf{Definition} \\
\midrule
\textbf{Task} &
A specific outcome that the model is evaluated for, such as malignancy diagnosis, HER2 status, or chronic gastritis severity grading. Defines \emph{what} is predicted. The same task is evaluated across multiple cohorts. \\
\addlinespace
\textbf{Cohort} &
The group of cases used in a particular evaluation, defined by a specific combination of hospital, study phase (retrospective or prospective), and specimen type (biopsy or resection). Defines \emph{whom} the task is evaluated on. For a given task, multiple cohorts may be evaluated separately across different hospitals or study settings.  \\

\addlinespace
\textbf{Case} &
A patient-level unit used for analysis and reporting.
Defines the unit of analysis at the patient level. A case may contain multiple slides or WSIs, whose predictions can be combined into a single case-level result. Multiple cases form a cohort.\\

\bottomrule
\end{tabular}
\end{table}

\begin{table}[!htbp]
\centering
\caption{\textbf{Dataset composition for downstream task evaluation.}}
\label{tab:hospital_dataset}
\renewcommand{\arraystretch}{1.5}
\begin{tabular}{lcc}
\toprule
\textbf{Dataset}& \textbf{Total Cases} & \textbf{Total Slides} \\
\midrule
Hospital H1     & 7,358  & 7,752  \\
Hospital H2     & 1,200   & 4,784  \\
Hospital H3     & 260   & 260   \\
Hospital H4     & 320   & 320   \\
Hospital H5     & 1,327 & 2,697 \\
Hospital H6     & 253 & 471 \\
Hospital H7     & 740 & 5,917 \\
Hospital H8     & 147 & 172 \\
Hospital H9     & 169 & 272 \\

\midrule
\textbf{Total} & \textbf{11,774} & \textbf{22,645} \\
\bottomrule
\end{tabular}
\end{table}

\FloatBarrier
\begin{table}[t]
    \centering
    \centering
    \footnotesize
    \setlength{\tabcolsep}{6pt}
    \caption{Performance on abnormal tissue diagnosis (Hospital H5). Metrics include AUROC, macro-ACC, specificity at 90\% sensitivity, positive-class NPV and PPV with 95\% CIs.}
    \label{tab:Normal_and_Abnormal_Tissue_Diagnosis}
    \begin{tabular}{lccccc}
        \toprule
        Model & AUROC & Macro-ACC & Spec at 90\% Sens & NPV & PPV \\
        \midrule
        \textbf{GRACE} & \textbf{0.9589 (0.9286-0.9821)} & 0.7600 (0.6832-0.8310) & \textbf{0.9204 (0.8194-0.9699)} & 0.8207 (0.6667-0.9394) & 0.9060 (0.8681-0.9394) \\
        Virchow2 & 0.9426 (0.9002-0.9754) & \textbf{0.8696 (0.8075-0.9247)} & 0.9039 (0.7191-0.9599) & 0.7665 (0.6382-0.8801) & 0.9535 (0.9252-0.9812) \\
        UNI & 0.9469 (0.9105-0.9747) & 0.8686 (0.8061-0.9262) & 0.8570 (0.7324-0.9465) & 0.7522 (0.6226-0.8751) & \textbf{0.9540 (0.9242-0.9800)} \\
        CONCH & 0.9467 (0.9068-0.9778) & 0.7926 (0.7248-0.8661) & 0.8564 (0.7793-0.9398) & \textbf{0.8593 (0.7390-0.9677)} & 0.9173 (0.8788-0.9511) \\
        \bottomrule
    \end{tabular}
\end{table}

\begin{table}[t]
\centering
\footnotesize
\setlength{\tabcolsep}{6pt}
\caption{Performance on \textit{H. pylori} infection diagnosis (Hospital H2). Metrics include AUROC, macro-ACC, specificity at 90\% sensitivity, and positive-class NPV/PPV with 95\% CIs.}
\label{tab:HP_Infection_Diagnosis}
\begin{tabular}{lccccc}
\toprule
Model & AUROC & Macro-ACC & Spec at 90\% Sens & NPV & PPV \\
\midrule
\textbf{GRACE} & \textbf{0.8980 (0.7753-0.9967)} & \textbf{0.7455 (0.6000-0.8857)} & \textbf{0.7607 (0.1405-0.9640)} & \textbf{0.7743 (0.5998-0.9231)} & \textbf{0.6620 (0.3000-1.0000)} \\
Virchow2 & 0.8740 (0.7427-0.9720) & 0.6886 (0.5429-0.8286) & 0.6727 (0.1405-0.9498) & 0.7179 (0.5484-0.8751) & 0.5690 (0.2000-1.0000) \\
UNI & 0.8562 (0.7197-0.9616) & 0.6857 (0.5429-0.8286) & 0.6236 (0.1303-0.9197) & 0.7046 (0.5385-0.8667) & 0.6127 (0.0000-1.0000) \\
CONCH & 0.8608 (0.7313-0.9686) & 0.7161 (0.5714-0.8571) & 0.6294 (0.1405-0.9131) & 0.7613 (0.5768-0.9201) & 0.5956 (0.2500-0.9000) \\
\bottomrule
\end{tabular}
\end{table}

\begin{table}[t]
\centering
\footnotesize
\setlength{\tabcolsep}{6pt}
\caption{Performance on \textit{H. pylori} infection diagnosis (Hospital H1). Metrics include AUROC, macro-ACC, specificity at 90\% sensitivity, and positive-class NPV/PPV with 95\% CIs.}
\label{tab:H_pylori_Infection_Diagnosis_External_H1}
\begin{tabular}{lccccc}
\toprule
Model & AUROC & Macro-ACC & Spec at 90\% Sens & NPV & PPV \\
\midrule
\textbf{GRACE} & \textbf{0.9072 (0.8717-0.9393)} & 0.7139 (0.6696-0.7632) & \textbf{0.7706 (0.6388-0.8797)} & 0.9478 (0.8813-1.0000) & 0.6665 (0.6140-0.7204) \\
Virchow2 & 0.8880 (0.8506-0.9221) & \textbf{0.7951 (0.7515-0.8392)} & 0.6688 (0.5819-0.7592) & 0.8760 (0.8085-0.9321) & \textbf{0.7594 (0.7061-0.8095)} \\
UNI & 0.8647 (0.8242-0.9022) & 0.7188 (0.6725-0.7661) & 0.6196 (0.4982-0.7292) & 0.8465 (0.7727-0.9243) & 0.6798 (0.6207-0.7386) \\
CONCH & 0.8959 (0.8541-0.9316) & 0.6408 (0.5906-0.6901) & 0.7559 (0.6120-0.8896) & \textbf{0.9644 (0.8750-1.0000)} & 0.6109 (0.5592-0.6635) \\
\bottomrule
\end{tabular}
\end{table}

\begin{table}[t]
\centering
\footnotesize
\setlength{\tabcolsep}{6pt}
\caption{Performance on diagnosis of autoimmune chronic gastritis with \textit{H. pylori} (Hospital H5). Metrics include AUROC, macro-ACC, specificity at 90\% sensitivity, and positive-class NPV/PPV with 95\% CIs.}
\label{tab:Autoimmune_Chronic_Gastritis_with_HP_Diagnosis}
\begin{tabular}{lccccc}
\toprule
Model & AUROC & Macro-ACC & Spec at 90\% Sens & NPV & PPV \\
\midrule
\textbf{GRACE} & \textbf{0.8854 (0.8156-0.9377)} & 0.6395 (0.5609-0.7304) & \textbf{0.6698 (0.4114-0.8562)} & 0.9161 (0.8809-0.9490) & 0.6481 (0.4000-0.8889) \\
Virchow2 & 0.8620 (0.7893-0.9238) & 0.6454 (0.5628-0.7350) & 0.6075 (0.4114-0.7993) & 0.9183 (0.8819-0.9545) & 0.5030 (0.2857-0.7273) \\
UNI & 0.8615 (0.7885-0.9170) & \textbf{0.6888 (0.5961-0.7813)} & 0.6312 (0.3311-0.8094) & \textbf{0.9314 (0.8965-0.9619)} & 0.3974 (0.2500-0.5556) \\
CONCH & 0.8602 (0.7760-0.9246) & 0.6409 (0.5582-0.7338) & 0.5995 (0.3579-0.8261) & 0.9173 (0.8832-0.9484) & \textbf{0.6944 (0.4286-0.9286)} \\
\bottomrule
\end{tabular}

\end{table}

\begin{table}[t]
\centering
\footnotesize
\setlength{\tabcolsep}{6pt}
\caption{Performance on \textit{H. pylori}-associated chronic gastritis diagnosis (Hospital H5). Metrics include AUROC, macro-ACC, specificity at 90\% sensitivity, and positive-class NPV/PPV with 95\% CIs.}
\label{tab:HP-Associated_Chronic_Gastritis_Diagnosis}
\begin{tabular}{lccccc}
\toprule
Model & AUROC & Macro-ACC & Spec at 90\% Sens & NPV & PPV \\
\midrule
\textbf{GRACE} & \textbf{0.9851 (0.9697-0.9959)} & 0.7524 (0.6575-0.8455) & \textbf{0.9600 (0.8863-0.9933)} & 0.9473 (0.9173-0.9727) & 0.8255 (0.6153-1.0000) \\
Virchow2 & 0.9449 (0.8784-0.9876) & \textbf{0.8418 (0.7509-0.9274)} & 0.8742 (0.6822-0.9699) & \textbf{0.9669 (0.9429-0.9876)} & 0.8291 (0.6522-0.9600) \\
UNI & 0.9707 (0.9486-0.9887) & 0.7001 (0.6042-0.7927) & 0.9114 (0.8094-0.9732) & 0.9371 (0.9059-0.9646) & \textbf{0.9173 (0.7333-1.0000)} \\
CONCH & 0.9696 (0.9480-0.9872) & 0.8035 (0.7068-0.8926) & 0.9230 (0.8361-0.9732) & 0.9590 (0.9325-0.9798) & 0.7372 (0.5499-0.9130) \\
\bottomrule
\end{tabular}
\end{table}

\begin{table}[t]
\centering
\footnotesize
\setlength{\tabcolsep}{6pt}
\caption{Performance on chronic gastritis diagnosis (Hospital H1). Metrics include AUROC, macro-ACC, specificity at 90\% sensitivity, and positive-class NPV/PPV with 95\% CIs.}
\label{tab:Chronic_Gastritis_Diagnosis}
\begin{tabular}{lccccc}
\toprule
Model & AUROC & Macro-ACC & Spec at 90\% Sens & NPV & PPV \\
\midrule
\textbf{GRACE} & 0.9707 (0.9590-0.9807) & \textbf{0.9278 (0.9122-0.9430)} & \textbf{0.9321 (0.8997-0.9599)} & \textbf{0.8987 (0.8684-0.9284)} & 0.9427 (0.9259-0.9595) \\
Virchow2 & \textbf{0.9709 (0.9597-0.9802)} & 0.9214 (0.9050-0.9367) & 0.9248 (0.8896-0.9532) & 0.8943 (0.8606-0.9251) & 0.9350 (0.9181-0.9525) \\
UNI & 0.9646 (0.9506-0.9758) & 0.9216 (0.9050-0.9376) & 0.9276 (0.8930-0.9565) & 0.8739 (0.8390-0.9074) & \textbf{0.9468 (0.9303-0.9629)} \\
CONCH & 0.9664 (0.9553-0.9756) & 0.9196 (0.9032-0.9340) & 0.9089 (0.8729-0.9431) & 0.8895 (0.8563-0.9201) & 0.9345 (0.9155-0.9518) \\
\bottomrule

\end{tabular}
\end{table}

\begin{table}[t]
\centering
\footnotesize
\setlength{\tabcolsep}{6pt}
\caption{Performance on chronic gastritis diagnosis (prospective validation on Hospital H1). Metrics include AUROC, macro-ACC, specificity at 90\% sensitivity, and positive-class NPV/PPV with 95\% CIs.}
\label{tab:Chronic_Gastritis_Diagnosis_Prospective_H1}
\begin{tabular}{lccccc}
\toprule
Model & AUROC & Macro-ACC & Spec at 90\% Sens & NPV & PPV \\
\midrule
\textbf{GRACE} & \textbf{0.9716 (0.9613-0.9807)} & 0.9089 (0.8916-0.9274) & 0.9310 (0.8963-0.9565) & \textbf{0.9689 (0.9515-0.9853)} & 0.8685 (0.8408-0.8962) \\
Virchow2 & 0.9702 (0.9601-0.9792) & \textbf{0.9220 (0.9052-0.9379)} & \textbf{0.9339 (0.9064-0.9565)} & 0.9586 (0.9375-0.9769) & 0.8945 (0.8667-0.9199) \\
UNI & 0.9677 (0.9562-0.9772) & 0.9192 (0.9021-0.9358) & 0.9337 (0.9030-0.9565) & 0.9443 (0.9220-0.9652) & \textbf{0.8984 (0.8721-0.9227)} \\
CONCH & 0.9645 (0.9526-0.9750) & 0.8949 (0.8758-0.9137) & 0.9187 (0.8863-0.9465) & 0.9684 (0.9501-0.9845) & 0.8467 (0.8183-0.8746) \\
\bottomrule
\end{tabular}
\vspace{-4pt}
\end{table}

\begin{table}[t]
\centering
\footnotesize
\setlength{\tabcolsep}{6pt}
\caption{Performance on chronic gastritis diagnosis (prospective validation on Hospital H9). Metrics include AUROC, macro-ACC, specificity at 90\% sensitivity, and positive-class NPV/PPV with 95\% CIs.}
\label{tab:Chronic_Gastritis_Diagnosis_Prospective_H9}
\begin{tabular}{lccccc}
\toprule
Model & AUROC & Macro-ACC & Spec at 90\% Sens & NPV & PPV \\
\midrule
\textbf{GRACE} & \textbf{0.9680 (0.9377-0.9919)} & \textbf{0.9331 (0.8932-0.9644)} & \textbf{0.9446 (0.8796-0.9900)} & \textbf{0.7186 (0.5918-0.8400)} & 0.9920 (0.9741-1.0000) \\
Virchow2 & 0.9621 (0.9349-0.9844) & 0.8933 (0.8415-0.9356) & 0.9080 (0.8026-0.9666) & 0.6245 (0.4884-0.7429) & 0.9827 (0.9561-1.0000) \\
UNI & 0.9678 (0.9372-0.9914) & 0.8062 (0.7659-0.8477) & 0.9205 (0.8294-0.9833) & 0.4202 (0.3187-0.5161) & \textbf{1.0000 (1.0000-1.0000)} \\
CONCH & 0.9430 (0.9053-0.9752) & 0.8755 (0.8173-0.9246) & 0.8398 (0.6652-0.9331) & 0.6071 (0.4754-0.7359) & 0.9737 (0.9417-1.0000) \\
\bottomrule
\end{tabular}
\vspace{-4pt}
\end{table}

\begin{table}[t]
\centering
\footnotesize
\setlength{\tabcolsep}{6pt}
\caption{Performance on chronic gastritis severity grading (Hospital H1). Metrics include macro-AUC, macro-ACC, specificity at 90\% sensitivity, and positive-class NPV/PPV with 95\% CIs.}
\label{tab:Chronic_Gastritis_Severity_Grading}
\begin{tabular}{lccccc}
\toprule
Model & Macro-AUC & Macro-ACC & Spec at 90\% Sens & NPV & PPV \\
\midrule
\textbf{GRACE} & \textbf{0.9218 (0.9059-0.9368)} & \textbf{0.8030 (0.7753-0.8301)} & \textbf{0.8021 (0.7458-0.8528)}& 0.8030 (0.7558-0.8476)  & 0.9082 (0.8788-0.9349) \\
Virchow2 & 0.9165 (0.8994-0.9318) & 0.7869 (0.7548-0.8178) & 0.7961 (0.7391-0.8462) & \textbf{0.8087 (0.7601-0.8537)}& 0.8810 (0.8505-0.9099)  \\
UNI & 0.9195 (0.9039-0.9344) & 0.7887 (0.7589-0.8178) & 0.7788 (0.7258-0.8294)  & 0.7630 (0.7147-0.8088)& \textbf{0.9263 (0.9002-0.9495)} \\
CONCH & 0.9128 (0.8968-0.9285) & 0.7837 (0.7548-0.8137) & 0.7644 (0.7156-0.8194)  & 0.7696 (0.7239-0.8136) & 0.9028 (0.8724-0.9298)\\
\bottomrule
\end{tabular}
\vspace{2pt}

\begin{minipage}{0.97\linewidth}
\raggedright
\footnotesize
\textit{For severity grading in chronic gastritis, NPV and PPV were calculated with moderate and severe cases treated as the positive class and mild cases treated as the negative class. Macro-AUC and macro-ACC were calculated across all severity grades.}
\end{minipage}
\vspace{-4pt}
\end{table}

\begin{table}[t]
\centering
\footnotesize
\setlength{\tabcolsep}{6pt}
\caption{Performance on chronic gastritis severity grading (prospective validation on Hospital H1). Metrics include macro-AUC, macro-ACC, specificity at 90\% sensitivity, and positive-class NPV/PPV with 95\% CIs.}
\label{tab:Chronic_Gastritis_Severity_Grading_Prospective_H1}
\begin{tabular}{lccccc}
\toprule
Model & Macro-AUC & Macro-ACC & Spec at 90\% Sens & NPV & PPV \\
\midrule
\textbf{GRACE} & \textbf{0.9103 (0.8728-0.9428)} & \textbf{0.7773 (0.7056-0.8389)} & \textbf{0.7806 (0.6756-0.8629)} & \textbf{0.7954 (0.7065-0.8750)}& 0.9307 (0.8658-0.9783)  \\
Virchow2 & 0.9079 (0.8741-0.9380) & 0.7514 (0.6889-0.8113) & 0.7529 (0.6355-0.8495) & 0.7548 (0.6697-0.8351)& 0.9627 (0.9176-1.0000)  \\
UNI & 0.9001 (0.8664-0.9305) & 0.7386 (0.6722-0.8000) & 0.7281 (0.6288-0.8161) & 0.7744 (0.6860-0.8589)& 0.8970 (0.8222-0.9545)  \\
CONCH & 0.8910 (0.8538-0.9264) & 0.7510 (0.6833-0.8111) & 0.7264 (0.5886-0.8194) & 0.7319 (0.6486-0.8163) & \textbf{0.9733 (0.9286-1.0000)}  \\
\bottomrule
\end{tabular}
\vspace{-4pt}
\end{table}

\begin{table}[t]
\centering
\footnotesize
\setlength{\tabcolsep}{6pt}
\caption{Performance on chronic gastritis severity grading (prospective validation on Hospital H2). Metrics include macro-AUC, macro-ACC, specificity at 90\% sensitivity, and positive-class NPV/PPV with 95\% CIs.}
\label{tab:Chronic_Gastritis_Severity_Grading_Prospective_H2}
\begin{tabular}{lccccc}
\toprule
Model & Macro-AUC & Macro-ACC & Spec at 90\% Sens & NPV & PPV \\
\midrule
\textbf{GRACE} & \textbf{0.9194 (0.6326-0.9829)} & \textbf{0.7227 (0.5522-0.9296)} & \textbf{0.9275 (0.8629-0.9699)} &\textbf{0.9196 (0.8864-0.9505)} & \textbf{1.0000 (1.0000-1.0000)}  \\
Virchow2 & 0.9186 (0.6364-0.9805) & 0.5324 (0.5000-0.5659) & 0.9223 (0.8461-0.9732) & 0.8834 (0.8454-0.9184)& \textbf{1.0000 (1.0000-1.0000)}  \\
UNI & 0.9054 (0.6231-0.9721) & 0.6175 (0.5467-0.8860) & 0.8837 (0.7992-0.9498) & 0.9068 (0.8723-0.9391)& \textbf{1.0000 (1.0000-1.0000)}  \\
CONCH & 0.9173 (0.6314-0.9817) & 0.5145 (0.4806-0.5506) & 0.9253 (0.8594-0.9666)  & 0.8693 (0.8304-0.9043) & \textbf{1.0000 (1.0000-1.0000)} \\
\bottomrule
\end{tabular}
\vspace{-4pt}
\end{table}

\begin{table}[t]
\centering
\footnotesize
\setlength{\tabcolsep}{6pt}
\caption{Performance on diagnosis of acute activity in chronic gastritis (Hospital H1). Metrics include AUROC, macro-ACC, specificity at 90\% sensitivity, and positive-class NPV/PPV with 95\% CIs.}
\label{tab:diagnosis_of_Acute_Activity_in_Chronic_Gastritis}
\begin{tabular}{lccccc}
\toprule
Model & AUROC & Macro-ACC & Spec at 90\% Sens & NPV & PPV \\
\midrule
\textbf{GRACE} & \textbf{0.9203 (0.8796-0.9561)} & \textbf{0.9391 (0.9142-0.9616)} & \textbf{0.7305 (0.6288-0.8829)} & 0.8045 (0.6364-0.9412) & \textbf{0.9472 (0.9240-0.9688)} \\
Virchow2 & 0.9006 (0.8574-0.9437) & 0.9301 (0.9052-0.9526) & 0.6853 (0.5452-0.8329) & 0.7435 (0.5554-0.9091) & 0.9403 (0.9155-0.9617) \\
UNI & 0.8981 (0.8506-0.9414) & 0.9236 (0.8984-0.9458) & 0.6808 (0.5351-0.7893) & \textbf{0.8378 (0.6150-1.0000)} & 0.9254 (0.9014-0.9482) \\
CONCH & 0.9121 (0.8741-0.9498) & 0.9181 (0.8894-0.9413) & 0.7294 (0.6054-0.8562) & 0.6595 (0.4286-0.8889) & 0.9293 (0.9049-0.9515) \\
\bottomrule
\end{tabular}
\vspace{-4pt}
\end{table}

\begin{table}[t]
\centering
\footnotesize
\setlength{\tabcolsep}{6pt}
\caption{Performance on diagnosis of acute activity in chronic gastritis (prospective validation on Hospital H7). Metrics include AUROC, macro-ACC, specificity at 90\% sensitivity, and positive-class NPV/PPV with 95\% CIs.}
\label{tab:Diagnosis_of_Acute_Activity_in_Chronic_Gastritis_Prospective_H7}
\begin{tabular}{lccccc}
\toprule
Model & AUROC & Macro-ACC & Spec at 90\% Sens & NPV & PPV \\
\midrule
\textbf{GRACE} & \textbf{0.8909 (0.8518-0.9226)} & 0.5262 (0.5079-0.5476) & \textbf{0.7352 (0.6354-0.8227)} & \textbf{1.0000 (1.0000-1.0000)} & 0.6095 (0.5582-0.6636) \\
Virchow2 & 0.8169 (0.7693-0.8590) & \textbf{0.6630 (0.6118-0.7101)} & 0.5279 (0.4314-0.6389) & 0.7001 (0.6000-0.7975) & \textbf{0.7024 (0.6445-0.7610)} \\
UNI & 0.8640 (0.8253-0.9013) & 0.5701 (0.5391-0.6054) & 0.6500 (0.5518-0.7425) & 0.8170 (0.6520-0.9546) & 0.6336 (0.5784-0.6854) \\
CONCH & 0.8648 (0.8251-0.9029) & 0.5000 (0.5000-0.5000) & 0.6371 (0.5385-0.7258) &  - & 0.5973 (0.5481-0.6506) \\
\bottomrule
\end{tabular}
\vspace{2pt}

\begin{minipage}{0.97\linewidth}
\raggedright
\footnotesize
\textit{Unless otherwise specified, PPV or NPV is not reported in cases where predictions are restricted to a single class (all positive or all negative), for which the corresponding metric is undefined.}
\end{minipage}
\vspace{-4pt}
\end{table}

\begin{table}[t]
\centering
\footnotesize
\setlength{\tabcolsep}{6pt}
\caption{Performance on diagnosis of acute activity in chronic gastritis (prospective validation on Hospital H8). Metrics include AUROC, macro-ACC, specificity at 90\% sensitivity, and positive-class NPV/PPV with 95\% CIs.}
\label{tab:Diagnosis_of_Acute_Activity_in_Chronic_Gastritis_Prospective_H8}
\begin{tabular}{lccccc}
\toprule
Model & AUROC & Macro-ACC & Spec at 90\% Sens & NPV & PPV \\
\midrule
\textbf{GRACE} & \textbf{0.9217 (0.8038-1.0000)} & \textbf{0.8738 (0.7667-0.9706)} & \textbf{0.7727 (0.3742-0.9967)} & \textbf{1.0000 (1.0000-1.0000)} &\textbf{0.7661 (0.5385-0.9413)} \\
Virchow2 & 0.9132 (0.7895-1.0000) & 0.5325 (0.5000-0.6071) & 0.7232 (0.3512-0.9967) & \textbf{1.0000 (1.0000-1.0000)} & 0.4696 (0.2857-0.6538) \\
UNI & 0.8239 (0.6363-0.9685) & 0.6147 (0.4789-0.7500) & 0.5146 (0.0735-0.9231) & 0.8274 (0.5000-1.0000) & 0.5177 (0.3043-0.7202) \\
CONCH & 0.9075 (0.7777-0.9905) & 0.5000 (0.5000-0.5000) & 0.7350 (0.3539-0.9967) & - & 0.4451 (0.2759-0.6207)  \\
\bottomrule
\end{tabular}
\vspace{-4pt}
\end{table}

\begin{table}[t]
\centering
\footnotesize
\setlength{\tabcolsep}{6pt}
\caption{Performance on diagnosis of acute activity in chronic gastritis (prospective validation on Hospital H9). Metrics include AUROC, macro-ACC, specificity at 90\% sensitivity, and positive-class NPV/PPV with 95\% CIs.}
\label{tab:Diagnosis_of_Acute_Activity_in_Chronic_Gastritis_Prospective_H9}
\begin{tabular}{lccccc}
\toprule
Model & AUROC & Macro-ACC & Spec at 90\% Sens & NPV & PPV \\
\midrule
\textbf{GRACE} & \textbf{0.8968 (0.8264-0.9562)} & \textbf{0.6974 (0.6369-0.7560)} & 0.7052 (0.4648-0.9064) & 0.9729 (0.9091-1.0000) & \textbf{0.4052 (0.3038-0.5068)} \\
Virchow2 & 0.8928 (0.8053-0.9579) & 0.5033 (0.4659-0.5330) & \textbf{0.7246 (0.4848-0.9130)} & 0.7636 (0.2500-1.0000) & 0.2897 (0.2136-0.3772) \\
UNI & 0.8423 (0.7425-0.9265) & 0.5000 (0.5000-0.5000) & 0.5838 (0.2742-0.8596) &  - & 0.2900 (0.2165-0.3750) \\
CONCH & 0.8363 (0.7424-0.9224) & 0.5241 (0.5057-0.5476) & 0.5383 (0.2341-0.8595) & \textbf{1.0000 (1.0000-1.0000)} & 0.3021 (0.2203-0.3879) \\
\bottomrule
\end{tabular}
\vspace{-4pt}
\end{table}

\begin{table}[t]
\centering
\footnotesize
\setlength{\tabcolsep}{6pt}
\caption{Performance on acute activity severity grading (Hospital H1). Metrics include macro-AUC, macro-ACC, specificity at 90\% sensitivity, and positive-class NPV/PPV with 95\% CIs.}
\label{tab:Acute_Activity_Severity_Grading}
\begin{tabular}{lccccc}
\toprule
Model & Macro-AUC & Macro-ACC & Spec at 90\% Sens & NPV & PPV \\
\midrule
\textbf{GRACE} & \textbf{0.9439 (0.9279-0.9582)} & \textbf{0.8550 (0.8288-0.8812)} & 0.8228 (0.7591-0.8829) & 0.9033 (0.8747-0.9292) & 0.8617 (0.8131-0.9027)  \\
Virchow2 & 0.9432 (0.9284-0.9572) & 0.8490 (0.8209-0.8764) & 0.8232 (0.7592-0.8797) & \textbf{0.9161 (0.8900-0.9416)}& 0.8255 (0.7729-0.8756)  \\
UNI & 0.9388 (0.9229-0.9536) & 0.8228 (0.7940-0.8510) & 0.8319 (0.7892-0.8729)& 0.8670 (0.8365-0.8970)  & \textbf{0.8890 (0.8418-0.9341)} \\
CONCH & 0.9423 (0.9268-0.9566) & 0.8146 (0.7845-0.8447) & \textbf{0.8358 (0.7692-0.8863)} & 0.9051 (0.8762-0.9314) & 0.8272 (0.7731-0.8757)  \\
\bottomrule
\end{tabular}
\vspace{2pt}

\begin{minipage}{0.97\linewidth}
\raggedright
\footnotesize
\textit{For inflammatory activity severity grading, NPV and PPV were calculated with moderate or severe acute activity defined as the positive class and mild acute activity defined as the negative class.}
\end{minipage}
\vspace{-4pt}
\end{table}

\begin{table}[t]
\centering
\footnotesize
\setlength{\tabcolsep}{6pt}
\caption{Performance on acute activity severity grading (prospective validation on Hospital H1). Metrics include macro-AUC, macro-ACC, specificity at 90\% sensitivity, and positive-class NPV/PPV with 95\% CIs.}
\label{tab:Acute_Activity_Severity_Grading_Prospective_H1}
\begin{tabular}{lccccc}
\toprule
Model & Macro-AUC & Macro-ACC & Spec at 90\% Sens & NPV & PPV \\
\midrule
\textbf{GRACE} & \textbf{0.9305 (0.9089-0.9495)} & \textbf{0.8403 (0.8020-0.8738)} & \textbf{0.8559 (0.8127-0.8963)} & 0.9291 (0.8982-0.9583) & \textbf{0.8170 (0.7459-0.8830)}  \\
Virchow2 & 0.9216 (0.9002-0.9430) & 0.8339 (0.7946-0.8688) & 0.8259 (0.7793-0.8729) & 0.9652 (0.9400-0.9848) & 0.7681 (0.6950-0.8357) \\
UNI & 0.9093 (0.8864-0.9303) & 0.7865 (0.7450-0.8267) & 0.7820 (0.7123-0.8495) & \textbf{0.9668 (0.9430-0.9851)} & 0.7936 (0.7266-0.8582) \\
CONCH & 0.9178 (0.8949-0.9394) & 0.7623 (0.7178-0.8045) & 0.8032 (0.7525-0.8495) & 0.9468 (0.9179-0.9737) & 0.7366 (0.6643-0.8069)  \\
\bottomrule
\end{tabular}
\vspace{-4pt}
\end{table}

\begin{table}[t]
\centering
\footnotesize
\setlength{\tabcolsep}{6pt}
\caption{Performance on acute activity severity grading (prospective validation on Hospital H7). Metrics include macro-AUC, macro-ACC, specificity at 90\% sensitivity, and positive-class NPV/PPV with 95\% CIs.}
\label{tab:Acute_Activity_Severity_Grading_Prospective_H7}
\begin{tabular}{lccccc}
\toprule
Model & Macro-AUC & Macro-ACC & Spec at 90\% Sens & NPV & PPV \\
\midrule
\textbf{GRACE} & \textbf{0.9360 (0.9043-0.9635)} & 0.6852 (0.5409-0.8364) & \textbf{0.8537 (0.7759-0.9164)} & 0.9405 (0.8928-0.9823) & \textbf{0.8144 (0.7142-0.9000)}  \\
Virchow2 & 0.9347 (0.9011-0.9663) & 0.6452 (0.5786-0.7599) & 0.8407 (0.7726-0.8964) & \textbf{1.0000 (1.0000-1.0000)} & 0.7277 (0.6267-0.8140) \\
UNI & 0.9187 (0.8802-0.9515) & \textbf{0.7755 (0.6554-0.8611)} & 0.7798 (0.6990-0.8696) & 0.9187 (0.8614-0.9691) & 0.7712 (0.6667-0.8677)  \\
CONCH & 0.9217 (0.8854-0.9531) & 0.7687 (0.6419-0.8580) & 0.7930 (0.7124-0.8863) & 0.9564 (0.9100-0.9899)& 0.7627 (0.6622-0.8571)  \\
\bottomrule
\end{tabular}
\vspace{-4pt}
\end{table}

\begin{table}[t]
\centering
\footnotesize
\setlength{\tabcolsep}{6pt}
\caption{Performance on atrophy diagnosis in chronic gastritis (Hospital H1). Metrics include AUROC, macro-ACC, specificity at 90\% sensitivity, and positive-class NPV/PPV with 95\% CIs.}
\label{tab:diagnosis_of_Atrophy_in_Chronic_Gastritis}
\begin{tabular}{lccccc}
\toprule
Model & AUROC & Macro-ACC & Spec at 90\% Sens & NPV & PPV \\
\midrule
\textbf{GRACE} & \textbf{0.8960 (0.8720-0.9187)} & \textbf{0.8100 (0.7804-0.8398)} & \textbf{0.6864 (0.6254-0.7492)} & 0.7385 (0.6885-0.7862) & \textbf{0.8761 (0.8402-0.9078)} \\
Virchow2 & 0.8838 (0.8573-0.9080) & 0.8072 (0.7774-0.8368) & 0.6741 (0.6020-0.7492) & \textbf{0.7652 (0.7132-0.8153)} & 0.8391 (0.8030-0.8747) \\
UNI & 0.8897 (0.8646-0.9120) & 0.8097 (0.7819-0.8383) & 0.6549 (0.5819-0.7592) & 0.7481 (0.7013-0.7945) & 0.8633 (0.8285-0.8962) \\
CONCH & 0.8803 (0.8540-0.9039) & 0.7974 (0.7671-0.8294) & 0.6421 (0.5753-0.7191) & 0.7575 (0.7020-0.8085) & 0.8252 (0.7872-0.8608) \\
\bottomrule
\end{tabular}
\vspace{-4pt}
\end{table}

\begin{table}[t]
\centering
\footnotesize
\setlength{\tabcolsep}{6pt}
\caption{Performance on atrophy diagnosis in chronic gastritis (prospective validation on Hospital H1). Metrics include AUROC, macro-ACC, specificity at 90\% sensitivity, and positive-class NPV/PPV with 95\% CIs.}
\label{tab:Diagnosis_of_Atrophy_in_Chronic_Gastritis_Prospective_H1}
\begin{tabular}{lccccc}
\toprule
Model & AUROC & Macro-ACC & Spec at 90\% Sens & NPV & PPV \\
\midrule
\textbf{GRACE} & \textbf{0.9015 (0.8698-0.9288)} & 0.8266 (0.7911-0.8587) & \textbf{0.7526 (0.6355-0.8395)} & 0.8564 (0.8082-0.8996) & 0.8005 (0.7536-0.8502) \\
Virchow2 & 0.8943 (0.8628-0.9239) & 0.7829 (0.7405-0.8186) & 0.7382 (0.6421-0.8161) & \textbf{0.8713 (0.8206-0.9209)} & 0.7272 (0.6786-0.7767) \\
UNI & 0.9002 (0.8701-0.9255) & 0.8012 (0.7658-0.8354) & 0.7346 (0.6254-0.8261) & 0.8659 (0.8177-0.9096) & 0.7568 (0.7050-0.8077) \\
CONCH & 0.8924 (0.8603-0.9239) & \textbf{0.8372 (0.8038-0.8713)} & 0.7236 (0.6120-0.8294) & 0.8561 (0.8125-0.8996) & \textbf{0.8208 (0.7692-0.8680)} \\
\bottomrule
\end{tabular}
\vspace{-4pt}
\end{table}

\begin{table}[t]
\centering
\footnotesize
\setlength{\tabcolsep}{6pt}
\caption{Performance on atrophy diagnosis in chronic gastritis (prospective validation on Hospital H2). Metrics include AUROC, macro-ACC, specificity at 90\% sensitivity, and positive-class NPV/PPV with 95\% CIs.}
\label{tab:Diagnosis_of_Atrophy_in_Chronic_Gastritis_Prospective_H2}
\begin{tabular}{lccccc}
\toprule
Model & AUROC & Macro-ACC & Spec at 90\% Sens & NPV & PPV \\
\midrule
\textbf{GRACE} & \textbf{0.9121 (0.8759-0.9438)} & \textbf{0.8106 (0.7806-0.8410)} & 0.7711 (0.6521-0.8863) & \textbf{1.0000 (1.0000-1.0000)} & \textbf{0.3127 (0.2388-0.3910)} \\
Virchow2 & 0.9052 (0.8712-0.9353) & 0.7421 (0.7132-0.7727) & \textbf{0.8012 (0.6957-0.8730)} & \textbf{1.0000 (1.0000-1.0000)} & 0.2508 (0.1893-0.3155) \\
UNI & 0.9117 (0.8743-0.9439) & 0.7225 (0.6921-0.7520) & 0.7882 (0.6120-0.8796) & \textbf{1.0000 (1.0000-1.0000)} & 0.2369 (0.1792-0.3005) \\
CONCH & 0.8729 (0.8267-0.9141) & 0.7401 (0.7020-0.7726) & 0.6666 (0.4649-0.8127) & 0.9923 (0.9743-1.0000) & 0.2534 (0.1887-0.3192) \\
\bottomrule
\end{tabular}
\vspace{-4pt}
\end{table}

\begin{table}[t]
\centering
\footnotesize
\setlength{\tabcolsep}{6pt}
\caption{Performance on atrophy diagnosis in chronic gastritis (prospective validation on Hospital H8). Metrics include AUROC, macro-ACC, specificity at 90\% sensitivity, and positive-class NPV/PPV with 95\% CIs.}
\label{tab:Diagnosis_of_Atrophy_in_Chronic_Gastritis_Prospective_H8}
\begin{tabular}{lccccc}
\toprule
Model & AUROC & Macro-ACC & Spec at 90\% Sens & NPV & PPV \\
\midrule
\textbf{GRACE} & \textbf{0.9024 (0.7840-0.9861)} & \textbf{0.8214 (0.6825-0.9412)} & \textbf{0.7005 (0.3311-0.9431)} & 0.9309 (0.7691-1.0000) & \textbf{0.6666 (0.4165-0.9168)} \\
Virchow2 & 0.8783 (0.7411-0.9789) & 0.8126 (0.6944-0.9118) & 0.6953 (0.2843-0.9431) & \textbf{1.0000 (1.0000-1.0000)} & 0.6121 (0.3683-0.8333) \\
UNI & 0.8329 (0.6699-0.9665) & 0.7197 (0.5609-0.8693) & 0.6204 (0.0000-0.9565) & 0.8630 (0.6667-1.0000) & 0.5606 (0.3124-0.8000) \\
CONCH & 0.8773 (0.7399-0.9800) & 0.7734 (0.6180-0.9063) & 0.6487 (0.3311-0.9432) & 0.8725 (0.6923-1.0000) & 0.6421 (0.3750-0.8889) \\
\bottomrule
\end{tabular}
\vspace{-4pt}
\end{table}

\begin{table}[t]
\centering
\footnotesize
\setlength{\tabcolsep}{6pt}
\caption{Performance on atrophy diagnosis in chronic gastritis (prospective validation on Hospital H9). Metrics include AUROC, macro-ACC, specificity at 90\% sensitivity, and positive-class NPV/PPV with 95\% CIs.}
\label{tab:Diagnosis_of_Atrophy_in_Chronic_Gastritis_Prospective_H9}
\begin{tabular}{lccccc}
\toprule
Model & AUROC & Macro-ACC & Spec at 90\% Sens & NPV & PPV \\
\midrule
\textbf{GRACE} & \textbf{0.9269 (0.8788-0.9653)} & \textbf{0.8692 (0.8097-0.9198)} & \textbf{0.8156 (0.6355-0.9331)} & 0.9707 (0.9219-1.0000) & \textbf{0.6928 (0.5689-0.8095)} \\
Virchow2 & 0.9232 (0.8744-0.9664) & 0.8385 (0.7804-0.8941) & 0.8052 (0.6019-0.9264) & 0.9683 (0.9153-1.0000) & 0.6382 (0.5156-0.7538) \\
UNI & 0.9109 (0.8518-0.9615) & 0.7053 (0.6459-0.7641) & 0.7730 (0.5385-0.9031) & \textbf{0.9726 (0.9090-1.0000)} & 0.4708 (0.3777-0.5700) \\
CONCH & 0.8983 (0.8444-0.9479) & 0.8458 (0.7835-0.9043) & 0.7406 (0.5351-0.8763) & 0.9412 (0.8788-0.9865) & 0.6856 (0.5645-0.8000) \\
\bottomrule
\end{tabular}
\vspace{-4pt}
\end{table}

\begin{table}[t]
\centering
\footnotesize
\setlength{\tabcolsep}{6pt}
\caption{Performance on atrophy severity grading (Hospital H1). Metrics include AUROC, macro-ACC, specificity at 90\% sensitivity, and positive-class NPV/PPV with 95\% CIs.}
\label{tab:Atrophy_Severity_Grading}
\begin{tabular}{lccccc}
\toprule
Model & AUROC & Macro-ACC & Spec at 90\% Sens & NPV & PPV \\
\midrule
\textbf{GRACE} & \textbf{0.8844 (0.8257-0.9392)} & \textbf{0.8108 (0.7391-0.8696)} & \textbf{0.7029 (0.4983-0.8930)} & \textbf{0.8412 (0.7664-0.9048)} & \textbf{0.6959 (0.5357-0.8571)} \\
Virchow2 & 0.8706 (0.8020-0.9316) & 0.7538 (0.6812-0.8261) & 0.6584 (0.4013-0.8329) & 0.7659 (0.6897-0.8404) & 0.6429 (0.3885-0.9000) \\
UNI & 0.8629 (0.7983-0.9206) & 0.7743 (0.7027-0.8478) & 0.6206 (0.3411-0.8930) & 0.8171 (0.7407-0.8868) & 0.6203 (0.4399-0.7917) \\
CONCH & 0.8840 (0.8278-0.9382) & 0.7898 (0.7246-0.8551) & 0.6856 (0.4883-0.8897) & 0.8086 (0.7315-0.8739) & 0.6941 (0.4996-0.8824) \\
\bottomrule
\end{tabular}
\vspace{-4pt}
\end{table}

\begin{table}[t]
\centering
\footnotesize
\setlength{\tabcolsep}{6pt}
\caption{Performance on polyp diagnosis (Hospital H1). Metrics include AUROC, macro-ACC, specificity at 90\% sensitivity, and positive-class NPV/PPV with 95\% CIs.}
\label{tab:Polyp_Diagnosis}
\begin{tabular}{lccccc}
\toprule
Model & AUROC & Macro-ACC & Spec at 90\% Sens & NPV & PPV \\
\midrule
\textbf{GRACE} & \textbf{0.9807 (0.9735-0.9869)} & 0.9145 (0.8984-0.9307) & 0.9491 (0.9298-0.9666) & 0.9026 (0.8816-0.9222) & \textbf{0.9515 (0.9237-0.9755)} \\
Virchow2 & 0.9793 (0.9706-0.9870) & 0.9299 (0.9151-0.9446) & 0.9519 (0.9298-0.9699) & \textbf{0.9747 (0.9621-0.9853)} & 0.8440 (0.8069-0.8776) \\
UNI & 0.9798 (0.9699-0.9880) & \textbf{0.9301 (0.9141-0.9446)} & \textbf{0.9536 (0.9264-0.9699)} & 0.9736 (0.9606-0.9848) & 0.8430 (0.8048-0.8790) \\
CONCH & 0.9739 (0.9635-0.9829) & 0.9166 (0.8984-0.9335) & 0.9455 (0.9164-0.9632) & 0.9744 (0.9621-0.9856) & 0.8121 (0.7699-0.8555) \\
\bottomrule
\end{tabular}
\vspace{-4pt}
\end{table}

\begin{table}[t]
\centering
\footnotesize
\setlength{\tabcolsep}{6pt}
\caption{Performance on polyp diagnosis (prospective validation on Hospital H1). Metrics include AUROC, macro-ACC, specificity at 90\% sensitivity, and positive-class NPV/PPV with 95\% CIs.}
\label{tab:Polyp_Diagnosis_Prospective_H1}
\begin{tabular}{lccccc}
\toprule
Model & AUROC & Macro-ACC & Spec at 90\% Sens & NPV & PPV \\
\midrule
\textbf{GRACE} & \textbf{0.9812 (0.9701-0.9897)} & 0.8438 (0.8134-0.8707) & \textbf{0.9649 (0.9398-0.9833)} & 0.6770 (0.6205-0.7340) & \textbf{0.9968 (0.9902-1.0000)} \\
Virchow2 & 0.9807 (0.9710-0.9891) & 0.9458 (0.9280-0.9624) & 0.9624 (0.9365-0.9799) & 0.8878 (0.8424-0.9277) & 0.9771 (0.9622-0.9900) \\
UNI & 0.9797 (0.9683-0.9892) & \textbf{0.9482 (0.9296-0.9656)} & 0.9622 (0.9398-0.9799) & \textbf{0.9078 (0.8667-0.9458)} & 0.9681 (0.9489-0.9833) \\
CONCH & 0.9758 (0.9644-0.9854) & 0.9019 (0.8789-0.9247) & 0.9423 (0.9130-0.9666) & 0.7829 (0.7310-0.8327) & 0.9833 (0.9688-0.9946) \\
\bottomrule
\end{tabular}
\vspace{-4pt}
\end{table}

\begin{table}[t]
\centering
\footnotesize
\setlength{\tabcolsep}{6pt}
\caption{Performance on polyp diagnosis (prospective validation on Hospital H7). Metrics include AUROC, macro-ACC, specificity at 90\% sensitivity, and positive-class NPV/PPV with 95\% CIs.}
\label{tab:Polyp_Diagnosis_Prospective_H7}
\begin{tabular}{lccccc}
\toprule
Model & AUROC & Macro-ACC & Spec at 90\% Sens & NPV & PPV \\
\midrule
\textbf{GRACE} & \textbf{0.9209 (0.8563-0.9737)} & \textbf{0.9051 (0.8440-0.9570)} & 0.8518 (0.5115-0.9732) & 0.9804 (0.9639-0.9943) & \textbf{0.8335 (0.7222-0.9376)} \\
Virchow2 & 0.9199 (0.8575-0.9704) & 0.8776 (0.8200-0.9268) & 0.8114 (0.5284-0.9498) & 0.9818 (0.9650-0.9940) & 0.5067 (0.3971-0.6167) \\
UNI & 0.9192 (0.8611-0.9694) & 0.8767 (0.8272-0.9249) & 0.8352 (0.4749-0.9699) & \textbf{0.9844 (0.9693-0.9969)} & 0.4548 (0.3511-0.5679) \\
CONCH & 0.9184 (0.8490-0.9759) & 0.8819 (0.8301-0.9306) & \textbf{0.8606 (0.5986-0.9666)} & \textbf{0.9844 (0.9712-0.9969)} & 0.4720 (0.3662-0.5833) \\
\bottomrule
\end{tabular}
\vspace{-4pt}
\end{table}

\begin{table}[t]
\centering
\footnotesize
\setlength{\tabcolsep}{6pt}
\caption{Performance on polyp diagnosis (prospective validation on Hospital H8). Metrics include AUROC, macro-ACC, specificity at 90\% sensitivity, and positive-class NPV/PPV with 95\% CIs.}
\label{tab:Polyp_Diagnosis_Prospective_H8}
\begin{tabular}{lccccc}
\toprule
Model & AUROC & Macro-ACC & Spec at 90\% Sens & NPV & PPV \\
\midrule
\textbf{GRACE} & \textbf{0.9511 (0.8803-1.0000)} & 0.8919 (0.7757-0.9750) & \textbf{0.8907 (0.7492-0.9967)} & 0.9367 (0.8421-1.0000) & \textbf{0.8124 (0.5882-1.0000)} \\
Virchow2 & 0.9411 (0.8613-0.9958) & 0.8445 (0.7169-0.9531) & 0.8475 (0.6321-0.9732) & 0.9050 (0.7931-1.0000) & 0.7494 (0.5293-0.9412) \\
UNI & 0.9425 (0.8625-1.0000) & \textbf{0.8960 (0.8226-0.9595)} & 0.8803 (0.6989-0.9967) & \textbf{1.0000 (1.0000-1.0000)} & 0.6859 (0.5000-0.8824) \\
CONCH & 0.9136 (0.7869-0.9958) & 0.8331 (0.7222-0.9219) & 0.7890 (0.2341-0.9732) & 0.9607 (0.8696-1.0000) & 0.6043 (0.4091-0.8097) \\
\bottomrule
\end{tabular}
\vspace{-4pt}
\end{table}

\begin{table}[t]
\centering
\footnotesize
\setlength{\tabcolsep}{6pt}
\caption{Performance on ulcer diagnosis (Hospital H1). Metrics include AUROC, macro-ACC, specificity at 90\% sensitivity, and positive-class NPV/PPV with 95\% CIs.}
\label{tab:Ulcer_Diagnosis}
\begin{tabular}{lccccc}
\toprule
Model & AUROC & Macro-ACC & Spec at 90\% Sens & NPV & PPV \\
\midrule
\textbf{GRACE} & \textbf{0.8828 (0.7876-0.9525)} & 0.9768 (0.9677-0.9852) & \textbf{0.7088 (0.3177-0.9064)} & 0.9766 (0.9667-0.9852) & \textbf{1.0000 (1.0000-1.0000)} \\
Virchow2 & 0.8820 (0.7972-0.9491) & 0.9760 (0.9668-0.9843) & 0.6952 (0.4080-0.8663) & 0.9776 (0.9685-0.9861) & 0.6643 (0.2500-1.0000) \\
UNI & 0.8470 (0.7615-0.9173) & 0.9595 (0.9474-0.9705) & 0.5689 (0.3845-0.7659) & \textbf{0.9821 (0.9734-0.9896)} & 0.2640 (0.1176-0.4244) \\
CONCH & 0.8674 (0.7798-0.9475) & \textbf{0.9769 (0.9677-0.9843)} & 0.6625 (0.2207-0.8930) & 0.9778 (0.9686-0.9861) & 0.7955 (0.3333-1.0000) \\
\bottomrule
\end{tabular}
\vspace{-4pt}
\end{table}

\begin{table}[t]
\centering
\footnotesize
\setlength{\tabcolsep}{6pt}
\caption{Performance on ulcer diagnosis (Hospital H5). Metrics include AUROC, macro-ACC, specificity at 90\% sensitivity, and positive-class NPV/PPV with 95\% CIs.}
\label{tab:Ulcer_Diagnosis_External_H5}
\begin{tabular}{lccccc}
\toprule
Model & AUROC & Macro-ACC & Spec at 90\% Sens & NPV & PPV \\
\midrule
\textbf{GRACE} & \textbf{0.8923 (0.8514-0.9263)} & 0.5768 (0.5470-0.6107) & \textbf{0.7273 (0.5752-0.8495)} & 0.7004 (0.6492-0.7493) & 0.9121 (0.7777-1.0000) \\
Virchow2 & 0.8696 (0.8234-0.9083) & 0.6137 (0.5790-0.6521) & 0.6268 (0.5084-0.7559) & 0.7186 (0.6716-0.7656) & \textbf{0.9368 (0.8333-1.0000)} \\
UNI & 0.8868 (0.8479-0.9234) & \textbf{0.7475 (0.6970-0.7958)} & 0.6902 (0.5686-0.8060) & \textbf{0.8094 (0.7610-0.8540)} & 0.7960 (0.7158-0.8818) \\
CONCH & 0.8741 (0.8324-0.9117) & 0.5422 (0.5179-0.5684) & 0.6596 (0.5317-0.7692) & 0.6846 (0.6371-0.7318) & 0.9155 (0.7143-1.0000) \\

\bottomrule
\end{tabular}
\vspace{-4pt}
\end{table}

\begin{table}[t]
\centering
\footnotesize
\setlength{\tabcolsep}{6pt}
\caption{Performance on intestinal metaplasia diagnosis (Hospital H1). Metrics include AUROC, macro-ACC, specificity at 90\% sensitivity, and positive-class NPV/PPV with 95\% CIs.}
\label{tab:Intestinal_Metaplasia_Diagnosis}
\begin{tabular}{lccccc}
\toprule
Model & AUROC & Macro-ACC & Spec at 90\% Sens & NPV & PPV \\
\midrule
\textbf{GRACE} & \textbf{0.9873 (0.9780-0.9947)} & 0.9573 (0.9410-0.9720) & \textbf{0.9770 (0.9599-0.9900)} & \textbf{0.9726 (0.9549-0.9886)} & 0.9394 (0.9117-0.9655) \\
Virchow2 & 0.9841 (0.9719-0.9936) & 0.9569 (0.9395-0.9720) & 0.9763 (0.9599-0.9900) & 0.9573 (0.9358-0.9763) & 0.9569 (0.9333-0.9795) \\
UNI & 0.9822 (0.9705-0.9920) & \textbf{0.9635 (0.9483-0.9779)} & 0.9700 (0.9498-0.9866) & 0.9604 (0.9396-0.9786) & \textbf{0.9672 (0.9451-0.9867)} \\
CONCH & 0.9814 (0.9706-0.9907) & 0.9555 (0.9395-0.9705) & 0.9710 (0.9532-0.9866) & 0.9628 (0.9435-0.9808) & 0.9478 (0.9216-0.9703) \\
\bottomrule
\end{tabular}
\vspace{-4pt}
\end{table}

\begin{table}[t]
\centering
\footnotesize
\setlength{\tabcolsep}{6pt}
\caption{Performance on intestinal metaplasia diagnosis (Hospital H5). Metrics include AUROC, macro-ACC, specificity at 90\% sensitivity, and positive-class NPV/PPV with 95\% CIs.}
\label{tab:Intestinal_Metaplasia_Diagnosis_External_H5}
\begin{tabular}{lccccc}
\toprule
Model & AUROC & Macro-ACC & Spec at 90\% Sens & NPV & PPV \\
\midrule
\textbf{GRACE} & \textbf{0.9423 (0.9225-0.9607)} & 0.8928 (0.8637-0.9127) & 0.8821 (0.8395-0.9264) & 0.9990 (0.9971-1.0000) & 0.1449 (0.1054-0.1905) \\
Virchow2 & 0.9179 (0.8699-0.9514) & \textbf{0.9112 (0.8840-0.9310)} & 0.8763 (0.8427-0.9030) & \textbf{0.9991 (0.9972-1.0000)} & \textbf{0.1736 (0.1278-0.2254)} \\
UNI & 0.9414 (0.9151-0.9631) & 0.8967 (0.8685-0.9171) & \textbf{0.8890 (0.8395-0.9264)} & 0.9990 (0.9971-1.0000) & 0.1489 (0.1076-0.1932) \\
CONCH & 0.9380 (0.9104-0.9601) & 0.8808 (0.8525-0.9015) & 0.8718 (0.8127-0.9197) & 0.9990 (0.9969-1.0000) & 0.1303 (0.0961-0.1667) \\
\bottomrule
\end{tabular}
\vspace{-4pt}
\end{table}

\begin{table}[t]
\centering
\footnotesize
\setlength{\tabcolsep}{6pt}
\caption{Performance on intestinal metaplasia diagnosis (prospective validation on Hospital H1). Metrics include AUROC, macro-ACC, specificity at 90\% sensitivity, and positive-class NPV/PPV with 95\% CIs.}
\label{tab:Intestinal_Metaplasia_Diagnosis_Prospective_H1}
\begin{tabular}{lccccc}
\toprule
Model & AUROC & Macro-ACC & Spec at 90\% Sens & NPV & PPV \\
\midrule
\textbf{GRACE} & \textbf{0.9845 (0.9729-0.9940)} & 0.9519 (0.9311-0.9708) & 0.9690 (0.9431-0.9900) & 0.9706 (0.9504-0.9889) & 0.9265 (0.8932-0.9600) \\
Virchow2 & 0.9782 (0.9618-0.9917) & \textbf{0.9580 (0.9394-0.9749)} & \textbf{0.9701 (0.9431-0.9900)} & 0.9707 (0.9495-0.9892) & \textbf{0.9401 (0.9058-0.9709)} \\
UNI & 0.9824 (0.9696-0.9922) & 0.9373 (0.9144-0.9582) & 0.9620 (0.9331-0.9833) & \textbf{0.9731 (0.9520-0.9893)} & 0.8936 (0.8495-0.9333) \\
CONCH & 0.9706 (0.9524-0.9866) & 0.9392 (0.9186-0.9603) & 0.9576 (0.9298-0.9799) & 0.9669 (0.9457-0.9856) & 0.9045 (0.8670-0.9412) \\
\bottomrule
\end{tabular}
\vspace{-4pt}
\end{table}

\begin{table}[t]
\centering
\footnotesize
\setlength{\tabcolsep}{6pt}
\caption{Performance on intestinal metaplasia diagnosis (prospective validation on Hospital H2). Metrics include AUROC, macro-ACC, specificity at 90\% sensitivity, and positive-class NPV/PPV with 95\% CIs.}
\label{tab:Intestinal_Metaplasia_Diagnosis_Prospective_H2}
\begin{tabular}{lccccc}
\toprule
Model & AUROC & Macro-ACC & Spec at 90\% Sens & NPV & PPV \\
\midrule
\textbf{GRACE} & \textbf{0.9696 (0.9414-0.9916)} & \textbf{0.9627 (0.9383-0.9848)} & \textbf{0.9508 (0.9097-0.9833)} & \textbf{0.9645 (0.9402-0.9849)} & \textbf{0.9914 (0.9722-1.0000)} \\
Virchow2 & 0.9643 (0.9371-0.9869) & 0.9505 (0.9240-0.9742) & 0.9347 (0.8862-0.9732) & 0.9602 (0.9336-0.9839) & 0.9587 (0.9189-0.9915) \\
UNI & 0.9670 (0.9383-0.9906) & 0.9524 (0.9261-0.9751) & 0.9419 (0.8997-0.9766) & 0.9634 (0.9370-0.9842) & 0.9511 (0.9083-0.9847) \\
CONCH & 0.9636 (0.9378-0.9877) & 0.9369 (0.9073-0.9644) & 0.9337 (0.8829-0.9732) & 0.9490 (0.9201-0.9737) & 0.9495 (0.9062-0.9835) \\
\bottomrule
\end{tabular}
\vspace{-4pt}
\end{table}

\begin{table}[t]
\centering
\footnotesize
\setlength{\tabcolsep}{6pt}
\caption{Performance on intestinal metaplasia diagnosis (prospective validation on Hospital H8). Metrics include AUROC, macro-ACC, specificity at 90\% sensitivity, and positive-class NPV/PPV with 95\% CIs.}
\label{tab:Intestinal_Metaplasia_Diagnosis_Prospective_H8}
\begin{tabular}{lccccc}
\toprule
Model & AUROC & Macro-ACC & Spec at 90\% Sens & NPV & PPV \\
\midrule
\textbf{GRACE} & \textbf{0.9411 (0.8166-1.0000)} & \textbf{0.9082 (0.8040-1.0000)} & 0.8821 (0.6351-0.9967) & 0.8955 (0.7273-1.0000) & \textbf{0.9352 (0.7692-1.0000)} \\
Virchow2 & 0.9350 (0.8042-1.0000) & 0.8755 (0.7604-0.9723) & 0.8854 (0.5549-0.9967) & 0.8489 (0.6667-1.0000) & 0.9304 (0.7500-1.0000) \\
UNI & 0.9378 (0.8104-1.0000) & \textbf{0.9082 (0.8040-1.0000)} & \textbf{0.8951 (0.5845-0.9967)} & \textbf{0.8969 (0.7497-1.0000)} & 0.9334 (0.8000-1.0000) \\
CONCH & 0.9259 (0.8194-1.0000) & \textbf{0.9082 (0.8040-1.0000)} & 0.8154 (0.4180-0.9967) & 0.8900 (0.7333-1.0000) & 0.9341 (0.7692-1.0000) \\
\bottomrule
\end{tabular}
\vspace{-4pt}
\end{table}

\begin{table}[t]
\centering
\footnotesize
\setlength{\tabcolsep}{6pt}
\caption{Performance on intestinal metaplasia diagnosis (prospective validation on Hospital H9). Metrics include AUROC, macro-ACC, specificity at 90\% sensitivity, and positive-class NPV/PPV with 95\% CIs.}
\label{tab:Intestinal_Metaplasia_Diagnosis_Prospective_H9}
\begin{tabular}{lccccc}
\toprule
Model & AUROC & Macro-ACC & Spec at 90\% Sens & NPV & PPV \\
\midrule
\textbf{GRACE} & \textbf{0.9668 (0.9320-0.9926)} & 0.9198 (0.8742-0.9611) & \textbf{0.9289 (0.8462-0.9866)} & 0.9738 (0.9365-1.0000) & 0.8197 (0.7192-0.9153) \\
Virchow2 & 0.9587 (0.9207-0.9884) & \textbf{0.9317 (0.8887-0.9695)} & 0.9231 (0.8528-0.9766) & \textbf{0.9745 (0.9382-1.0000)} & \textbf{0.8510 (0.7600-0.9412)} \\
UNI & 0.9518 (0.9014-0.9900) & 0.9017 (0.8514-0.9480) & 0.9128 (0.8093-0.9766) & 0.9712 (0.9275-1.0000) & 0.7807 (0.6818-0.8769) \\
CONCH & 0.9572 (0.9181-0.9883) & 0.9215 (0.8726-0.9651) & 0.9075 (0.7993-0.9732) & 0.9610 (0.9155-1.0000) & 0.8493 (0.7500-0.9423) \\
\bottomrule
\end{tabular}
\vspace{-4pt}
\end{table}

\begin{table}[t]
\centering
\footnotesize
\setlength{\tabcolsep}{6pt}
\caption{Performance on malignancy diagnosis (Hospital H1). Metrics include AUROC, macro-ACC, specificity at 90\% sensitivity, and positive-class NPV/PPV with 95\% CIs.}
\label{tab:Malignancy_Diagnosis}
\begin{tabular}{lccccc}
\toprule
Model & AUROC & Macro-ACC & Spec at 90\% Sens & NPV & PPV \\
\midrule
\textbf{GRACE} & \textbf{0.9979 (0.9947-1.0000)} & \textbf{0.9973 (0.9937-1.0000)} & \textbf{0.9946 (0.9799-0.9967)} & \textbf{0.9973 (0.9936-1.0000)} & \textbf{1.0000 (1.0000-1.0000)} \\
Virchow2 & 0.9976 (0.9937-1.0000) & 0.9955 (0.9910-0.9991) & 0.9940 (0.9833-0.9967) & 0.9955 (0.9908-0.9991) & \textbf{1.0000 (1.0000-1.0000)} \\
UNI & 0.9823 (0.9406-1.0000) & 0.9927 (0.9873-0.9973) & 0.9511 (0.8494-0.9967) & 0.9925 (0.9871-0.9972) & \textbf{1.0000 (1.0000-1.0000)} \\
CONCH & 0.9632 (0.8808-1.0000) & 0.9955 (0.9910-0.9991) & 0.9525 (0.8495-0.9967) & 0.9972 (0.9935-1.0000) & 0.9085 (0.7825-1.0000) \\
\bottomrule
\end{tabular}
\vspace{-4pt}
\end{table}

\begin{table}[t]
\centering
\footnotesize
\setlength{\tabcolsep}{6pt}
\caption{Performance on malignancy diagnosis (prospective validation on Hospital H7). Metrics include AUROC, macro-ACC, specificity at 90\% sensitivity, and positive-class NPV/PPV with 95\% CIs.}
\label{tab:Malignancy_Diagnosis_Prospective_H7}
\begin{tabular}{lccccc}
\toprule
Model & AUROC & Macro-ACC & Spec at 90\% Sens & NPV & PPV \\
\midrule
\textbf{GRACE} & \textbf{0.9799 (0.9631-0.9927)} & \textbf{0.9192 (0.8865-0.9500)} & 0.9511 (0.9064-0.9833) & \textbf{0.9516 (0.9308-0.9704)} & \textbf{0.9912 (0.9716-1.0000)} \\
Virchow2 & 0.9786 (0.9601-0.9918) & 0.8694 (0.8321-0.9070) & \textbf{0.9530 (0.8997-0.9833)} & 0.9239 (0.9000-0.9471) & 0.9900 (0.9677-1.0000) \\
UNI & 0.9748 (0.9560-0.9884) & 0.8486 (0.8108-0.8877) & 0.9458 (0.8963-0.9833) & 0.9142 (0.8876-0.9386) & 0.9788 (0.9451-1.0000) \\
CONCH & 0.9784 (0.9619-0.9910) & 0.9143 (0.8821-0.9427) & 0.9475 (0.9030-0.9833) & 0.9511 (0.9313-0.9697) & 0.9563 (0.9160-0.9908) \\
\bottomrule
\end{tabular}
\vspace{-4pt}
\end{table}

\begin{table}[t]
\centering
\footnotesize
\setlength{\tabcolsep}{6pt}
\caption{Performance on malignancy diagnosis (prospective validation on Hospital H8). Metrics include AUROC, macro-ACC, specificity at 90\% sensitivity, and positive-class NPV/PPV with 95\% CIs.}
\label{tab:Malignancy_Diagnosis_Prospective_H8}
\begin{tabular}{lccccc}
\toprule
Model & AUROC & Macro-ACC & Spec at 90\% Sens & NPV & PPV \\
\midrule
\textbf{GRACE} & \textbf{0.9975 (0.9908-1.0000)} & 0.9366 (0.8845-0.9804) & \textbf{0.9930 (0.9498-0.9967)} & 0.9178 (0.8428-0.9818) & 0.9777 (0.9302-1.0000) \\
Virchow2 & 0.9919 (0.9712-1.0000) & 0.9257 (0.8734-0.9756) & 0.9802 (0.9398-0.9967) & 0.9024 (0.8167-0.9683) & 0.9772 (0.9286-1.0000) \\
UNI & 0.9971 (0.9894-1.0000) & 0.9471 (0.8992-0.9886) & 0.9897 (0.9431-0.9967) & 0.9312 (0.8571-0.9844) & 0.9771 (0.9231-1.0000) \\
CONCH & 0.9947 (0.9826-1.0000) & \textbf{0.9580 (0.9158-1.0000)} & 0.9810 (0.9398-0.9967) & \textbf{0.9478 (0.8823-1.0000)} & \textbf{0.9780 (0.9268-1.0000)} \\
\bottomrule
\end{tabular}
\vspace{-4pt}
\end{table}

\begin{table}[t]
\centering
\footnotesize
\setlength{\tabcolsep}{6pt}
\caption{Performance on pathology subtyping (Hospital H2). Metrics include macro-AUC, macro-ACC, specificity at 90\% sensitivity, and positive-class NPV/PPV with 95\% CIs.}
\label{tab:Pathology_Subtyping}
\begin{tabular}{lccccc}
\toprule
Model & Macro-AUC & Macro-ACC & Spec at 90\% Sens & NPV & PPV \\
\midrule
\textbf{GRACE} & \textbf{0.9019 (0.8609-0.9351)} & 0.7051 (0.6331-0.7770) & 0.7190 (0.6187-0.8294) & 0.9229 (0.8759-0.9622) & 0.6394 (0.3000-1.0000) \\
Virchow2 & 0.9006 (0.8582-0.9379) & 0.7493 (0.6763-0.8201) & 0.7292 (0.5987-0.8361) & \textbf{0.9293 (0.8819-0.9688)} & 0.6691 (0.3846-0.9167) \\
UNI & 0.8965 (0.8541-0.9337) & 0.7407 (0.6691-0.8059) & 0.7197 (0.6020-0.8328) & 0.8908 (0.8432-0.9412) & \textbf{1.0000 (1.0000-1.0000)} \\
CONCH & 0.8999 (0.8642-0.9368) & \textbf{0.7556 (0.6835-0.8201)} & \textbf{0.7417 (0.6387-0.8495)} & 0.9233 (0.8730-0.9685) & 0.8802 (0.6000-1.0000) \\
\bottomrule
\end{tabular}
\vspace{2pt}

\begin{minipage}{0.97\linewidth}
\raggedright
\footnotesize
\textit{For pathology subtype detection, NPV and PPV were calculated with signet-ring cell carcinoma treated as the positive class and non-specified gastric adenocarcinoma and tubular adenocarcinoma grouped as the negative class.}
\end{minipage}
\vspace{-4pt}
\end{table}

\begin{table}[t]
\centering
\footnotesize
\setlength{\tabcolsep}{6pt}
\caption{Performance on pathology subtyping (prospective validation on Hospital H1). Metrics include macro-AUC, macro-ACC, specificity at 90\% sensitivity, and positive-class NPV/PPV with 95\% CIs.}
\label{tab:Pathology_Subtyping_Prospective_H1}
\begin{tabular}{lccccc}
\toprule
Model & Macro-AUC & Macro-ACC & Spec at 90\% Sens & NPV & PPV \\
\midrule
\textbf{GRACE} & \textbf{0.8937 (0.7955-0.9821)} & \textbf{0.6797 (0.4704-0.8333)} & 0.6245 (0.3545-0.9431) & 0.6778 (0.4815-0.8462) & \textbf{1.0000 (1.0000-1.0000)} \\
Virchow2 & 0.8552 (0.6944-0.9882) & 0.6207 (0.4174-0.8206) & 0.5094 (0.1605-0.9967) & \textbf{0.6923 (0.5000-0.8621)} & 0.8589 (0.5714-1.0000) \\
UNI & 0.8909 (0.7726-0.9838) & 0.6343 (0.4111-0.7982) & 0.6202 (0.3110-0.9967) & 0.6807 (0.5000-0.8401) & \textbf{1.0000 (1.0000-1.0000)} \\
CONCH & 0.8840 (0.7881-0.9575) & 0.6715 (0.5457-0.7821) & \textbf{0.7033 (0.4983-0.9264)} & 0.6755 (0.5000-0.8462) & \textbf{1.0000 (1.0000-1.0000)} \\
\bottomrule
\end{tabular}
\vspace{-4pt}
\end{table}

\begin{table}[t]
\centering
\footnotesize
\setlength{\tabcolsep}{6pt}
\caption{Performance on pathology subtyping (prospective validation on Hospital H7). Metrics include macro-AUC, macro-ACC, specificity at 90\% sensitivity, and positive-class NPV/PPV with 95\% CIs.}
\label{tab:Pathology_Subtyping_Prospective_H7}
\begin{tabular}{lccccc}
\toprule
Model & Macro-AUC & Macro-ACC & Spec at 90\% Sens & NPV & PPV \\
\midrule
\textbf{GRACE} & \textbf{0.8862 (0.6051-0.9437)} & \textbf{0.6959 (0.4664-0.8517)} & \textbf{0.7623 (0.6084-0.8863)} & \textbf{0.9940 (0.9811-1.0000)} & 0.1338 (0.0000-0.3333) \\
Virchow2 & 0.8377 (0.5679-0.8959) & 0.4984 (0.4436-0.5480) & 0.6143 (0.4816-0.7726) & 0.9825 (0.9595-1.0000) & - \\
UNI & 0.8686 (0.5916-0.9274) & 0.6214 (0.4808-0.8615) & 0.7279 (0.5719-0.8629) & 0.9883 (0.9706-1.0000) & \textbf{0.1675 (0.0000-0.5000)} \\
CONCH & 0.8137 (0.5272-0.8815) & 0.4785 (0.4185-0.5336) & 0.5225 (0.3344-0.7325) & 0.9823 (0.9638-1.0000) & - \\
\bottomrule
\end{tabular}
\vspace{-4pt}
\end{table}

\begin{table}[t]
\centering
\footnotesize
\setlength{\tabcolsep}{6pt}
\caption{Performance on Lauren classification (Hospital H1, H3, H4). Metrics include AUROC, macro-ACC, specificity at 90\% sensitivity, and positive-class NPV/PPV with 95\% CIs.}
\label{tab:Lauren_Classification}
\begin{tabular}{lccccc}
\toprule
Model & AUROC & Macro-ACC & Spec at 90\% Sens & NPV & PPV \\
\midrule
\textbf{GRACE} & \textbf{0.9566 (0.9143-0.9882)} & 0.8691 (0.8017-0.9174) & 0.8946 (0.7793-0.9766) & 0.9698 (0.9194-1.0000) & 0.7556 (0.6429-0.8546) \\
Virchow2 & 0.9440 (0.8922-0.9843) & 0.8767 (0.8099-0.9339) & 0.8874 (0.7659-0.9732) & \textbf{0.9702 (0.9206-1.0000)} & 0.7673 (0.6595-0.8710) \\
UNI & 0.9414 (0.8914-0.9826) & 0.8679 (0.8017-0.9256) & 0.8529 (0.7090-0.9565) & 0.8776 (0.7999-0.9487) & 0.8554 (0.7333-0.9524) \\
CONCH & 0.9400 (0.8798-0.9863) & \textbf{0.8935 (0.8428-0.9421)} & \textbf{0.9008 (0.7659-0.9732)} & 0.8986 (0.8243-0.9620) & \textbf{0.8841 (0.7805-0.9750)} \\
\bottomrule
\end{tabular}
\vspace{2pt}

\begin{minipage}{0.97\linewidth}
\raggedright
\footnotesize
\textit{For Lauren classification, NPV and PPV were calculated with intestinal-type treated as the positive class and diffuse-type treated as the negative class.}
\end{minipage}
\vspace{-4pt}
\end{table}

\FloatBarrier
\begin{table}[t]
\centering
\footnotesize
\setlength{\tabcolsep}{6pt}
\caption{Performance on Lauren classification (prospective validation on Hospital H7). Metrics include AUROC, macro-ACC, specificity at 90\% sensitivity, and positive-class NPV/PPV with 95\% CIs.}
\label{tab:Lauren_Classification_Prospective_H7}
\begin{tabular}{lccccc}
\toprule
Model & AUROC & Macro-ACC & Spec at 90\% Sens & NPV & PPV \\
\midrule
\textbf{GRACE} & \textbf{0.9727 (0.9143-1.0000)} & \textbf{0.9724 (0.9302-1.0000)} & \textbf{0.9666 (0.8997-0.9967)} & \textbf{0.9731 (0.9091-1.0000)} & \textbf{0.9737 (0.9142-1.0000)} \\
Virchow2 & 0.9712 (0.9091-1.0000) & 0.9307 (0.8624-0.9861) & 0.9606 (0.8796-0.9967) & 0.9204 (0.8286-1.0000) & 0.9434 (0.8571-1.0000) \\
UNI & 0.9674 (0.9087-1.0000) & 0.8789 (0.8045-0.9456) & 0.9552 (0.8930-0.9967) & 0.8135 (0.6818-0.9268) & 0.9672 (0.8889-1.0000) \\
CONCH & 0.9697 (0.9136-1.0000) & 0.8917 (0.8197-0.9545) & 0.9560 (0.8896-0.9967) & 0.8339 (0.7143-0.9413) & 0.9694 (0.8918-1.0000) \\
\bottomrule
\end{tabular}
\vspace{-4pt}
\end{table}

\begin{table}[t]
\centering
\footnotesize
\setlength{\tabcolsep}{6pt}
\caption{Performance on histologic differentiation grading (Hospital H2). Metrics include AUROC, macro-ACC, specificity at 90\% sensitivity, and positive-class NPV/PPV with 95\% CIs.}
\label{tab:Tumor_Differentiation_Grading}
\begin{tabular}{lccccc}
\toprule
Model & AUROC & Macro-ACC & Spec at 90\% Sens & NPV & PPV \\
\midrule
\textbf{GRACE} & \textbf{0.9585 (0.9263-0.9831)} & \textbf{0.9015 (0.8584-0.9358)} & \textbf{0.8828 (0.7792-0.9599)} & 0.6040 (0.4643-0.7347) & \textbf{1.0000 (1.0000-1.0000)} \\
Virchow2 & 0.9524 (0.9171-0.9804) & 0.8314 (0.7475-0.9092) & 0.8684 (0.7793-0.9365) & 0.7920 (0.6286-0.9356) & 0.9169 (0.8648-0.9630) \\
UNI & 0.9219 (0.8701-0.9645) & 0.6041 (0.5390-0.6786) & 0.7442 (0.6053-0.9030) & \textbf{0.8757 (0.5827-1.0000)} & 0.8071 (0.7401-0.8731) \\
CONCH & 0.9305 (0.8829-0.9704) & 0.7527 (0.6665-0.8390) & 0.7938 (0.6220-0.9164) & 0.8480 (0.6667-1.0000) & 0.8741 (0.8073-0.9333) \\
\bottomrule
\end{tabular}
\vspace{-4pt}
\end{table}

\begin{table}[t]
\centering
\footnotesize
\setlength{\tabcolsep}{6pt}
\caption{Performance on histologic differentiation grading (Hospital H3). Metrics include AUROC, macro-ACC, specificity at 90\% sensitivity, and positive-class NPV/PPV with 95\% CIs.}
\label{tab:Tumor_Differentiation_Grading_External_H3}
\begin{tabular}{lccccc}
\toprule
Model & AUROC & Macro-ACC & Spec at 90\% Sens & NPV & PPV \\
\midrule
\textbf{GRACE} & \textbf{0.9320 (0.8779-0.9750)} & \textbf{0.9201 (0.8709-0.9600)} & \textbf{0.9030 (0.7458-0.9533)} & 0.7735 (0.6667-0.8731) & \textbf{0.9747 (0.9471-0.9940)} \\
Virchow2 & 0.9268 (0.8860-0.9634) & 0.7775 (0.7081-0.8500) & 0.8870 (0.7525-0.9398) & 0.7035 (0.5652-0.8333) & 0.8978 (0.8546-0.9390) \\
UNI & 0.8999 (0.8439-0.9462) & 0.6231 (0.5646-0.6898) & 0.7894 (0.5149-0.9030) & \textbf{0.8139 (0.6000-1.0000)} & 0.8209 (0.7688-0.8718) \\
CONCH & 0.9097 (0.8535-0.9556) & 0.7933 (0.7219-0.8604) & 0.8274 (0.6254-0.9130) & 0.7315 (0.5945-0.8537) & 0.9016 (0.8523-0.9464) \\
\bottomrule
\end{tabular}
\vspace{-4pt}
\end{table}

\begin{table}[t]
\centering
\footnotesize
\setlength{\tabcolsep}{6pt}
\caption{Performance on histologic differentiation grading (Hospital H4). Metrics include AUROC, macro-ACC, specificity at 90\% sensitivity, and positive-class NPV/PPV with 95\% CIs.}
\label{tab:Tumor_Differentiation_Grading_External_H4}
\begin{tabular}{lccccc}
\toprule
Model & AUROC & Macro-ACC & Spec at 90\% Sens & NPV & PPV \\
\midrule
\textbf{GRACE} & \textbf{0.9255 (0.8717-0.9661)} & \textbf{0.8713 (0.8070-0.9257)} & 0.7945 (0.5986-0.9130) & \textbf{0.8912 (0.8000-0.9636)} & \textbf{0.8534 (0.7656-0.9333)} \\
Virchow2 & 0.9169 (0.8578-0.9634) & 0.7455 (0.6706-0.8167) & \textbf{0.8025 (0.6287-0.9231)} & \textbf{0.8912 (0.7750-0.9750)} & 0.6788 (0.5696-0.7821) \\
UNI & 0.8899 (0.8226-0.9420) & 0.5598 (0.5137-0.6102) & 0.6946 (0.5183-0.8863) & 0.8885 (0.6361-1.0000) & 0.5322 (0.4392-0.6204) \\
CONCH & 0.9115 (0.8542-0.9575) & 0.7613 (0.6862-0.8348) & 0.7443 (0.5685-0.9097) & 0.8612 (0.7500-0.9524) & 0.7079 (0.6053-0.8031) \\
\bottomrule
\end{tabular}
\vspace{-4pt}
\end{table}

\begin{table}[t]
\centering
\footnotesize
\setlength{\tabcolsep}{6pt}
\caption{Performance on histologic differentiation grading (prospective validation on Hospital H2). Metrics include AUROC, macro-ACC, specificity at 90\% sensitivity, and positive-class NPV/PPV with 95\% CIs.}
\label{tab:Tumor_Differentiation_Grading_Prospective_H2}
\begin{tabular}{lccccc}
\toprule
Model & AUROC & Macro-ACC & Spec at 90\% Sens & NPV & PPV \\
\midrule
\textbf{GRACE} & \textbf{0.9485 (0.9007-0.9818)} & \textbf{0.8289 (0.7470-0.8965)} & \textbf{0.8162 (0.6755-0.9365)} & 0.8854 (0.7667-0.9737) & \textbf{0.8097 (0.7059-0.9048)} \\
Virchow2 & 0.9321 (0.8711-0.9761) & 0.7776 (0.6925-0.8517) & 0.8150 (0.6722-0.9431) & 0.9219 (0.8075-1.0000) & 0.7402 (0.6296-0.8413) \\
UNI & 0.9135 (0.8442-0.9675) & 0.5246 (0.5000-0.5612) & 0.7664 (0.5752-0.9197) & \textbf{1.0000 (1.0000-1.0000)} & 0.5550 (0.4505-0.6594) \\
CONCH & 0.9270 (0.8671-0.9724) & 0.7289 (0.6449-0.8121) & 0.8064 (0.6086-0.9431) & 0.9548 (0.8500-1.0000) & 0.6883 (0.5735-0.7971) \\
\bottomrule
\end{tabular}
\vspace{-4pt}
\end{table}

\begin{table}[t]
\centering
\footnotesize
\setlength{\tabcolsep}{6pt}
\caption{Performance on histologic differentiation grading (prospective validation on Hospital H7). Metrics include AUROC, macro-ACC, specificity at 90\% sensitivity, and positive-class NPV/PPV with 95\% CIs.}
\label{tab:Tumor_Differentiation_Grading_Prospective_H7}
\begin{tabular}{lccccc}
\toprule
Model & AUROC & Macro-ACC & Spec at 90\% Sens & NPV & PPV \\
\midrule
\textbf{GRACE} & \textbf{0.9458 (0.8914-0.9830)} & 0.8515 (0.7947-0.9049) & 0.8518 (0.7258-0.9498) & 0.6480 (0.5373-0.7500) & \textbf{0.9621 (0.9157-1.0000)} \\
Virchow2 & 0.9364 (0.8790-0.9761) & \textbf{0.8783 (0.8120-0.9335)} & \textbf{0.8556 (0.7258-0.9499)} & 0.7804 (0.6610-0.8889) & 0.9373 (0.8830-0.9806) \\
UNI & 0.9306 (0.8740-0.9723) & 0.6957 (0.6220-0.7703) & 0.8171 (0.6788-0.9197) & \textbf{0.9454 (0.8234-1.0000)} & 0.7849 (0.7087-0.8522) \\
CONCH & 0.9081 (0.8515-0.9566) & 0.7921 (0.7148-0.8623) & 0.7435 (0.5686-0.9030) & 0.8735 (0.7428-0.9714) & 0.8523 (0.7863-0.9123) \\
\bottomrule
\end{tabular}
\vspace{-4pt}
\end{table}

\begin{table}[t]
\centering
\footnotesize
\setlength{\tabcolsep}{6pt}
\caption{Performance on histologic differentiation grading (prospective validation on Hospital H8). Metrics include AUROC, macro-ACC, specificity at 90\% sensitivity, and positive-class NPV/PPV with 95\% CIs.}
\label{tab:Tumor_Differentiation_Grading_Prospective_H8}
\begin{tabular}{lccccc}
\toprule
Model & AUROC & Macro-ACC & Spec at 90\% Sens & NPV & PPV \\
\midrule
\textbf{GRACE} & \textbf{0.9319 (0.8423-0.9917)} & \textbf{0.8098 (0.7297-0.8871)} & \textbf{0.8770 (0.4983-0.9967)} & \textbf{1.0000 (1.0000-1.0000)} & \textbf{0.2990 (0.1111-0.5263)} \\
Virchow2 & 0.8732 (0.7477-0.9667) & 0.7688 (0.6892-0.8462) & 0.6936 (0.2462-0.9498) & \textbf{1.0000 (1.0000-1.0000)} & 0.2608 (0.0951-0.4545) \\
UNI & 0.9172 (0.8134-0.9910) & 0.5251 (0.5000-0.5694) & 0.7892 (0.3962-0.9732) & \textbf{1.0000 (1.0000-1.0000)} & 0.1467 (0.0488-0.2683) \\
CONCH & 0.8799 (0.7210-0.9821) & 0.7699 (0.6842-0.8530) & 0.7118 (0.3311-0.9699) & \textbf{1.0000 (1.0000-1.0000)} & 0.2636 (0.1053-0.4546) \\
\bottomrule
\end{tabular}
\vspace{-4pt}
\end{table}

\begin{table}[t]
\centering
\footnotesize
\setlength{\tabcolsep}{6pt}
\caption{Performance on T-staging (Hospital H2). Metrics include macro-AUC, macro-ACC, specificity at 90\% sensitivity, and positive-class NPV/PPV with 95\% CIs.}
\label{tab:T-Staging_Internal_H2}
\begin{tabular}{lccccc}
\toprule
Model & Macro-AUC & Macro-ACC & Spec at 90\% Sens & NPV & PPV \\
\midrule
\textbf{GRACE} & \textbf{0.8927 (0.8471-0.9341)} & 0.5873 (0.5160-0.6533) & \textbf{0.7111 (0.5853-0.8161)} & 0.9752 (0.9111-1.0000) & 0.7840 (0.6825-0.8834) \\
Virchow2 & 0.8880 (0.8408-0.9307) & 0.5587 (0.4859-0.6283) & 0.7037 (0.5484-0.8261) & \textbf{0.9769 (0.9184-1.0000)} & 0.8224 (0.7231-0.9139) \\
UNI & 0.8626 (0.8109-0.9101) & 0.5784 (0.4928-0.6745) & 0.6490 (0.4247-0.7760) & 0.9766 (0.9211-1.0000) & 0.8098 (0.7118-0.9105) \\
CONCH & 0.8917 (0.8484-0.9298) & \textbf{0.6285 (0.5438-0.7138)} & 0.6734 (0.5350-0.8094) & 0.9599 (0.8980-1.0000) & \textbf{0.8737 (0.7818-0.9608)} \\
\bottomrule
\end{tabular}
\vspace{2pt}

\begin{minipage}{0.97\linewidth}
\raggedright
\footnotesize
\textit{For NPV and PPV calculation, stages T3 and T4 were treated as the positive class, and stages T1 and T2 were grouped as the negative class.}
\end{minipage}
\vspace{-4pt}
\end{table}

\begin{table}[t]
\centering
\footnotesize
\setlength{\tabcolsep}{6pt}
\caption{Performance on T-staging (Hospital H6). Metrics include macro-AUC, macro-ACC, specificity at 90\% sensitivity, and positive-class NPV/PPV with 95\% CIs.}
\label{tab:T-Staging_External_H6}
\begin{tabular}{lccccc}
\toprule
Model & Macro-AUC & Macro-ACC & Spec at 90\% Sens & NPV & PPV \\
\midrule
\textbf{GRACE} & \textbf{0.8800 (0.8420-0.9170)} & \textbf{0.6695 (0.6337-0.6990)} & \textbf{0.6662 (0.5149-0.7692)} & \textbf{0.9770 (0.9250-1.0000)} & 0.8449 (0.7817-0.9014) \\
Virchow2 & 0.8706 (0.8344-0.9038) & 0.5067 (0.4605-0.5473) & 0.5663 (0.3745-0.7593) & 0.8774 (0.7884-0.9516) & 0.8914 (0.8358-0.9412) \\
UNI & 0.8672 (0.8349-0.8990) & 0.5972 (0.5263-0.6749) & 0.6311 (0.5585-0.7057) & 0.8649 (0.7719-0.9465) & 0.9062 (0.8512-0.9538) \\
CONCH & 0.7984 (0.7505-0.8428) & 0.5753 (0.5143-0.6365) & 0.4339 (0.3010-0.5585) & 0.8272 (0.7333-0.9130) & \textbf{0.9484 (0.9065-0.9835)} \\
\bottomrule
\end{tabular}
\vspace{-4pt}
\end{table}

\begin{table}[t]
\centering
\footnotesize
\setlength{\tabcolsep}{6pt}
\caption{Performance on T-staging (prospective validation on Hospital H2). Metrics include macro-AUC, macro-ACC, specificity at 90\% sensitivity, and positive-class NPV/PPV with 95\% CIs.}
\label{tab:T-Staging_Prospective_H2}
\begin{tabular}{lccccc}
\toprule
Model & Macro-AUC & Macro-ACC & Spec at 90\% Sens & NPV & PPV \\
\midrule
\textbf{GRACE} & \textbf{0.8838 (0.8376-0.9264)} & \textbf{0.5665 (0.4890-0.6393)} & \textbf{0.6994 (0.5485-0.8227)} & \textbf{0.9753 (0.9167-1.0000)} & 0.7425 (0.6271-0.8491) \\
Virchow2 & 0.8829 (0.8363-0.9249) & 0.5429 (0.4688-0.6059) & 0.6438 (0.5083-0.7894) & 0.9739 (0.9142-1.0000) & 0.7149 (0.5937-0.8276) \\
UNI & 0.8555 (0.8101-0.8996) & 0.5011 (0.4244-0.5767) & 0.5718 (0.4448-0.7191) & 0.9720 (0.9000-1.0000) & 0.6811 (0.5614-0.8001) \\
CONCH & 0.8330 (0.7742-0.8903) & 0.5575 (0.4846-0.6347) & 0.5256 (0.3410-0.6890) & 0.9573 (0.8919-1.0000) & \textbf{0.7922 (0.6667-0.9039)} \\
\bottomrule
\end{tabular}
\vspace{-4pt}
\end{table}

\begin{table}[t]
\centering
\footnotesize
\setlength{\tabcolsep}{6pt}
\caption{Performance on N-staging prediction (Hospital H2). Metrics include AUROC, macro-ACC, specificity at 90\% sensitivity, and positive-class NPV/PPV with 95\% CIs.}
\label{tab:Lymph_Node_Metastasis_Prediction}
\begin{tabular}{lccccc}
\toprule
Model & AUROC & Macro-ACC & Spec at 90\% Sens & NPV & PPV \\
\midrule
\textbf{GRACE} & \textbf{0.8381 (0.7569-0.9050)} & \textbf{0.8075 (0.7357-0.8714)} & \textbf{0.5852 (0.2641-0.8562)} & \textbf{0.8312 (0.7419-0.9178)} & \textbf{0.7868 (0.6857-0.8769)} \\
Virchow2 & 0.8236 (0.7435-0.8990) & 0.7811 (0.7071-0.8502) & 0.5197 (0.2107-0.7926) & 0.7874 (0.6986-0.8767) & 0.7744 (0.6713-0.8719) \\
UNI & 0.7990 (0.7259-0.8724) & 0.7127 (0.6427-0.7857) & 0.4604 (0.2875-0.6589) & 0.7210 (0.6154-0.8219) & 0.7085 (0.5968-0.8182) \\
CONCH & 0.8225 (0.7501-0.8907) & 0.7650 (0.6929-0.8357) & 0.5185 (0.3043-0.7593) & 0.7497 (0.6486-0.8391) & 0.7844 (0.6724-0.8889) \\
\bottomrule
\end{tabular}
\vspace{-4pt}
\end{table}

\begin{table}[t]
\centering
\footnotesize
\setlength{\tabcolsep}{6pt}
\caption{Performance on N-staging prediction (Hospital H1). Metrics include AUROC, macro-ACC, specificity at 90\% sensitivity, and positive-class NPV/PPV with 95\% CIs.}
\label{tab:Lymph_Node_Metastasis_Prediction_External_H1}
\begin{tabular}{lccccc}
\toprule
Model & AUROC & Macro-ACC & Spec at 90\% Sens & NPV & PPV \\
\midrule
\textbf{GRACE} & \textbf{0.8009 (0.7538-0.8431)} & 0.7179 (0.6709-0.7638) & \textbf{0.4796 (0.3645-0.5819)} & 0.6878 (0.6237-0.7486) & 0.7497 (0.6885-0.8083) \\
Virchow2 & 0.7859 (0.7414-0.8298) & 0.7314 (0.6910-0.7714) & 0.4320 (0.3578-0.5385) & \textbf{0.7929 (0.7253-0.8550)} & 0.6996 (0.6423-0.7528) \\
UNI & 0.7245 (0.6746-0.7738) & 0.6066 (0.5628-0.6558) & 0.3291 (0.2341-0.4314) & 0.7720 (0.6530-0.8814) & 0.5798 (0.5287-0.6332) \\
CONCH & 0.7889 (0.7452-0.8312) & \textbf{0.7381 (0.6960-0.7815)} & 0.4483 (0.3077-0.5519) & 0.7161 (0.6480-0.7795) & \textbf{0.7603 (0.6996-0.8182)} \\
\bottomrule
\end{tabular}
\vspace{-4pt}
\end{table}

\begin{table}[t]
\centering
\footnotesize
\setlength{\tabcolsep}{6pt}
\caption{Performance on N-staging prediction (prospective validation on Hospital H1). Metrics include AUROC, macro-ACC, specificity at 90\% sensitivity, and positive-class NPV/PPV with 95\% CIs.}
\label{tab:Lymph_Node_Metastasis_Prediction_Prospective_H1}
\begin{tabular}{lccccc}
\toprule
Model & AUROC & Macro-ACC & Spec at 90\% Sens & NPV & PPV \\
\midrule
\textbf{GRACE} & \textbf{0.7876 (0.6507-0.8991)} & \textbf{0.7379 (0.6147-0.8562)} & \textbf{0.4563 (0.1572-0.7391)} & 0.8671 (0.7200-1.0000) & \textbf{0.5868 (0.4137-0.7667)} \\
Virchow2 & 0.7857 (0.6441-0.9048) & 0.6826 (0.5431-0.7988) & 0.4249 (0.1170-0.8362) & 0.8163 (0.6428-0.9545) & 0.5341 (0.3571-0.7188) \\
UNI & 0.7021 (0.5364-0.8332) & 0.6482 (0.5568-0.7420) & 0.3223 (0.0970-0.5921) & \textbf{0.9162 (0.7143-1.0000)} & 0.4782 (0.3256-0.6411) \\
CONCH & 0.7127 (0.5650-0.8464) & 0.6795 (0.5558-0.8046) & 0.3385 (0.0501-0.6187) & 0.8185 (0.6400-0.9644) & 0.5340 (0.3546-0.7084) \\
\bottomrule
\end{tabular}
\vspace{-4pt}
\end{table}

\begin{table}[t]
\centering
\footnotesize
\setlength{\tabcolsep}{6pt}
\caption{Performance on perineural invasion (Hospital H1). Metrics include AUROC, macro-ACC, specificity at 90\% sensitivity, and positive-class NPV/PPV with 95\% CIs.}
\label{tab:Perineural_Invasion_}
\begin{tabular}{lccccc}
\toprule
Model & AUROC & Macro-ACC & Spec at 90\% Sens & NPV & PPV \\
\midrule
\textbf{GRACE} & \textbf{0.9381 (0.8749-0.9857)} & 0.8373 (0.7625-0.9125) & 0.8226 (0.5920-0.9599) & 0.7199 (0.5556-0.8611) & 0.9339 (0.8537-1.0000) \\
Virchow2 & 0.9216 (0.8472-0.9773) & 0.8638 (0.7875-0.9375) & 0.7849 (0.5652-0.9332) & 0.7610 (0.6110-0.8860) & 0.9366 (0.8571-1.0000) \\
UNI & 0.9363 (0.8637-0.9906) & \textbf{0.9128 (0.8500-0.9625)} & \textbf{0.8708 (0.5582-0.9799)} & \textbf{0.9190 (0.7998-1.0000)} & 0.9077 (0.8276-0.9804) \\
CONCH & 0.9254 (0.8509-0.9800) & 0.8008 (0.7125-0.8875) & 0.7927 (0.5885-0.9365) & 0.6452 (0.5116-0.7879) & \textbf{0.9484 (0.8718-1.0000)} \\
\bottomrule
\end{tabular}
\vspace{-4pt}
\end{table}

\begin{table}[t]
\centering
\footnotesize
\setlength{\tabcolsep}{6pt}
\caption{Performance on perineural invasion (Hospital H6). Metrics include AUROC, macro-ACC, specificity at 90\% sensitivity, and positive-class NPV/PPV with 95\% CIs.}
\label{tab:Perineural_Invasion_External_H6}
\begin{tabular}{lccccc}
\toprule
Model & AUROC & Macro-ACC & Spec at 90\% Sens & NPV & PPV \\
\midrule
\textbf{GRACE} & \textbf{0.9141 (0.8702-0.9518)} & 0.8311 (0.7737-0.8842) & 0.7452 (0.6421-0.8496) & 0.8272 (0.7302-0.9131) & 0.8876 (0.8385-0.9305) \\
Virchow2 & 0.8894 (0.8401-0.9353) & 0.8204 (0.7613-0.8753) & 0.6986 (0.5552-0.8295) & 0.7605 (0.6618-0.8507) & 0.8894 (0.8395-0.9342) \\
UNI & 0.8634 (0.8039-0.9150) & 0.7740 (0.7100-0.8319) & 0.6294 (0.5050-0.7425) & \textbf{0.8278 (0.7166-0.9231)} & 0.8442 (0.7907-0.8927) \\
CONCH & 0.9136 (0.8713-0.9530) & \textbf{0.8359 (0.7811-0.8881)} & \textbf{0.7516 (0.6320-0.8696)} & 0.7312 (0.6373-0.8261) & \textbf{0.9097 (0.8621-0.9533)} \\
\bottomrule
\end{tabular}
\vspace{-4pt}
\end{table}

\begin{table}[t]
\centering
\footnotesize
\setlength{\tabcolsep}{6pt}
\caption{Performance on perineural invasion (prospective validation on Hospital H1). Metrics include AUROC, macro-ACC, specificity at 90\% sensitivity, and positive-class NPV/PPV with 95\% CIs.}
\label{tab:Perineural_Invasion_Prospective_H1}
\begin{tabular}{lccccc}
\toprule
Model & AUROC & Macro-ACC & Spec at 90\% Sens & NPV & PPV \\
\midrule
\textbf{GRACE} & \textbf{0.8905 (0.7990-0.9585)} & 0.7475 (0.6371-0.8542) & 0.6623 (0.4514-0.9030) & 0.8019 (0.6249-0.9524) & 0.7142 (0.5588-0.8684) \\
Virchow2 & 0.8685 (0.7486-0.9630) & \textbf{0.8146 (0.7129-0.9057)} & \textbf{0.6892 (0.3208-0.9298)} & 0.8538 (0.6957-0.9677) & \textbf{0.7861 (0.6471-0.9231)} \\
UNI & 0.8618 (0.7596-0.9509) & 0.7305 (0.6333-0.8261) & 0.6221 (0.3578-0.8897) & \textbf{0.9374 (0.7997-1.0000)} & 0.6555 (0.5122-0.7917) \\
CONCH & 0.8390 (0.7302-0.9310) & 0.7634 (0.6532-0.8667) & 0.5564 (0.2742-0.8495) & 0.8673 (0.7143-1.0000) & 0.7109 (0.5749-0.8485) \\
\bottomrule
\end{tabular}
\vspace{-4pt}
\end{table}

\begin{table}[t]
\centering
\footnotesize
\setlength{\tabcolsep}{6pt}
\caption{Performance on perineural invasion (prospective validation on Hospital H8). Metrics include AUROC, macro-ACC, specificity at 90\% sensitivity, and positive-class NPV/PPV with 95\% CIs.}
\label{tab:Perineural_Invasion_Prospective_H8}
\begin{tabular}{lccccc}
\toprule
Model & AUROC & Macro-ACC & Spec at 90\% Sens & NPV & PPV \\
\midrule
\textbf{GRACE} & \textbf{0.9169 (0.8161-0.9893)} & \textbf{0.8459 (0.7051-0.9643)} & \textbf{0.7519 (0.4783-0.9599)} & 0.8361 (0.5830-1.0000) & 0.8865 (0.7500-1.0000) \\
Virchow2 & 0.9076 (0.7913-0.9904) & 0.8061 (0.6607-0.9355) & 0.7317 (0.3445-0.9599) & 0.7200 (0.4614-0.9333) & 0.8774 (0.7391-1.0000) \\
UNI & 0.9099 (0.7874-0.9884) & 0.8443 (0.7000-0.9615) & 0.7292 (0.3678-0.9575) & \textbf{1.0000 (1.0000-1.0000)} & 0.8621 (0.7308-0.9667) \\
CONCH & 0.9101 (0.7949-0.9888) & 0.8226 (0.6827-0.9460) & 0.7246 (0.4180-0.9566) & 0.6890 (0.4667-0.9048) & \textbf{0.9089 (0.7778-1.0000)} \\
\bottomrule
\end{tabular}
\vspace{-4pt}
\end{table}

\begin{table}[t]
\centering
\footnotesize
\setlength{\tabcolsep}{6pt}
\caption{Performance on vascular invasion (Hospital H1, H3, H4). Metrics include AUROC, macro-ACC, specificity at 90\% sensitivity, and positive-class NPV/PPV with 95\% CIs.}
\label{tab:Vascular_Invasion}
\begin{tabular}{lccccc}
\toprule
Model & AUROC & Macro-ACC & Spec at 90\% Sens & NPV & PPV \\
\midrule
\textbf{GRACE} & \textbf{0.8371 (0.7828-0.8896)} & \textbf{0.7418 (0.6789-0.8000)} & \textbf{0.5542 (0.4247-0.6756)} & \textbf{0.8274 (0.7407-0.9091)} & 0.6856 (0.6000-0.7658) \\
Virchow2 & 0.8289 (0.7692-0.8835) & 0.7200 (0.6526-0.7842) & 0.5402 (0.4047-0.6790) & 0.7347 (0.6392-0.8182) & \textbf{0.7095 (0.6185-0.7979)} \\
UNI & 0.8036 (0.7361-0.8636) & 0.7365 (0.6736-0.8000) & 0.4968 (0.3512-0.6421) & 0.8172 (0.7200-0.8974) & 0.6837 (0.5966-0.7607) \\
CONCH & 0.7926 (0.7264-0.8547) & 0.7373 (0.6737-0.7947) & 0.4364 (0.2809-0.5819) & 0.8146 (0.7222-0.8986) & 0.6848 (0.6017-0.7642) \\
\bottomrule
\end{tabular}
\vspace{-4pt}
\end{table}

\begin{table}[t]
\centering
\footnotesize
\setlength{\tabcolsep}{6pt}
\caption{Performance on tumor regression grade (Hospital H2). Metrics include AUROC, macro-ACC, specificity at 90\% sensitivity, and positive-class NPV/PPV with 95\% CIs.}
\label{tab:Response_to_Chemotherapy_Diagnosis}
\begin{tabular}{lccccc}
\toprule
Model & AUROC & Macro-ACC & Spec at 90\% Sens & NPV & PPV \\
\midrule
\textbf{GRACE} & \textbf{0.9013 (0.7762-0.9871)} & \textbf{0.8144 (0.6757-0.9189)} & \textbf{0.7289 (0.3746-0.9632)} & \textbf{0.6671 (0.0000-1.0000)} & 0.8250 (0.6970-0.9412) \\
Virchow2 & 0.8760 (0.7327-0.9834) & 0.8108 (0.6757-0.9189) & 0.6706 (0.3303-0.9632) & 0.6664 (0.0000-1.0000) & 0.8250 (0.7027-0.9412) \\
UNI & 0.8769 (0.6989-0.9921) & 0.8094 (0.6757-0.9189) & 0.6663 (0.2976-0.9666) & 0.5948 (0.0000-1.0000) & \textbf{0.8464 (0.7097-0.9667)} \\
CONCH & 0.8723 (0.7412-0.9784) & 0.7575 (0.6216-0.8919) & 0.6847 (0.1973-0.9632) & - & 0.7771 (0.6389-0.8919) \\
\bottomrule
\end{tabular}
\vspace{2pt}
\raggedright
\footnotesize

\textit{For NPV and PPV calculation, high residual tumor was treated as the positive class and low residual tumor was treated as the negative class.}

\vspace{-4pt}
\end{table}

\begin{table}[t]
\centering
\footnotesize
\setlength{\tabcolsep}{6pt}
\caption{Performance on HER2 status prediction (Hospital H2). Metrics include AUROC, macro-ACC, specificity at 90\% sensitivity, and positive-class NPV/PPV with 95\% CIs.}
\label{tab:HER2_Expression_Prediction}
\begin{tabular}{lccccc}
\toprule
Model & AUROC & Macro-ACC & Spec at 90\% Sens & NPV & PPV \\
\midrule
\textbf{GRACE} & \textbf{0.8811 (0.5764-1.0000)} & 0.8535 (0.4808-1.0000) & \textbf{0.6995 (0.0000-0.9967)} & 0.9785 (0.9362-1.0000) & 0.5981 (0.0000-1.0000) \\
Virchow2 & 0.8454 (0.4113-1.0000) & 0.6161 (0.4700-0.9898) & 0.6503 (0.0000-0.9967) & 0.9408 (0.8654-1.0000) & 0.4912 (0.0000-1.0000) \\
UNI & 0.8084 (0.5385-0.9667) & 0.6140 (0.4706-0.9808) & 0.5892 (0.0000-0.9574) & 0.9399 (0.8722-1.0000) & 0.5090 (0.0000-1.0000) \\
CONCH & 0.8512 (0.3846-1.0000) & \textbf{0.8817 (0.5000-1.0000)} & 0.6811 (0.0000-0.9967) & \textbf{0.9812 (0.9388-1.0000)} & \textbf{1.0000 (1.0000-1.0000)} \\
\bottomrule
\end{tabular}
\vspace{-4pt}
\end{table}

\begin{table}[t]
\centering
\footnotesize
\setlength{\tabcolsep}{6pt}
\caption{Performance on MMR status prediction (Hospital H2). Metrics include AUROC, macro-ACC, specificity at 90\% sensitivity, and positive-class NPV/PPV with 95\% CIs.}
\label{tab:MMR_Prediction}
\begin{tabular}{lccccc}
\toprule
Model & AUROC & Macro-ACC & Spec at 90\% Sens & NPV & PPV \\
\midrule
\textbf{GRACE} & \textbf{0.9236 (0.8503-0.9779)} & 0.9134 (0.8519-0.9753) & \textbf{0.7861 (0.5351-0.9264)} & 0.9092 (0.8421-0.9620) & \textbf{1.0000 (1.0000-1.0000)} \\
Virchow2 & 0.9140 (0.8185-0.9815) & \textbf{0.9259 (0.8642-0.9753)} & 0.7253 (0.4983-0.9365) & \textbf{0.9213 (0.8553-0.9737)} & \textbf{1.0000 (1.0000-1.0000)} \\
UNI & 0.8725 (0.7569-0.9605) & 0.9014 (0.8395-0.9630) & 0.6237 (0.3539-0.8863) & 0.9070 (0.8354-0.9627) & 0.7964 (0.3333-1.0000) \\
CONCH & 0.8360 (0.6662-0.9753) & 0.9142 (0.8519-0.9753) & 0.5562 (0.2709-0.9064) & 0.9198 (0.8571-0.9737) & 0.8397 (0.5000-1.0000) \\
\bottomrule
\end{tabular}
\vspace{2pt}

\begin{minipage}{0.97\linewidth}
\raggedright
\footnotesize
\textit{NPV and PPV were calculated using dMMR as the positive class and pMMR as the negative class.}
\end{minipage}
\vspace{-4pt}
\end{table}

\begin{table}[t]
\centering
\footnotesize
\setlength{\tabcolsep}{6pt}
\caption{Performance on MMR status prediction (Hospital H1). Metrics include AUROC, macro-ACC, specificity at 90\% sensitivity, and positive-class NPV/PPV with 95\% CIs.}
\label{tab:MMR_Prediction_External_H1}
\begin{tabular}{lccccc}
\toprule
Model & AUROC & Macro-ACC & Spec at 90\% Sens & NPV & PPV \\
\midrule
\textbf{GRACE} & \textbf{0.9252 (0.8421-0.9877)} & \textbf{0.9321 (0.8824-0.9748)} & \textbf{0.7905 (0.4916-0.9699)} & 0.9520 (0.9072-0.9901) & 0.8207 (0.6250-1.0000) \\
Virchow2 & 0.9244 (0.8564-0.9768) & 0.8820 (0.8235-0.9412) & 0.7605 (0.4983-0.9298) & \textbf{0.9581 (0.9175-0.9899)} & 0.6013 (0.4118-0.8000) \\
UNI & 0.9162 (0.8286-0.9864) & 0.9154 (0.8655-0.9580) & 0.7362 (0.4883-0.9532) & 0.9500 (0.9020-0.9896) & 0.7331 (0.5263-0.9231) \\
CONCH & 0.8980 (0.7910-0.9798) & 0.9240 (0.8739-0.9664) & 0.7216 (0.3144-0.9532) & 0.9333 (0.8811-0.9806) & \textbf{0.8579 (0.6364-1.0000)} \\
\bottomrule
\end{tabular}
\vspace{-4pt}
\end{table}

\begin{table}[t]
\centering
\footnotesize
\setlength{\tabcolsep}{6pt}
\caption{Performance on MMR status prediction (prospective validation on Hospital H2). Metrics include AUROC, macro-ACC, specificity at 90\% sensitivity, and positive-class NPV/PPV with 95\% CIs.}
\label{tab:MMR_Prediction_Prospective_H2}
\begin{tabular}{lccccc}
\toprule
Model & AUROC & Macro-ACC & Spec at 90\% Sens & NPV & PPV \\
\midrule
\textbf{GRACE} & \textbf{0.9265 (0.8529-0.9838)} & \textbf{0.8321 (0.6650-0.9718)} & \textbf{0.8265 (0.5549-0.9565)} & \textbf{0.9693 (0.9242-1.0000)} & \textbf{0.5028 (0.2000-0.8000)} \\
Virchow2 & 0.9121 (0.8310-0.9744) & 0.7614 (0.5783-0.9481) & 0.8126 (0.4983-0.9465) & 0.9542 (0.9000-1.0000) & 0.4187 (0.1250-0.7143) \\
UNI & 0.8704 (0.7439-0.9730) & 0.7543 (0.5717-0.9392) & 0.6527 (0.3597-0.9097) & 0.9542 (0.8971-1.0000) & 0.3892 (0.1247-0.6667) \\
CONCH & 0.8900 (0.7916-0.9701) & 0.7549 (0.5732-0.9416) & 0.7193 (0.3980-0.9197) & 0.9552 (0.8970-1.0000) & 0.3920 (0.1000-0.6923) \\
\bottomrule
\end{tabular}
\vspace{-4pt}
\end{table}

\begin{table}[t]
\centering
\footnotesize
\setlength{\tabcolsep}{6pt}
\caption{Performance on Ki-67 expression prediction (Hospital H1, H3, H4). Metrics include macro-AUC, macro-ACC, specificity at 90\% sensitivity, and positive-class NPV/PPV with 95\% CIs.}
\label{tab:Ki-67_Prediction}
\begin{tabular}{lccccc}
\toprule
Model & Macro-AUC & Macro-ACC & Spec at 90\% Sens & NPV & PPV \\
\midrule
\textbf{GRACE} & \textbf{0.7467 (0.6767-0.8128)} & 0.5845 (0.5070-0.6690) & \textbf{0.4024 (0.1437-0.5786)} & 0.7291 (0.6182-0.8333) & \textbf{0.6975 (0.5977-0.7901)} \\
Virchow2 & 0.7256 (0.6548-0.7938) & 0.5215 (0.4437-0.6056) & 0.3065 (0.1405-0.4515) &  - & 0.5207 (0.4437-0.5986) \\
UNI & 0.7363 (0.6615-0.8066) & \textbf{0.6120 (0.5350-0.6972)} & 0.3125 (0.1773-0.4783) & \textbf{0.8004 (0.6756-0.9250)} & 0.6459 (0.5464-0.7429) \\
CONCH & 0.7355 (0.6610-0.8008) & 0.6118 (0.5282-0.6901) & 0.3624 (0.1872-0.5251) & 0.7616 (0.6250-0.8800) & 0.6550 (0.5618-0.7423) \\
\bottomrule
\end{tabular}
\vspace{2pt}
\raggedright
\footnotesize

\textit{NPV and PPV were calculated using high Ki-67 expression as the positive class and low- and medium-expression cases grouped as the negative class.}

\vspace{-4pt}
\end{table}

\begin{table}[t]
\centering
\footnotesize
\setlength{\tabcolsep}{6pt}
\caption{Performance on EBV status prediction (Hospital H2). Metrics include AUROC, macro-ACC, specificity at 90\% sensitivity, and positive-class NPV/PPV with 95\% CIs.}
\label{tab:EBV_Status_Prediction}
\begin{tabular}{lccccc}
\toprule
Model & AUROC & Macro-ACC & Spec at 90\% Sens & NPV & PPV \\
\midrule
\textbf{GRACE} & \textbf{0.8061 (0.6157-1.0000)} & 0.9526 (0.8889-1.0000) & 0.4109 (0.0000-0.9967) & 0.9518 (0.8889-1.0000) &  - \\
Virchow2 & 0.7167 (0.4764-1.0000) & 0.9380 (0.8730-1.0000) & 0.3210 (0.0000-0.9967) & 0.9658 (0.9138-1.0000) & 0.3283 (0.0000-1.0000) \\
UNI & 0.7640 (0.4028-1.0000) & 0.9672 (0.9206-1.0000) & \textbf{0.5443 (0.0000-0.9967)} & 0.9678 (0.9180-1.0000) & \textbf{1.0000 (1.0000-1.0000)} \\
CONCH & 0.6647 (0.2064-1.0000) & \textbf{0.9676 (0.9206-1.0000)} & 0.3365 (0.0000-0.9967) & \textbf{0.9679 (0.9180-1.0000)} & \textbf{1.0000 (1.0000-1.0000)} \\
\bottomrule
\end{tabular}
\vspace{-4pt}
\end{table}

\begin{table}[t]
\centering
\footnotesize
\setlength{\tabcolsep}{6pt}
\caption{Performance on disease-free survival (DFS) and overall survival (OS) prediction  (Hospital H3, H4). The C-index and the 95\% CIs are reported.}
\label{tab:Survival_DFS_OS_prediction}
\begin{tabular}{cccc}
\toprule
Model &  DFS & OS \\
\midrule
\textbf{GRACE}  & 0.7329 (0.5829-0.8434) & \textbf{0.6869 (0.5623-0.7868)}\\
Virchow2  & \textbf{0.7403 (0.5712-0.8482)} & 0.6856 (0.5747-0.7763) \\
CONCH & 0.7253 (0.5632-0.8364)  & 0.6534 (0.5451-0.7506)\\
UNI  & 0.7304 (0.5928-0.8376) & 0.6633 (0.5618-0.7568) \\
\bottomrule
\end{tabular}
\end{table}

\begin{table}[t]
\centering
\footnotesize
\setlength{\tabcolsep}{6pt}
\caption{Performance on disease-free survival (DFS) and overall survival (OS) prediction (Hospital H1). The C-index and the 95\% CIs are reported.}
\label{tab:Survival_DFS_OS_prediction_external}
\begin{tabular}{ccc}
\toprule
Model & DFS & OS \\
\midrule
\textbf{GRACE} 
& \textbf{0.7032 (0.6451-0.7570)} 
& \textbf{0.7144 (0.6525-0.7782)} \\

Virchow2  
& 0.6643 (0.5954-0.7264) 
& 0.7076 (0.6336-0.7813) \\

CONCH 
& 0.6935 (0.6347-0.7524)  
& 0.6794 (0.5968-0.7579) \\

UNI  
& 0.6498 (0.5851-0.7053) 
& 0.6761 (0.5948-0.7612) \\

\bottomrule
\end{tabular}
\end{table}

\begin{table}[!htbp]
\setlength{\tabcolsep}{5pt}
\renewcommand{\arraystretch}{1.15}
\caption{\textbf{Comparison of continued-pretraining strategies based on low-rank adaptation (LoRA) and full fine-tuning (FFT) on the gastric dataset for malignancy diagnosis.} "Trainable params" reports the number of parameters that are optimized. "Tuned/Total" reports the fraction of total backbone parameters that were trainable. The best result is shown in bold.}
\label{tab:ablation_malignancy_diagnosis}
\begin{tabular}{lccccc}
\toprule
\textbf{Method} & \textbf{Trainable params} & \textbf{Tuned/Total} & \textbf{AUROC} & \textbf{Macro-ACC}  & \textbf{Spec at 90\% Sens} \\
\midrule
\multicolumn{6}{l}{\textbf{Internal cohorts H1, N = 5,519 cases (5,765 slides)}} \\
\midrule
FFT & 655,426,560 & 100\% & 0.9951 (0.9891-0.9994) & 0.9343 (0.8481-0.9982) & 0.9895 (0.9565-0.9967) \\
LoRA (Ours) & 26,153,216 & 3.978\% & \textbf{0.9979 (0.9947-1.0000)}  & \textbf{0.9973 (0.9937-1.0000)} & \textbf{0.9946 (0.9799-0.9967)} \\
\midrule
\multicolumn{6}{l}{\textbf{External validation cohort H8, N = 102 cases (127 slides)}} \\
\midrule
FFT & 655,426,560 & 100\% & 0.9939 (0.9794-1.0000) & 0.9045 (0.8477-0.9583) & 0.9805 (0.9398-0.9967) \\
LoRA (Ours) & 26,153,216 & 3.978\%  & \textbf{0.9975 (0.9908-1.0000)} &  \textbf{0.9366 (0.8845-0.9804)} & \textbf{0.9930 (0.9498-0.9967)} \\
\bottomrule
\end{tabular}
\vspace{2pt}

\footnotesize
Abbreviations: LoRA, low-rank adaptation; FFT, full fine-tuning; params, parameters; Spec at 90\% Sens, specificity at 90\% sensitivity.
\end{table}

\begin{table}[!htbp]
\setlength{\tabcolsep}{5pt}
\renewcommand{\arraystretch}{1.15}
\caption{\textbf{Comparison of continued-pretraining strategies based on low-rank adaptation (LoRA) and full fine-tuning (FFT) on the gastric dataset for chronic gastritis severity grading.} "Trainable params" reports the number of parameters that are optimized. "Tuned/Total" reports the fraction of total backbone parameters that were trainable. The best result is shown in bold.}
\label{tab:ablation_diagnosis_of_grading_chronic_gastritis}
\begin{tabular}{lccccc}
\toprule
\textbf{Method} & \textbf{Trainable params} & \textbf{Tuned/Total} & \textbf{Macro-AUC} & \textbf{Macro-ACC} & \textbf{Spec at 90\% Sens} \\
\midrule
\multicolumn{6}{l}{\textbf{Internal cohort H1, N = 3,632 cases (3,779 slides)}} \\
\midrule
FFT & 655,426,560 & 100\% & 0.9204 (0.9045-0.9352) & 0.7201 (0.6707-0.7741)  & \textbf{0.8056 (0.7559-0.8428)} \\
LoRA (Ours) & 26,153,216 & 3.978\% & \textbf{0.9218 (0.9059-0.9368)} & \textbf{0.8030 (0.7753-0.8301)}  & 0.8021 (0.7458-0.8528) \\
\midrule
\multicolumn{6}{l}{\textbf{Prospective validation cohort H2, N = 347 cases (348 slides)}} \\
\midrule
FFT & 655,426,560 & 100\%  & 0.9040 (0.6224-0.9717) & 0.5829 (0.5495-0.6123)& 0.8876 (0.7926-0.9431) \\
LoRA (Ours) & 26,153,216 & 3.978\%  & \textbf{0.9194 (0.6326-0.9829)} &\textbf{0.7227 (0.5522-0.9296)}& \textbf{0.9275 (0.8629-0.9699)} \\
\bottomrule
\end{tabular}
\vspace{2pt}

\footnotesize
Abbreviations: LoRA, low-rank adaptation; FFT, full fine-tuning; params, parameters; Spec at 90\% Sens, specificity at 90\% sensitivity.
\end{table}

\begin{table}[!htbp]
\setlength{\tabcolsep}{5pt}
\renewcommand{\arraystretch}{1.15}
\caption{\textbf{Comparison of continued-pretraining strategies based on low-rank adaptation (LoRA) and full fine-tuning (FFT) on the gastric dataset for atrophy severity grading.} "Trainable params" reports the number of parameters that are optimized. "Tuned/Total" reports the fraction of total backbone parameters that were trainable. The best result is shown in bold.}
\label{tab:ablation_atrophy_severity_grading}
\begin{tabular}{lccccc}
\toprule
\textbf{Method} & \textbf{Trainable params} & \textbf{Tuned/Total}   & \textbf{AUROC} &\textbf{Macro-ACC}  & \textbf{Spec at 90\% Sens} \\
\midrule
\multicolumn{6}{l}{\textbf{Internal cohort H1, N = 685 cases (720 slides)}} \\
\midrule
FFT & 655,426,560 & 100\%  & 0.8759 (0.8091-0.9312)  & 0.8009 (0.7162-0.8751) & 0.6962 (0.4649-0.8696) \\
LoRA (Ours) & 26,153,216 & 3.978\%   & \textbf{0.8844 (0.8257-0.9392)} & \textbf{0.8108 (0.7391-0.8696)} & \textbf{0.7029 (0.4983-0.8930)} \\
\bottomrule
\end{tabular}
\vspace{2pt}

\footnotesize
Abbreviations: LoRA, low-rank adaptation; FFT, full fine-tuning; params, parameters; Spec at 90\% Sens, specificity at 90\% sensitivity.
\end{table}

\begin{table}[!htbp]
\centering
\renewcommand{\arraystretch}{1.2}
\begin{threeparttable}
\captionsetup{justification=raggedright,singlelinecheck=false,width=\linewidth}
\caption{\textbf{Unified regression analysis of crossover effects.}}
\label{tab:crossover_regression}

\begin{tabular}{p{8.8cm} c c}
\toprule
\textbf{Covariate} & \textbf{OR (95\% CI)} & \textbf{$P$ value} \\
\midrule

\multicolumn{3}{l}{\textbf{Main effects}} \\

Crossover sequence (Group B vs A) 
& 1.34 (0.77-2.30) & 0.298 \\

\addlinespace
\multicolumn{3}{l}{\textbf{Interaction terms}} \\

Group B (vs A)$\times$ Atrophy diagnosis in chronic gastritis
& 0.97 (0.69-1.36) & 0.848 \\

Group B (vs A) $\times$ Intestinal metaplasia diagnosis 
& 0.80 (0.28-2.26) & 0.674 \\

Group B (vs A) $\times$ Senior reader 
& 1.60 (0.76-3.38) & 0.216 \\

\bottomrule
\end{tabular}

\begin{tablenotes}
\footnotesize
\item Abbreviations: OR, odds ratio; CI, confidence interval.
\item Group A was specified as the reference crossover sequence. 
All odds ratios are interpreted relative to the stated reference categories. Odds ratios were estimated using GEEs with a logit link and exchangeable correlation structure, clustered by pathologist. Models adjusted for diagnostic task, experience level, crossover sequence, and their interaction terms.
\end{tablenotes}
\end{threeparttable}
\end{table}

\begin{table}[!htbp]
\centering
\caption{\textbf{Distribution of diagnostic observations classified incorrectly in the unassisted condition and correctly in the AI-assisted condition (raw counts, \(N\)).} In the AI-assisted condition, the final diagnoses for these same observations matched the ground truth.}
\label{tab:task_true_label_predicted}
\begin{threeparttable}
\setlength{\tabcolsep}{6pt}
\begin{tabular}{l>{\centering\arraybackslash}m{2.8cm}
                >{\centering\arraybackslash}m{1.3cm}
                >{\centering\arraybackslash}m{1.3cm}
                >{\centering\arraybackslash}m{1.3cm}}
\toprule
\textbf{Task} 
& \textbf{True label} 
& \multicolumn{3}{c}{\textbf{Unassisted diagnosis}} \\
\midrule
\multirow{4}{*}{AA grading}
& \textit{Label}     
& \textbf{Mild} & \textbf{Moderate} & \textbf{Severe} \\
\cmidrule(lr){3-5}
& Mild    
& \cellcolor{cyan!10}0  & \cellcolor{cyan!45}49 & \cellcolor{cyan!15}2  \\
& Moderate
& \cellcolor{cyan!25}15 & \cellcolor{cyan!10}0  & \cellcolor{cyan!28}17 \\
& Severe 
& \cellcolor{cyan!10}0  & \cellcolor{cyan!18}4  & \cellcolor{cyan!10}0  \\
\midrule

\multirow{3}{*}{Atrophy diagnosis}
& \textit{Label}     
& \textbf{Absent} & \textbf{Present} & \textbf{N/A} \\
\cmidrule(lr){3-5}
& Absent  
& \cellcolor{cyan!10}0  & \cellcolor{cyan!35}35 & -- \\
& Present 
& \cellcolor{cyan!45}49 & \cellcolor{cyan!10}0  & -- \\
\midrule

\multirow{3}{*}{IM diagnosis}
& \textit{Label}     
& \textbf{Absent} & \textbf{Present} & \textbf{N/A} \\
\cmidrule(lr){3-5}
& Absent  
& \cellcolor{cyan!10}0  & \cellcolor{cyan!40}44 & -- \\
& Present 
& \cellcolor{cyan!28}16 & \cellcolor{cyan!10}0  & -- \\
\bottomrule
\end{tabular}

\begin{tablenotes}
\footnotesize
\item Abbreviations: AI, artificial intelligence; AA, acute activity; N/A, not applicable; IM, intestinal metaplasia.
\end{tablenotes}

\end{threeparttable}
\end{table}

\FloatBarrier
\begin{figure}[!htbp]
    \centering
    \includegraphics[width=0.75\textwidth]{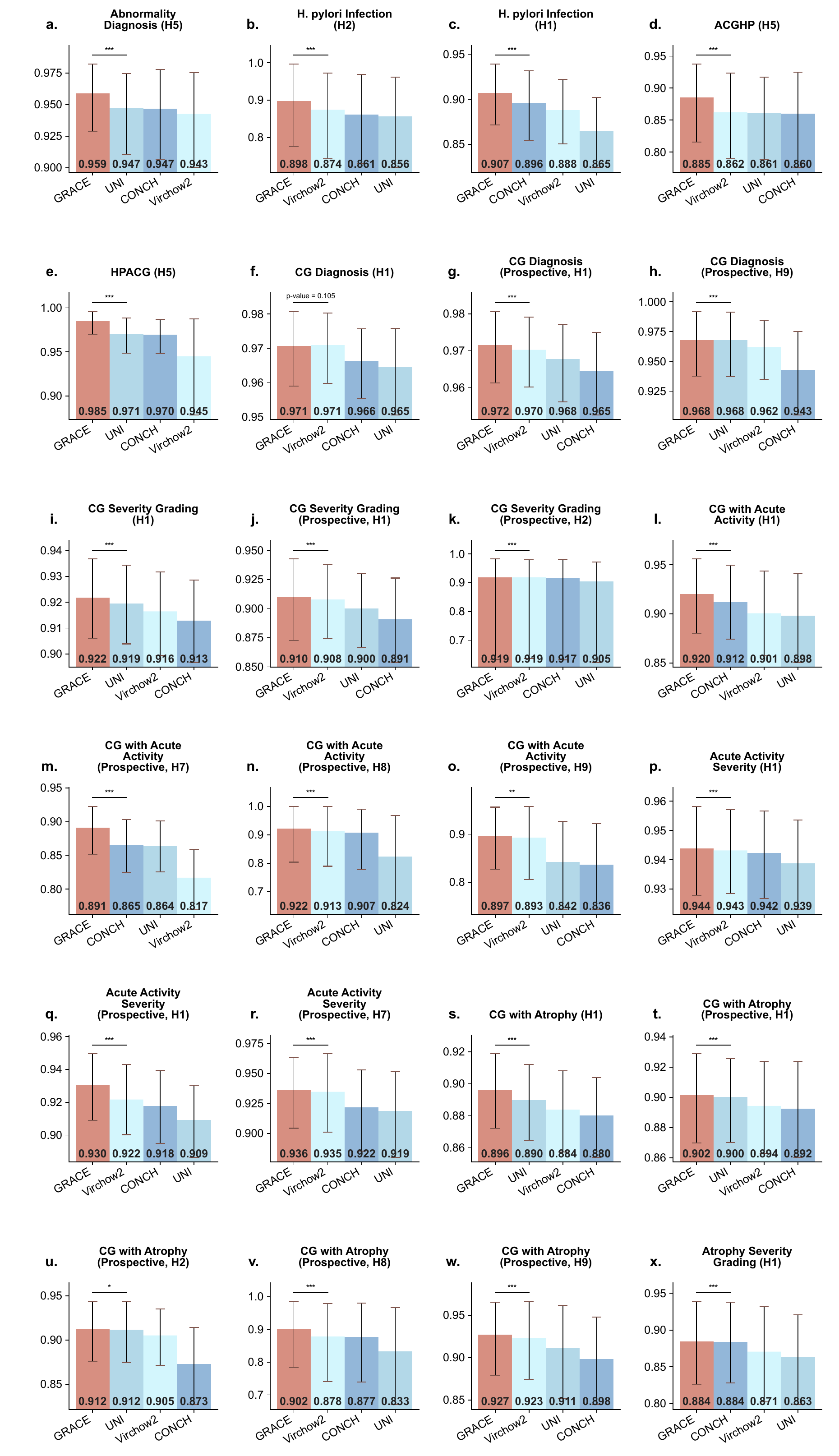}
\caption{\textbf{AUROC and macro-AUC performance comparison between GRACE and representative PFMs with 95\% CIs.}
    (\textbf{a-x}) Statistical significance is annotated using the following convention: "***" for $P < 0.001$, "**" for $P < 0.01$, and "*" for $P < 0.05$.}
    \label{fig:bar-1}
\end{figure}

\FloatBarrier

\begin{figure}[!htbp]
    \centering
    \includegraphics[width=0.75\textwidth]{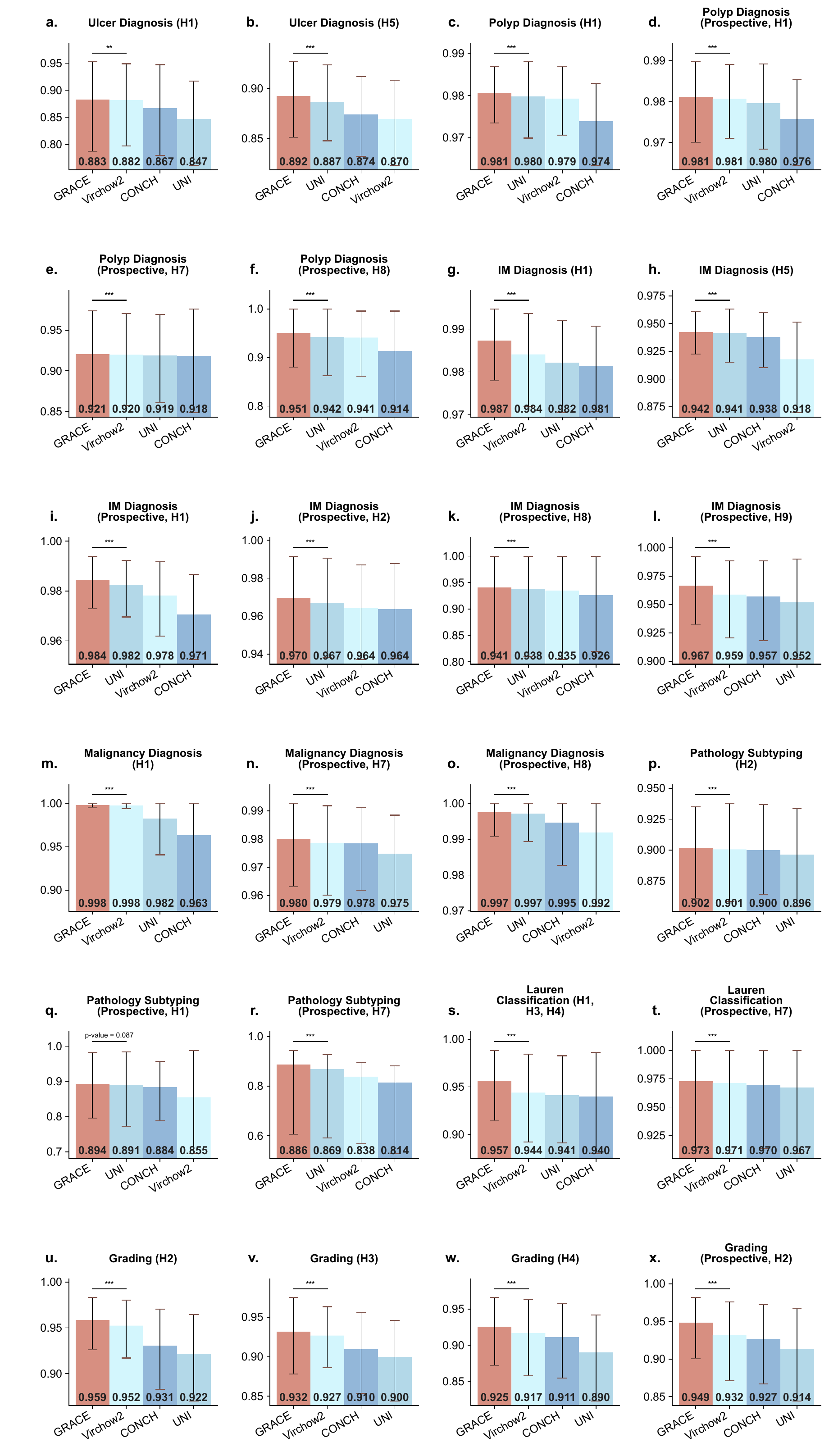}
\caption{\textbf{AUROC and macro-AUC performance comparison between GRACE and representative PFMs with 95\% CIs (continued).}
    (\textbf{a-x}) Statistical significance is annotated using the following convention: "***" for $P < 0.001$, "**" for $P < 0.01$, and "*" for $P < 0.05$.}
    \label{fig:bar-2}
\end{figure}
\FloatBarrier

\begin{figure}[!htbp]
    \centering
    \includegraphics[width=0.75\textwidth]{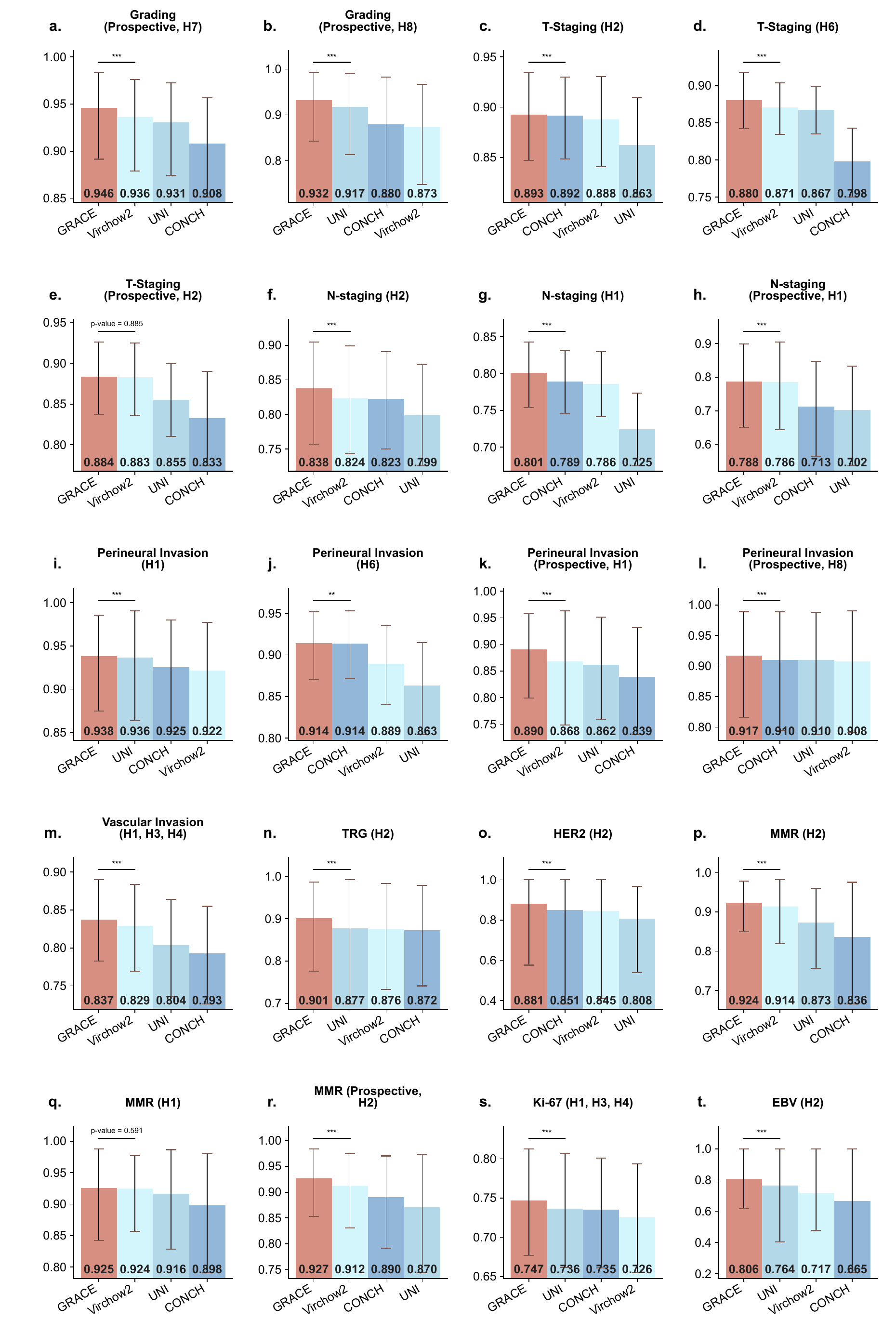}
\caption{\textbf{AUROC and macro-AUC performance comparison between GRACE and representative PFMs with 95\% CIs (continued).}
    (\textbf{a-t}) Statistical significance is annotated using the following convention: "***" for $P < 0.001$, "**" for $P < 0.01$, and "*" for $P < 0.05$.}
    \label{fig:bar-3}
\end{figure}

\FloatBarrier
\begin{table}[!htbp]
\centering
\caption{\textbf{Friedman omnibus comparison of model macro-AUC across paired tasks.}}
\label{tab:friedman_results}
\begin{tabular}{lcccccc}
\toprule
\textbf{Analysis set} & \textbf{Models} & \textbf{Paired tasks, $n$} & \textbf{$\chi^{2}$} & \textbf{df} & \textbf{\textit{P} value} & \textbf{Kendall's $W$} \\
\midrule
All tasks & 4 & 72 & 137.28 & 3 & $ < 0.001$ & 0.64 \\
Internal & 4 & 28 & 51.13 & 3 & $ < 0.001$ & 0.61 \\
External & 4 & 44 & 86.26 & 3 & $ < 0.001$ & 0.65 \\
\bottomrule
\end{tabular}
\end{table}

\begin{table}[!htbp]
\centering
\caption{\textbf{Paired Wilcoxon signed-rank tests comparing each method against GRACE.}}
\label{tab:wilcoxon_results}
\begin{tabular}{llccc}
\toprule
\textbf{Analysis set} & \textbf{Comparison} & \textbf{Paired tasks, $n$} & \textbf{GRACE higher} & \textbf{Holm-adjusted \textit{P}} \\
\midrule
All tasks & GRACE vs CONCH & 72 & 72/72 & $<0.001$ \\
All tasks & GRACE vs UNI & 72 & 72/72 & $<0.001$ \\
All tasks & GRACE vs Virchow2 & 72 & 70/72 & $<0.001$ \\
Internal & GRACE vs CONCH & 28 & 28/28 & $<0.001$ \\
Internal & GRACE vs UNI & 28 & 28/28 & $<0.001$ \\
Internal & GRACE vs Virchow2 & 28 & 26/28 & $<0.001$ \\
External & GRACE vs CONCH & 44 & 44/44 & $<0.001$ \\
External & GRACE vs UNI & 44 & 44/44 & $<0.001$ \\
External & GRACE vs Virchow2 & 44 & 44/44 & $<0.001$ \\
\bottomrule
\end{tabular}
\end{table}

\begin{table}[!htbp]
\centering
\begin{threeparttable}
\captionsetup{justification=raggedright,singlelinecheck=false,width=\linewidth}
\caption{\textbf{Conditional regression models of AI-associated correction among initially incorrect diagnoses and diagnostic harm among initially correct diagnoses.}}
\label{tab:transition_models}
\begin{tabular}{llcc}
\toprule
\textbf{Model} & \textbf{Variable} & \textbf{OR (95\% CI)} & \textbf{$P$ value} \\
\midrule
\multicolumn{4}{l}{\textbf{AI-associated correction model}} \\
 & Experience level & 1.03 (0.52--2.02) & 0.934 \\
 & Crossover group & 1.51 (0.80--2.82) & 0.201 \\
 & Atrophy diagnosis & 0.77 (0.50--1.17) & 0.224 \\
 & Intestinal metaplasia diagnosis & 1.31 (0.62--2.75) & 0.482 \\
\addlinespace
\multicolumn{4}{l}{\textbf{AI-associated diagnostic harm model}} \\
 & Experience level & 0.36 (0.16--0.82) & 0.015 \\
 & Crossover group & 2.25 (0.88--5.71) & 0.089 \\
 & Atrophy diagnosis in chronic gastritis & 1.10 (0.59--2.03) & 0.769 \\
 & Intestinal metaplasia diagnosis & 0.72 (0.23--2.30) & 0.586 \\
\bottomrule
\end{tabular}
\begin{tablenotes}
\footnotesize
\item Abbreviations: OR, odds ratio; CI, confidence interval.
\item ORs were estimated using GEEs with a logit link and an exchangeable working correlation structure, clustered by pathologist and adjusted for experience level, crossover sequence, and diagnostic task.
\item Pathologist experience was coded as senior versus junior, crossover sequence as Group B versus Group A, and diagnostic task using indicator variables with \textit{acute activity severity grading} as the reference task. 
\end{tablenotes}
\end{threeparttable}
\end{table}
\FloatBarrier
  
\end{document}